\documentclass[twoside,11pt]{article}

\pdfoutput=1 

\usepackage{jmlr2e}
\usepackage{amssymb}
\usepackage{amsmath}
\usepackage{parskip}
\usepackage{multirow}

\usepackage{amssymb}
\usepackage{amsmath}
\usepackage{parskip}
\usepackage{multirow}

\usepackage{ocgx2}
\usepackage{media9}
\usepackage{pdfbase}
\usepackage{animate}
\usepackage{graphicx}
\usepackage{graphics}
\usepackage {framed}
\usepackage {amsxtra}
\usepackage {enumerate}
\usepackage{commath}
\usepackage[margin=1 in]{geometry}
\usepackage{float}
\usepackage[caption = false]{subfig}

\usepackage{epstopdf}

\usepackage{pdfpages}
\usepackage{bm}
\usepackage[utf8]{inputenc}
\usepackage[english]{babel}

\DeclareMathOperator*{\argminB}{argmin}   

\ShortHeadings{Learning with Nonconvex Penalties}{Vettam and John}
\firstpageno{1}

\begin{document}

\title{Regularized Deep Learning with Nonconvex Penalties}

\author{\name Sujit Vettam \email svettam@uchicago.edu \\
       \addr The University of Chicago\\
       Booth School of Business\\
       Chicago, IL, USA
       \AND
       \name Majnu John \email majnu.john@hofstra.edu \\
       \addr Feinstein Institute of Medical Research\\
             Northwell Health System, Manhasset, NY, USA \\
             Department of Mathematics, Hofstra University\\
             Hempstead, NY, USA}

\editor{(arXiv version)}

\maketitle

\begin{abstract}
 Regularization methods are often employed in deep learning neural networks (DNNs) to prevent overfitting. For penalty based DNN regularization methods, convex penalties are typically considered because of their optimization guarantees.  Recent theoretical work have shown that nonconvex penalties that satisfy certain regularity conditions are also guaranteed to perform well with standard optimization algorithms. In this paper, we examine new and currently existing nonconvex penalties for DNN regularization. We provide theoretical justifications for the new penalties and also assess the performance of all penalties with DNN analyses of seven datasets.
\end{abstract}


\section{Introduction}

The success of DNNs in learning complex relationships between inputs and outputs may be mainly attributed to multiple non-linear hidden layers (Rumelhart, Hinton and Williams, 1986a, 1986b). As a consequence of having multiple layers, DNNs typically have tens of thousands of parameters, sometimes even millions. Such large number of parameters gives the method incredible amount of flexibility. However on the downside, this may lead to overfitting the data, especially if the training sample is not large enough. Overfitting means that the method may work well in the training set but not in the test set. Since overfitting is a typical problem for DNNs, many methods have been suggested to reduce it. Adding weight penalties to the cost function, \textit{Dropout}, early stopping, max-norm regularization and data augmentation are some of the popular regularization methods used to avoid overfitting. In this paper, we narrow our focus to regularization methods based on weight penalties appended to the cost function.

Two most commonly considered penalties for DNN regularization are the $L_{1}$ and $L_{2}$ penalties. In statistical literature, these two penalties are known as Lasso (Tibshirani, 1996) and Ridge penalties (Hoerl and Kennard, 1970; Frank and Friedman, 1993) respectively. One of the main advantages of working with these two penalties is convexity of the optimization problem which guarantees that a local optimum will always be a global optimum. $L_{1}$ penalization is also a selection procedure as it sets many parameters to zero. $L_{2}$ penalization does not have this property. All the parameters after $L_{2}$ penalization and all non-zero parameters after $L_{1}$ penalization are shrunk towards zero. The resulting bias in the regularized solution of the above convex penalties has motivated a few authors to consider nonconvex penalties (Fan and Li, 2001; Zhang, 2010), which have the potential to yield nearly unbiased estimates for the parameters. Recent theoretical works (Loh and Wainwright, 2015; Loh and Wainwright, 2017) have also shown that although nonconvex regularizers may yield multiple local optima they are essentially as good as a global optimum from a statistical perspective.

We present nonconvex penalty functions which could be used as regularizers of the weight parameters in a DNN. In the next section we motivate the definition of these penalty functions based on the $L_{0}$ norm. The main focus of our paper is to compare nonconvex penalties with convex penalties in terms of their performance for DNN regularization. We provide theoretical justifications for our proposed regularization approaches and also assess their performance on seven datasets.

\section{Background and Motivation}

In this section we present the DNN-regularization methods proposed in this paper along with background and motivation. Since we will be referencing Loh and Wainwright's 2015 paper frequently, we will use `LW15' as a short name henceforth, for convenience.

\subsection{Background}

Consider a classifier $f_{\bm{w}}: \bm{x} \rightarrow \bm{y}$ parameterized by the weight vector $\bm{w}$, for input $\bm{x}$ and output $\bm{y}$. Optimal weights in a non-regularized setting are obtained by minimizing a cost function $\mathcal{L}(\bm{w})$. Typically the negative log-likelihood is taken as the cost function; in the case of a categorical output it will be the cross-entropy function. One general approach for regularizing DNNs is to append a penalty function $P_{\theta}(\bm{w})$ to the cost function, where $\theta$ denotes the vector of tuning parameters associated with the penalty function. As done in most of the literature we will be restricting our attention to co-ordinate separable penalty functions which could be expressed as a sum \[\displaystyle P_{\theta}(\bm{w}) = \sum_{j=1}^{p}p_{\theta}(w_{j}),\; \bm{w} = (w_{1}, \ldots, w_{p}) \in \mathbb{R}^{p}. \] Thus the regularized optimization problem that we are interested in is \begin{equation} \argminB_{\bm{w} \in \mathbb{R}^{p}} \{\mathcal{L}(\bm{w}) + P_{\theta}(\bm{w}) \} = \argminB_{\bm{w} \in \mathbb{R}^{p}} \{\mathcal{L}(\bm{w}) + \sum_{j=1}^{p} p_{\theta}(w_{j}) \}. \label{B1} \end{equation} The most commonly discussed approach, known as the `canonical selection procedure' is based on \begin{equation} P_{\theta}(\bm{w}) = \sum_{j=1}^{p}I(w_{i} \neq 0),\;\, \mathrm{where\,} I\, \mathrm{denotes\, the\, indicator\, function}; \label{B2} \end{equation} the penalty function in this case is referred to as the $L_{0}$ `norm'. The key ideas behind Akaike's, Bayesian and Minimax Risk based Information Criteria (AIC, BIC and RIC), and Mallow's $C_{p}$ are all based on the above $L_{0}$ norm. However, it is intractable for DNN applications because finding the minimum of the objective function in (1) with the penalty function in (2) is in general NP hard. It is combinatorial in nature and has exponential complexity as it requires an exhaustive search of order $O(2^{p})$.

The above-mentioned intractability has led to considerations of approximations for the penalty function in (2). The most widely considered approximations are of the class of Bridge functions (Fu, 1998; Knight and Fu, 2000) \[ \displaystyle \sum_{j=1}^{p} \abs{w_{j}}^{\kappa}, \; \kappa > 0, \] motivated by the fact that \[ \lim_{\kappa \rightarrow 0} \sum_{j=1}^{p} \abs{w_{j}}^{\kappa} = \sum_{j=1}^{p}I(w_{j} = 0). \] $\kappa = 1$ and $\kappa = 2$ cases ($L_{1}$ and $L_{2}$ penalties) are known  in the literature as Lasso and Ridge penalties. Note that the penalty function in (2) is singular at zero and the optimization problem based on it is nonconvex. Bridge penalty functions are convex when $\kappa \geq 1$ and nonconvex for $0 < \kappa < 1$. Bridge functions are singular at zero only in the case $0 < \kappa \leq 1$. Thus Lasso is the only case among the class of Bridge functions which is both convex and has a singularity at origin. Convex relaxation of a nonconvex problem has its advantage in the optimization setting based on the simple fact that the local minimum of a convex function is also a global minimum. Singularity at origin for the penalty function is essentially what guarantees the sparsity of the solution. Here sparsity means setting to zero small estimated weights to reduce model complexity.

Although Lasso has the above-mentioned advantages over other Bridge estimators, it differs from the $L_{0}$ norm in a crucial aspect: where as the $L_{0}$ norm is constant for any nonzero argument, the $L_{1}$ norm increases linearly with the absolute value of the argument. This linear increase results in a bias for the $L_{1}$-regularized solution (Fan and Li, 2001) which in turn could lead to modeling bias. As mentioned in Fan and Li, 2001, in addition to unbiasedness and sparsity, a good penalty function should result in an estimator with continuity property. Continuity is necessary to avoid instability in model prediction (Breiman, 1996). Note that the penalty function in (2) does not satisfy the continuity criterion. None of the Bridge penalty functions satisfy simultaneously all of the preceding three required properties. The solution for Bridge penalties is continuous only when $\kappa \geq 1$. However, when $\kappa > 1$ the Bridge penalties do not produce sparse solutions. When $\kappa = 1$ (Lasso) it produces continuous and sparse solution, but this comes at the price of shifting the resulting estimator by a constant (bias).

The above issues for the Bridge functions have led to considerations of other approximations for the penalty function in (2) (especially nonconvex approximations) with the hope that these new approximations will satisfy (or nearly satisfy) all the three desirable properties mentioned above. In this paper we present two nonconvex approximation functions: \begin{equation} \displaystyle \sum_{j=1}^{p}p_{\theta}(w_{j}) = \lambda\sum_{j=1}^{p}(1 - \varepsilon^{\abs{w_{j}}}), \;\; \theta = (\lambda, \varepsilon),\;\varepsilon \in (0,1), \; \lambda > 0, \label{B3} \end{equation} \begin{equation}\displaystyle \sum_{j=1}^{p}p_{\theta}(w_{j}) = \lambda\sum_{j=1}^{p}\frac{2}{\pi}\mathrm{arctan}(\gamma \abs{w_{j}}), \;\; \theta = (\lambda, \gamma),\;\gamma > 0, \lambda > 0.  \label{B4} \end{equation} The first penalty has appeared previously in the medical imaging literature (under a slightly different guise) in a method for magnetic resonance image reconstruction (Trzasko and Manduca, 2009), and it has been referred to as the Laplace penalty function. See also Lu \textit{et. al.}, 2014. The second penalty based on arctan function has not been considered in the literature so far to the best of our knowledge. Two other nonconvex penalties that currently exist in the literature are the SCAD penalty, \[ \sum_{j=1}^{p}p_{a, \lambda}(w_{i}), \; \mathrm{where}\; a > 2\; \mathrm{and}\; p_{a, \lambda}(t) = \left\{ \begin{array}{cl}
                    \lambda\abs{t}, & \mathrm{for} \; \abs{t} \leq \lambda\\
                  -(t^{2}-2a\lambda\abs{t} + \lambda^{2})/(2(a-1)), & \mathrm{for} \; \lambda < \abs{t} \leq a\lambda, \\ (a+1)\lambda^{2}/2, & \mathrm{for} \; \abs{t} > a\lambda, \end{array} \right. \] developed by Fan and Li, 2001, and the MCP regularizer (Zhang 2010), \[ \sum_{j=1}^{p}p_{b, \lambda}(w_{i}), \; \mathrm{where}\; b > 0\; \mathrm{and}\; p_{b, \lambda}(t) = \mathrm{sign}(t)\lambda \int_{0}^{\abs{t}}\left(1 - \frac{z}{\lambda b} \right)_{+}dz. \]

There are few other nonconvex regularizers that have appeared in the literature that we list below for the sake of completeness. We give brief reasons on why we restrict attention to the above-mentioned nonconvex penalties (Laplace, arctan, SCAD and MCP) in this paper.

One other nonconvex penalty that has appeared in the regularization literature is the Geman-McClure function (Geman and Yang, 1995) \[ p_{\theta}(w_{j}) = p(w_{j}) = \frac{ \abs{w_{j}} }{\sigma + \abs{w_{j}}}. \] Also known as the \textit{transformed} $L_{1}$ penalty, this function is exactly same as the function $p_{1}(w_{j})$ mentioned in section 3 if we replace $\sigma$ with $1/(-\log(\varepsilon))$. As will be shown below, the function $p_{1}(w_{j})$ is related to the Laplace penalty via Lemma 3.1. For the same parameter $\varepsilon$ the nonconvexity for $p_{1}$ is twice as that for the Laplace penalty. It can also be shown that the derivative of the Laplace penalty converges to zero at a faster rate than the that of $p_{1}$. Based on these considerations we did not study the Geman-McClure function in this paper.

Yet another nonconvex penalty that has appeared in the literature is the concave logarithmic penalty \[ p_{\theta}(w_{j}) = p(w_{j}) = \log \left( \frac{\abs{w_{j}}}{\sigma} + 1 \right), \; \sigma > 0. \] This function increases with the absolute value of the argument as in the case of $L_{1}$ and $L_{2}$ penalties; although the increase is at a lower rate than $L_{1}$ and $L_{2}$ for large $\abs{w_{j}}$, it is still an increasing function thereby resulting in bias. Hence we did not consider this latter penalty as well in this paper. $l_{1}-l_{2}$ function (Yin, Lou \textit{et. al.}, 2015) $P_{\theta}(w) = ||w||_{1} - ||w||_{2}$ is another nonconvex penalty; one disadvantage with this penalty is that it is not coordinate separable. Another nonconvex penalty is the \textit{capped}-$L_{1}$ penalty \[ p_{\theta}(w_{j}) = p(w_{j}) = \mathrm{min} \left(\frac{|w_{j}|}{c}, 1 \right), \; c > 0.\] A unique feature of this penalty is that it is non-differentiable at more than one point. Note also that the capped-$L_{1}$ penalty may be viewed as a limiting version of SCAD as $a \rightarrow$ 1. The error-bound-guarantees for local optima based on SCAD and MCP regularization scheme as shown in LW15 (and for Laplace and arctan as will be shown in this paper) are essentially based on convexification of the corresponding penalty function. Although such convexification is not possible for the capped-$L_{1}$ penalty, LW15's bounds for prediction error as well as $l_{1}$ and $l_{2}$ norm based errors may be modified for this penalty. However, such modified theoretical error bounds are weaker than the corresponding ones for penalties such as SCAD, MCP, Laplace and arctan. Finally, we did not consider nonconvex Bridge functions for the empirical studies in this paper because LW15's theoretical guarantees do not hold for such functions, primarily due to the fact that their derivative at zero is infinite.

\subsection{Motivation}

One of the main purposes of this paper is to study empirically the usefulness of regularization based on nonconvex penalties for deep neural networks. In our empirical studies the metric used to evaluate each method's performance is the \textit{test error} also known as \textit{prediction error}. The primary motivation behind considering nonconvex penalties is their potential for reducing bias more than their convex counterparts. It is well-known (see for example section 2.9, Hastie, Tibshirani and Freedman, 2009) that prediction error can be decomposed as the sum of squared bias and variance. Thus from a theoretical perspective, nonconvex penalties, by reducing bias, has the potential to reduce the prediction error as well.

Although regularization based on nonconvex penalties has shown better empirical performance in machine learning frameworks such as support vector machines (e.g. see Zhang \textit{et. al.}, 2005), thorough empirical comparison in a deep learning framework has not been done to the best of our knowledge. Another motivation for our work is to introduce two new nonconvex penalties (Laplace and arctan) to the deep learning literature. Penalties such as SCAD and MCP, although nonconvex, have a slightly more complex functional form than Laplace and arctan as they involve computing integrals or they require `conditional if' statements in practical code-writing to incorporate their different functional forms in different intervals on the real line. Although this is a negligible issue in a small-scale setting, it may potentially add up to non-negligible computational issues or round-off errors in a large-scale setting such as deep learning, eventually leading to sub-optimal performance empirically. Yet another motivation is to illustrate the theoretical guarantees based on recent work by Loh, Wainwright, Negahban, Agarwal and others for the two new nonconvex penalties introduced in this paper.

\section{Properties of Laplace and arctan penalty functions}

In this section we present a few properties satisfied by the Laplace and arctan penalties. There are altogether nine properties, numbered P1 to P9, that we list. We first give a brief overview of why these properties are important and worthwhile considering. The first five properties, P1-P5, are required for the theoretical guarantees presented in LW15. The relevance of LW15's results are as follows. The scheme typically used in papers prior to LW15, for example in Fan and Li, 2001 and Zhang and Zhang, 2012, was to first establish statistical consistency results concerning global minimizer of the formulation in (1) with nonconvex regularization, and then provide specialized algorithms for obtaining specific local optima that are provably close to the global solution. LW15's results showed that \textit{any} optimization algorithm guaranteed to converge to a stationary point of the objective function in (1), suffices.

Property P4 excludes penalties (such as the nonconvex Bridge penalties) which have infinite derivative at zero, and also penalties (such as capped-$L_{1}$) which has points of non-differentiability on the positive real line. The importance of P4, as pointed out in LW15 and Zhang and Zhang, 2012, is this: if a penalty function has an unbounded derivative at zero, then zero is always a local optima of the objective function in (1), and hence there is no hope for the corresponding error bound to be vanishingly small. Thus P4 is a key property necessary for the theoretical guarantees in LW15 to hold. Property P5 known as \textit{weak convexity} (Vial, 1982; Chen and Gu, 2014) is ``a type of curvature constraint that controls the level of nonconvexity of the penalty'' as quoted from LW15. Properties P1-P4 together imply that the penalty function $p_{\lambda}$ ($\equiv p_{\theta}$, but suppressing the notation for the parameters other than $\lambda$ in the vector $\theta$), is $\lambda L$-Lipshitz, where $L$ is defined below. In particular all subgradients and derivatives of $p_{\lambda}$ are bounded in magnitude by $\lambda L$, if P1-P4 are true. Combining P5 with P1-P4, we have $\lambda L||w||_{1} \leq p_{\lambda}(w) + (\mu/2)||w||_{2}^{2}$, a key inequality used to obtain the $l_{2}$-norm based error bound in LW15; $\mu$ is the convexity parameter defined in P5 below.

Properties P1-P3 were considered in Chen and Gu, 2014 and Zhang and Zhang, 2012, as well. Among other things, Zhang and Zhang, 2012 showed that penalties satisfying P1-P3 along with the extraneous subadditive property are bounded by a capped-$L_{1}$ penalty from below and the maximum of the $L_{0}$ and $L_{1}$ penalties from above, up to a factor 2. Yet another proposition from Zhang and Zhang, 2012, gives a $l_{2}$-regularity condition when properties P1-P3 hold. $l_{2}$-regularity condition is a generalization to the high-dimensional setting ($p \gg n$) of the standard regularity condition in the low-dimensional regression setting ($p \leq n$) that the rank of the design matrix is $p$. Under appropriate $l_{2}$-regularity conditions, Zhang and Zhang, 2012, provides further theory that roughly states that the global solution of nonconvex regularization problems is no worse than Lasso in terms of prediction errors. Finally, properties P6-P8 are related to the three desirable qualities of near-unbiasedness, sparsity and continuity recommended in Fan and Li, 2001, for a good penalty function and property P9 is required for a statistical consistency result in the same paper. Note that all the properties P1-P9 have been verified for SCAD and MCP in previous literature (LW15, Fan and Li, 2001 or Zhang, 2010).

\subsection*{Properties of the Laplace penalty function}

We begin with a useful lemma.

\textbf{Lemma 3.1.} For $\varepsilon \in (0, 1]$ and $x \geq 0$, \begin{equation} \varepsilon^{x} \leq \frac{1}{1 - x\log\varepsilon}. \end{equation}

\textbf{Proof:} Let $y = -x\log\varepsilon$. Note that $y \geq 0$ based on the assumptions. Taking logarithm on both sides of the inequality (5), we get $ \displaystyle x\log\varepsilon \leq -\log(1 - x\log\varepsilon).$ Multiplying by -1 on both sides and substituting $y$, we get $y \geq \log(1+y)$. But this follows from the inequality $z-1 \geq \log z$ for all $z > 0$ (in particular for $z \geq 1$) and the fact that $y \geq 0$. \hfill\BlackBox

  We present a few properties satisfied by the penalty function, \[ p_{\lambda}(t) = \lambda p(t) = \lambda (1 - \varepsilon^{\abs{t}}), \;\;\; \varepsilon \in (0,1), \;\;\; \lambda > 0. \]

\noindent (P1) $p_{\lambda}(0) = 0$ and $p_{\lambda}(t)$ is symmetric around zero. It is easily verified.

\noindent (P2) $p_{\lambda}(t)$ is increasing for $t \in [0, \infty)$. It is easy to see that $p'(t) = -\log(\varepsilon)\varepsilon^{t}$ is positive for $\varepsilon \in (0,1)$ and $t \geq 0$.

\noindent (P3) For $t > 0$, the function $g(t) = p(t)/t$ is non-increasing in $t$. Since, for $t > 0$, \[ g'(t) = \frac{t\left[-(\log\varepsilon)\varepsilon^{t} \right] - \left[ 1 - \varepsilon^{t} \right]}{t^{2}} = \frac{\varepsilon^{t}\left[1 - t\log\varepsilon \right] - 1 }{t^{2}} \] it suffices to show that the numerator $\displaystyle \varepsilon^{t}\left[1 - t\log\varepsilon \right] - 1 \leq 0$ for $t > 0$. But this follows from Lemma 3.1 above.

\noindent (P4) The function $p_{\lambda}$ is differentiable for all $t \neq 0$ and subdifferentiable at $t = 0$, with $\displaystyle \lim_{t \rightarrow 0+} p_{\lambda}'(t) = \lambda L, \;\; L = -\log\varepsilon$. It is easy to see that any point in the interval $\left[\lambda\log\varepsilon, -\lambda\log\varepsilon \right]$ is a subgradient of $p_{\lambda}(t)$  at $t = 0$.

\noindent (P5) There exists $\mu > 0$ such that $p_{\lambda, \mu}(t) = p_{\lambda}(t) + (\mu t^{2}/2)$ is convex: $\mu = \lambda(\log\varepsilon)^{2}$ will work. $\mu$ is a measure of the severity of nonconvexity of the penalty function.

  Since the penalty function $p_{\lambda}(t)$ satisfy the properties (P1) to (P5), we have $q_{\lambda}(t) := \lambda\abs{t} - p_{\lambda}(t)$ is everywhere differentiable. These properties also imply that $p_{\lambda}$ is $\lambda L$-Lipschitz as a function of $t$ (LW15). In particular, all subgradients and derivatives of $p_{\lambda}$ are bounded in magnitude by $\lambda L$ (LW15). We also see that for empirical loss $\mathcal{L}_{n}$ satisfying restricted strong convexity condition and conditions for $\lambda$ and sample size in Theorem 1 in LW15, the squared $l_{2}$-error of the estimator grows proportionally with the number of nonzeros in the target parameter and with $\lambda^{2}$. One condition that is not satisfied by $p_{\lambda}(t)$ is (P6) below.

\noindent (P6) There exists $r \in (0, \infty)$ such that $p_{\lambda}'(t) = 0, \forall t > r\lambda.$ It is clear that such a $r$ does not exist for Laplace penalty function. However, we note that $p_{\lambda}'(\abs{t})$ can be made arbitrarily close to zero for large $\abs{t}$. In other words, the following property is satisfied.

\noindent (P6$'$) $\displaystyle \lim_{\abs{t} \rightarrow \infty} p_{\lambda}'(t) = 0.$

  (P6) and (P6$'$) are related to unbiasedness as mentioned in Fan and Li, 2001. (P6) guarantees unbiasedness (and (P6$'$) near-unbiasedness) when the true unknown parameter is large and this in turn avoids unnecessary modeling bias.

Fan and Li, 2001, suggests that in addition to unbiasedness or near-unbiasedness (addressed in P6) a good penalty function $p_{\lambda}(t)$ should also allow for sparsity and continuity. The following two properties, (P7) and (P8) address sparsity and continuity of Laplace penalty.

\noindent (P7) Minimum of the function $\abs{t} + p_{\lambda}'(\abs{t})$ is positive when $\lambda > 1/e(\log \varepsilon)^{2}$ where $e$ is the exponential number. This property guarantees sparsity, at least in the $L_{2}$ empirical loss case. That is, the resulting estimator is a thresholding rule.

\noindent (P8) Minimum of the function $\abs{t} + p_{\lambda}'(\abs{t})$ is attained at $t = 0$ when $\lambda \leq 1/(\log \varepsilon)^{2}$. This property, at least in the $L_{2}$ empirical loss case, is related to the continuity of the resulting estimator. Continuity helps to avoid instability in model prediction.

Note that there is only a small range of $\lambda$ values ($\lambda \in (1/e(\log \varepsilon)^{2}, 1/(\log \varepsilon)^{2})$) for which properties (P7) and (P8) are simultaneously satisfied. For example, when $\varepsilon =$  0.01, the above interval is approximately (0.017, 0.047). It is easy to verify the properties (P7) and (P8) for the corresponding $\lambda$ ranges by considering the function $f(t) = t - (\lambda\log\varepsilon)\varepsilon^{t}$ for $t > 0$. The minimum of the function is attained at \begin{equation} t_{min} = \frac{\log\left[\lambda(\log\varepsilon)^{2}\right]}{(-\log\varepsilon)} \end{equation} and the minimum value is \[ f(t_{min}) = \frac{\log\left[\lambda(\log\varepsilon)^{2} \right] + 1}{(-\log\varepsilon)}. \]


  Theorem 1 in Fan and Li, 2001, provides necessary conditions for the $\sqrt{n}$-consistency of the estimator in a maximum likelihood framework and generalized linear models setting. The main assumption required for the penalty function is stated as the following property.

  \noindent (P9) $\mathrm{max}\left\{\abs{p_{\lambda}''(\abs{t_{s}})}: t_{s} \neq 0, s = 1, \ldots, p \right\} \rightarrow 0\;\;\mathrm{as}\;\; \lambda \rightarrow 0^{+}. $ In our case, this property is satisfied because $\abs{p_{\lambda}''(\abs{t})} = \lambda(\log\varepsilon)^{2}\varepsilon^{\abs{t}} \leq \lambda(\log\varepsilon)^{2} \rightarrow 0\;\;\mathrm{as}\;\; \lambda \rightarrow 0^{+}.$

  Lemma 3.1 suggests considering another penalty function $p_{1,\lambda}(t) = \lambda p_{1}(t)$ where \[ p_{1}(t) = \frac{ -\abs{t}\log\varepsilon }{1 - \abs{t}\log\varepsilon}. \] This penalty function ($p_{1}$) is equivalent to Geman and McClure's penalty function mentioned in Geman and Yang, 1995 (also mentioned in Trzasko and Manduca, 2009 and Lu \textit{et. al.}, 2014). Most of the properties listed above are satisfied by the penalty function $p_{1}$ as well. For example, \[ p_{1}'(t) = \frac{-\log\varepsilon}{(1 - t\log\varepsilon)^{2}} > 0, \;\;\mathrm{for}\;\;t \geq 0\;\;\mathrm{verifying}\;(\mathrm{P}2). \] In order to check (P3) we consider \[ g_{1}(x) = \frac{1}{x(1 - x\log\varepsilon)^{2}}.\] For $0 < x_{1} \leq x_{2}$, it is easy to check that \[ g_{1}(x_{1}) - g_{1}(x_{2}) = \frac{(x_{2}-x_{1})\left\{[1 - (\log\varepsilon)(x_{2} - x_{1})]^{2} + 3(\log\varepsilon)^{2}x_{1}x_{2} \right\} }{x_{1}x_{2}(1 - x_{1}\log\varepsilon)^{2}(1 - x_{2}\log\varepsilon)^{2} } \geq 0. \] \[\mathrm{Also}\;\; p_{1}''(t) = \frac{-2(\log\varepsilon)^{2}}{(1 - t\log\varepsilon)^{3}} < 0\;\;\mathrm{for}\;\;t \geq 0. \]However, this suggests that the $\mu$ required for (P5) is $2\lambda(\log\varepsilon)^{2}$ which is twice as that for $p_{\lambda}(t)$. That is, nonconvexity for $p_{1,\lambda}$ is twice as severe for $p_{\lambda}$. Also $p_{1}'(t)$ converges to zero (as do $p'(t)$) for large $\abs{t}$ satisfying (P6$'$) for near-unbiasedness. However, since it can be shown that \[ \varepsilon^{t} < \frac{1}{(1 - t\log\varepsilon)^{2}} \;\;\mathrm{for\;large}\;t,\;\mathrm{using\;the\;fact\;that\;} \frac{y}{2} > \log(1 + y)\;\;\mathrm{for\;large}\;y, \] we see that the convergence for $p'(t)$ is faster. Hence we did not consider the latter penalty function, $p_{1,\lambda}(t)$ in this paper.

  We may also generalize Laplace penalty function to $\lambda\left(1 - \varepsilon^{\alpha\abs{t}} \right)$ for $\alpha \geq 1$. However, the $\mu$ corresponding to this function will be $\alpha(\log\varepsilon)^{2}$, making it more severely nonconvex similar to the Bridge penalty function when $\kappa < 1$. Hence, in this paper we focus only on $\alpha = 1$ case. Further comparison with Bridge penalty is given in section 4.

\subsection*{Properties of the arctan penalty function} Here we check properties P1-P9 for the arctan penalty,
\[ p_{\lambda}(t) = \lambda p(t) = \lambda \left(\frac{2}{\pi}\mathrm{arctan}(\gamma\abs{t}) \right) , \;\;\; \gamma \in (0,\infty), \;\;\; \lambda > 0. \]

 Property (P1) ( - $p_{\lambda}(0) = 0$ and $p_{\lambda}(t)$ is symmetric around zero - ) is again easily verified.

\noindent (P2): For $t \geq 0$, \[p'(t) = \frac{2\gamma}{\pi}\frac{1}{1 + \gamma^{2}t^{2}}\] is positive for $\gamma > 0$. Hence $p_{\lambda}(t)$ is increasing for $t \in [0, \infty)$.

We state as a lemma a well-known fact about arctan function.

\textbf{Lemma 3.2.} For $y \geq 0$,
 \begin{equation} \frac{y}{1 + y^2} \leq \mathrm{arctan}(y) \leq y. \end{equation}

\textbf{Proof:} If we take $f(y) = y - \mathrm{arctan}(y)$, then $f'(y) = y^{2}/(1 + y^{2}) \geq 0$ for $y \geq 0$ and hence $f(\cdot)$ is non-decreasing in the interval $[0, \infty)$. In particular $f(y) \geq f(0) = 0$, which proves the right inequality. Similarly, by writing $f(y) = \mathrm{arctan}(y) - [y/(1 + y^{2})]$, we have $f'(y) = 2y^{2}/(1 + y^{2})^{2} \geq 0$ for $y \geq 0$, proving the left inequality. \hfill\BlackBox

\noindent (P3) For $t > 0$ consider the function $g(t) = p(t)/t$. \[ g'(t) = \frac{2}{\pi t^{2}} \left[\frac{\gamma t}{1 + \gamma^{2}t^{2}} - \mathrm{arctan}(\gamma t) \right]. \] Thus $g'(t) \leq 0$ by the above lemma and hence $p(t)/t$ is non-increasing.

\noindent (P4) $\displaystyle \lim_{t \rightarrow 0+} p_{\lambda}'(t) = \lambda L, \;\; L = 2\gamma/\pi$. Any point in the interval $\left[-2\lambda\gamma/\pi, 2\lambda\gamma/\pi\right]$ is a subgradient of $p_{\lambda}(t)$  at $t = 0$.

\noindent (P5) $\mu = 2\lambda\gamma^{2}/\pi > 0$ makes $p_{\lambda, \mu}(t) = p_{\lambda}(t) + (\mu t^{2}/2)$ convex.

\noindent (P6') It is easy to check that $\displaystyle \lim_{\abs{t} \rightarrow \infty} p_{\lambda}'(t) = 0.$

\noindent (P7) By considering the function $f(t) = t + [2\lambda\gamma/\pi(1 + \gamma^{2}t^{2})]$ for $t > 0$, it is easy to verify that minimum of $\abs{t} + p_{\lambda}'(\abs{t})$ is positive for all $\lambda > 0$. In other words, arctan penalty yields sparse solutions for \textit{all} $\lambda > 0$. Note that it was slightly different for the Laplace case where (P8) did not hold in a small range, ($0, 1/e(\log \varepsilon)^{2}$), of $\lambda$ values.

\noindent (P8) The minimum of $\abs{t} + p_{\lambda}'(\abs{t})$ is attained at zero when $\lambda < \pi/\gamma^{2}$. Thus for $\lambda < \pi/\gamma^{2}$, the arctan penalty yields stable solutions for the $L_{2}$ empirical loss. Unlike the Laplace case, it is not straightforward to verify this property analytically, but can be easily checked by plotting the function $f$.

Note also that only for $\lambda$ values in the interval ($0, \pi/\gamma^{2}$) are both properties (P7) and (P8) simultaneously satisfied. Also easy to see that $\displaystyle \abs{ p_{\lambda}''(t) } \leq 2\lambda\gamma^{2}/\pi \rightarrow 0$ which guarantees that (P9) is satisfied.

\section{Ordinary Least Squares Linear Regression}

  Although the focus of this paper is on applying nonconvex penalties in the deep learning setting, studying these penalties in ordinary least squares (OLS) linear regression setting will lead to deeper insights. First we will try to get a visual understanding, especially by considering the orthonormal design case. Secondly, we will study the asymptotic properties of the Laplace and arctan penalties, which will help to delineate them from convex and nonconvex Bridge penalties.

  In this section, our interest is on the linear regression model \begin{equation} \bm{y} = \bm{X}\bm{w} + \bm{e}. \label{OLS1} \end{equation} Let $\hat{\bm{w}}^{O}$, $\hat{\bm{w}}^{L_{\kappa}}$, $\hat{\bm{w}}^{\mathrm{Lapl.}}$ and $\hat{\bm{w}}^{\mathrm{atan}}$ represent the OLS estimates for the regression problem in (\ref{OLS1}) with no penalty, Bridge, Laplace and arctan penalties, respectively. That is, assuming $\bm{X}^{T}\bm{X}$ is invertible, $\bm{\hat{w}^{O}} = \left(\bm{X}^{T}\bm{X}\right)^{-1}\bm{X}^{T}y,$ and
  \[ \bm{\hat{w}^{L_{\kappa}}} = \argminB_{\bm{w}} \left\{ \sum_{i=1}^{n} (y_{i} - \bm{x}_{i}^{T}\bm{w})^{2} + \lambda \sum_{j=1}^{p}\abs{w_{j}}^{\kappa} \right\};\;\;\mathrm{e.g.}\;\kappa = 1\;\;\mathrm{for\;Lasso},\]
  \begin{equation} \hat{\bm{w}}^{\mathrm{Lapl.}} = \argminB_{\bm{w}} \left\{ \sum_{i=1}^{n} (y_{i} - \bm{x}_{i}^{T}\bm{w})^{2} + \lambda \sum_{j=1}^{p}\left(1 - \varepsilon^{\abs{w_{j}}} \right) \right\}  \label{OLS.Lapl} \end{equation} and  \begin{equation} \hat{\bm{w}}^{\mathrm{atan}} = \argminB_{\bm{w}} \left\{ \sum_{i=1}^{n} (y_{i} - \bm{x}_{i}^{T}\bm{w})^{2} + \lambda \sum_{j=1}^{p} (2/\pi)\arctan (\gamma w_{j}) \right\}  \label{OLS.atan} \end{equation}

    \begin{figure}[H]
  \begin{center}
  \hspace*{-0.6cm}
  \includegraphics[height=4in,width=6.5in,angle=0]{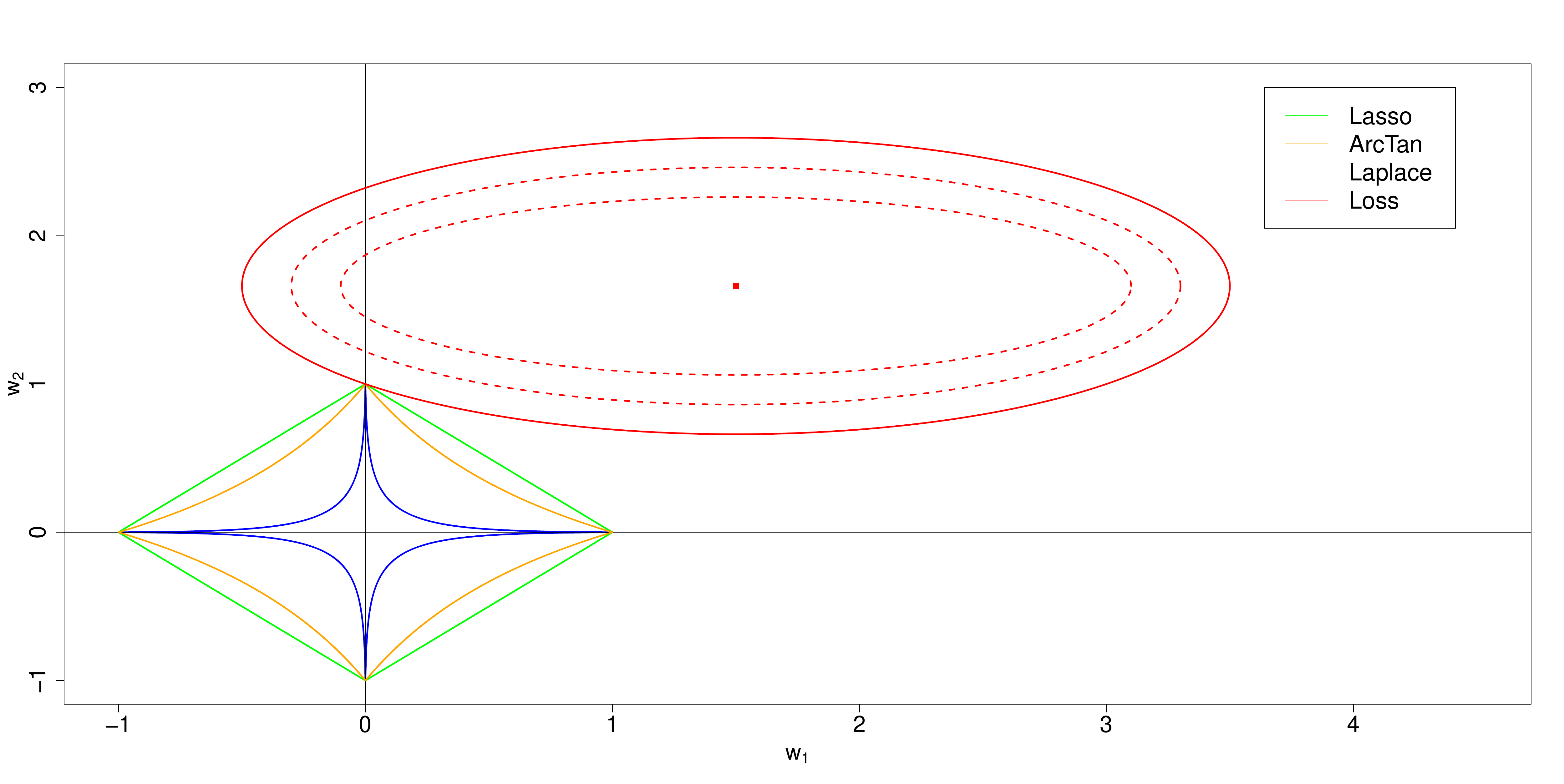}
  \caption{Contours of the quadratic form of OLS loss function (red curves) in the 2-dimensional case and constraints based on Lasso (green), Laplace with $\varepsilon$ = 0.01 (blue) and arctan with $\gamma$ = 1 (orange). The solution is where the loss function contours meet the constraints.}
  \end{center}
  \end{figure}

  The quadratic form of the objective function corresponding to the OLS loss $||y - \bm{X}\bm{w}||_{2}^{2}$ is ($\bm{w} - \hat{\bm{w}}^{O}$)$^{T}$($\bm{X}^{T}\bm{X}$)($\bm{w} - \hat{\bm{w}}^{O}$). For the special case $p = 2$, $\bm{w} =$ ($w_{1}, w_{2}$)$^{T}$ and the resulting elliptical contour lines are centered around the OLS estimate (see, for example, red lines plotted in Figure 1). The Laplace and arctan penalized estimates for the OLS loss solves the objective function under the constraints $\sum_{j=1}^{p}\left(1 - \varepsilon^{\abs{w_{j}}} \right) \leq s$ and $\sum_{j=1}^{p}\left((2/\pi)\arctan (\gamma w_{j}) \right) \leq s$, respectively. For the example plotted in Figure 1, the constraints space corresponding to the Laplace and arctan penalties are the areas within the blue curves and orange curves, respectively. The Laplace (or arctan) penalized OLS solution is the first point that the elliptical red contours touch the blue lines (or the orange lines) corresponding to the constraints and this sometimes occurs at a corner as in the plotted example, yielding a zero coefficient.

  \subsection*{Orthonormal design case} In the orthonormal design case $\bm{X}^{T}\bm{X} = I$ so that $\hat{\bm{w}}^{O} = \bm{X}^{T}\bm{y}$; this special case is particularly useful to obtain a clearer idea about the basic nature of the penalized estimators, by visual means. It is easy to show (Hardle and Simar, 2015, Chapter 7) that in the orthonormal case \begin{equation} \hat{w}^{L_{1}}_{j} = \mathrm{sign}\left(\hat{w}^{O}_{j}\right)\left(\abs{\hat{w}^{O}_{j}} - \frac{\lambda}{2} \right)^{+}, \; j \in \{1, \ldots, p\}.  \label{OD1} \end{equation} Also, in the orthonormal case, \begin{equation} \hat{w}_{j}^{\mathrm{Lapl.}} = \argminB_{w_{j}}  \left\{ -2\hat{w}^{O}_{j}w_{j} + w_{j}^{2} + \lambda\left(1 - \varepsilon^{\abs{w_{j}}} \right) \right\} \label{OD2} \end{equation} and \begin{equation} \hat{w}_{j}^{\mathrm{atan}} = \argminB_{w_{j}}  \left\{ -2\hat{w}^{O}_{j}w_{j} + w_{j}^{2} + \lambda\left((2/\pi)\arctan (\gamma\abs{w_{j}}) \right) \right\}. \label{OD3} \end{equation} When $\hat{w}^{O}_{j} > 0$, $\hat{w}_{j}^{\mathrm{Lapl.}}$ will be a solution to the following equation in the unknown variable $w$: \begin{equation} w = \hat{w}^{O}_{j} + \frac{\lambda}{2} \varepsilon^{w} \log \varepsilon. \label{OD4} \end{equation} Note that $\hat{w}^{O}_{j} > 0$ will force $\hat{w}_{j}^{\mathrm{Lapl.}} > 0$ since $w_{j}^{2} + \lambda (1 - \varepsilon^{|w_{j}|})$ is symmetric about zero and $-2\hat{w}^{O}_{j}w < -2\hat{w}^{O}_{j}(-w)$ for any $w > 0$. A similar argument shows that $\hat{w}^{O}_{j} < 0$ forces $\hat{w}_{j}^{\mathrm{Lapl.}} < 0$ and in this case $\hat{w}_{j}^{\mathrm{Lapl.}}$ will be a solution to \begin{equation} w = \hat{w}^{O}_{j} - \frac{\lambda}{2} \varepsilon^{-w} \log \varepsilon. \label{OD5} \end{equation} Combining the cases $\hat{w}^{O}_{j} > 0$ and $\hat{w}^{O}_{j} < 0$ (that is, combining equations (\ref{OD4}) and (\ref{OD5})), we may write $w_{j}^{\mathrm{Lapl.}}$ as the solution to \begin{equation} w = \hat{w}^{O}_{j} + \frac{\lambda}{2}\mathrm{sign} (\hat{w}^{O}_{j}) \varepsilon^{|w|} \log \varepsilon.  \label{OD6} \end{equation}  Using arguments similar to above, we may show $w_{j}^{\mathrm{atan}}$ is the solution to \begin{equation} w = \hat{w}^{O}_{j} + \frac{\gamma\lambda}{2}\mathrm{sign} (\hat{w}^{O}_{j}) \frac{1}{1 + \gamma^{2}w^{2}}.  \label{OD7} \end{equation}

  A better illustration of the difference between the Lasso solution and the solutions based on nonconvex penalties such as Laplace and arctan can be obtained by considering equations (\ref{OD1}), (\ref{OD2}) and (\ref{OD3}) for various values of $\lambda$. As a specific example, we consider the OLS solution $\hat{w}^{O}_{j}$ = 3 and $\lambda$ values on the grid $\{0.1, 1,2,3,\ldots,14,15\}$. The objective functions in (\ref{OD1}), (\ref{OD2}) and (\ref{OD3}) for these specific values of $\hat{w}^{O}_{j}$ and $\lambda$'s are plotted as animation movies in figures 2, 3 and 4, respectively. These animation movies can be played by clicking the triangle pointing to the right inside the 3$^{rd}$ button in the panel at the bottom (after downloading the pdf file for this paper). For the Laplace and arctan penalties considered in figures 3 and 4, we chose $\varepsilon =$ 0.01 and $\gamma$ = 10, respectively. Slower moving animation movies (with larger file sizes) corresponding to $\lambda$-spacing of 0.1 are available at the following web-appendix page for this paper: https://www.ncr-dnn.com/

\begin{figure}[H]
\begin{center}
\animategraphics[label=lassoskipwlambdai, height=3in, width=3.35in, controls, timeline=timelinesmall.txt]{4}{lassoskipwlambdai.}{1}{16}\caption{Animation movie showing one minima for the objective function $-2\hat{w_{0}}w + w^{2} + \lambda\abs{w}$ with $\hat{w}_{0} = 3$, as $\lambda$ increases from 0.1 to 15. A green vertical line through $w = \hat{w}_{0}$ is plotted for reference.}
\mediabutton[
jsaction={
if(anim[’lassoskipwlambdai’].isPlaying)
anim[’lassoskipwlambdai’].pause();
else
anim[’lassoskipwlambdai’].playFwd();
}
]{\fbox{Play/Pause}}
\end{center}
\end{figure}

In the Lasso case (Figure 2), when $\lambda$ is small ($\lambda =$ 0.1, for example), the argument of local minimum is close to the OLS solution $\hat{w}^{O}_{j}$ = 3. As $\lambda$ increases, the argument of local minimum shrinks and moves more and more closer to zero. After a particular $\lambda$-threshold (for some $\lambda$ value near 6, in the example in the Figure 2) the argument of local minimum is zero and stays at zero thereafter. Thus depending on whether $\lambda$ is below or above the threshold, one gets either a biased value or zero as the Lasso solution. Also note that in the Lasso case, because of convexity, the local minimum is always the global minimum.

In the Laplace case, as $\lambda$ increases, two local minima appear: one at zero and the other at or near $\hat{w}^{O}_{j}$ = 3. Thus an optimization algorithm seeking a local minimum will land at either zero or at the OLS solution.  For smaller values of $\lambda$ (that is, for values of $\lambda$ below a threshold), the global minimum is at $\hat{w}^{O}_{j}$ = 3; $\lambda$-threshold = 9 for the example plotted in Figure 3. Above the threshold, the local minimum at zero becomes the global minimum. Thus in the perfect world with an optimization algorithm that always correctly reaches the global minimum, we will get the unbiased (or nearly unbiased) solution for smaller values of $\lambda$ and the zero solution for larger values of $\lambda$. In other words, the regression weights which are `selected' at any $\lambda$ (i.e. the non-zero weights) will be nearly unbiased. The conclusions for the arctan case plotted in Figure 4 are also very similar.

\begin{figure}[H]
\begin{center}
\animategraphics[label=laplskipwlambdai, height=3in, width=3.35in, controls, timeline=timelinesmall.txt]{4}{laplskipwlambdai.}{1}{16}\caption{Animation movie showing two minima, one at 0 and the other at $\hat{w}_{0}$ for the objective function $-2\hat{w_{0}}w + w^{2} + \lambda(1 - \varepsilon^{\abs{w}})$ with $\hat{w}_{0} = 3$, as $\lambda$ increases from 0.1 to 15. We took for $\varepsilon = 0.01$ for this plot. A green vertical line through $w = \hat{w}_{0}$ is plotted for reference.}
\mediabutton[
jsaction={
if(anim[’laplskipwlambdai’].isPlaying)
anim[’laplskipwlambdai’].pause();
else
anim[’laplskipwlambdai’].playFwd();
}
]{\fbox{Play/Pause}}
\end{center}
\end{figure}

\subsection*{Asymptotics of Laplace and arctan solutions}

  In this subsection we state and prove three propositions to better understand the asymptotic properties of Laplace and arctan solutions. The first proposition is a consistency result, a theoretical asymptotic justification for the Laplace and arctan solutions. The second proposition is a $\sqrt{n}$-consistency result which helps to shed light onto the asymptotic biases of Laplace and arctan solutions. Since asymptotic biases for Lasso and nonconvex Bridge functions are already known from previous literature (see theorems 2 and 3 in Knight and Fu, 2000), our Proposition-2 will help us to compare the penalties in terms of their respective asymptotic biases. Specifically, we will see that unlike nonconvex Bridge penalties, Laplace and arctan penalties \textit{do} have asymptotic biases. However, we will also see that the asymptotic biases of Laplace and arctan penalties are substantially smaller compared to convex Bridge penalties. The third
  proposition below shows a consistency result similar to one proved in Huang, Horowitz and Ma, 2008, for nonconvex Bridge functions, where $p$ (the column dimension of the design matrix $\bm{X}$) is allowed to grow with $n$. The proof of Proposition-3 (or rather the assumptions required for the proof) nicely brings out the key differences between the new penalties introduced in this paper and nonconvex Bridge penalties. We also present a statistical consistency result for the logistic loss function in the Appendix-A by adapting slightly the theorems presented in Tarigan and van de Geer, 2006 and Meier, van de Geer and Buehlmann, 2008.

\begin{figure}[H]
\begin{center}
\animategraphics[label=lambdai, height=3in, width=3.35in, controls, timeline=timelinesmall.txt]{4}{atanskipwlambdai.}{1}{16}\caption{Animation movie showing two minima, one at 0 and the other at $\hat{w}_{0}$ for the objective function $-2\hat{w_{0}}w + w^{2} + \frac{2\lambda}{\pi}\arctan (\gamma w)$ with $\hat{w}_{0} = 3$, as $\lambda$ increases from 0.1 to 15. We took for $\gamma = 10$ for this plot. A green vertical line through $w = \hat{w}_{0}$ is plotted for reference.}
\mediabutton[
jsaction={
if(anim[’atanskipwlambdai’].isPlaying)
anim[’atanskipwlambdai’].pause();
else
anim[’atanskipwlambdai’].playFwd();
}
]{\fbox{Play/Pause}}
\end{center}
\end{figure}

  For the first two propositions we assume the following regularity conditions for the design matrix, $\bm{X}$ (as in Knight and Fu, 2000). If $\bm{x_{i}}$ denote the $i^{th}$ column of $\bm{X}^{T}$, we assume \[ C_{n} = \frac{1}{n} \sum_{i=1}^{n} \bm{x_{i}}\bm{x_{i}}^{T} \rightarrow C, \] where $C$ is a nonnegative definite matrix and \[ \frac{1}{n} \max_{1 \leq i \leq n} \bm{x_{i}}^{T}\bm{x_{i}} \rightarrow 0. \] Proposition-1 focuses only on Laplace solutions. However, a similar result can be easily obtained for arctan solutions as well. For Proposition-1 we include the subscript $n$ to emphasize the dependence on $n$ and skip the superscript \textit{Lapl.} (that is we write, $\bm{\hat{w}}_{n}$ for $\bm{\hat{w}}^{\mathrm{Lapl.}}$ ). Also we denote the random function in the definition of $\bm{\hat{w}}^{\mathrm{Lapl.}}$ as $Z_{n}(\bm{\phi})$: \[ Z_{n}(\bm{\phi}) = \frac{1}{n}\sum_{i=1}^{n} (y_{i} - \bm{x}_{i}^{T}\bm{\phi})^{2} + \frac{\lambda_{n}}{n} \sum_{j=1}^{p}\left(1 - \varepsilon^{\abs{\phi_{j}}} \right). \] We label theorems 1, 2 and 3 in Knight and Fu (2000) as Th-KF-1, Th-KF-2 and Th-KF-3.

  \noindent \textbf{\underline{Proposition-1}}: We assume $C$ is non-singular, $0 < \varepsilon < 1$, the components $e_{i}$'s of the error vector in (\ref{OLS1}) are i.i.d with mean zero and variance $\sigma^{2}$. If $\lambda_{n}/n \rightarrow \lambda_{0}\; (\lambda_{n}, \lambda_{0} \geq 0)$ then $\bm{\hat{w}}_{n} \rightarrow_{p} \argminB (Z)$ where \[ Z(\bm{\phi}) = (\bm{\phi} - \bm{w})^{T}C(\bm{\phi} - \bm{w}) + \lambda_{0} \sum_{j=1}^{p}\left(1 - \varepsilon^{\abs{\phi_{j}}} \right). \] Since $\argminB (Z) = \bm{w}$, $\bm{\hat{w}}_{n}$ is consistent provided $\lambda_{n} = o(n)$.

  \textbf{Proof:}
  Proof is easily adapted from proof of Th-KF-1 for Bridge  estimators. We write \begin{align*} Z_{n}(\bm{\phi}) &= \frac{1}{n}\sum_{i=1}^{n} \left[ (y_{i} - \bm{x}_{i}^{T}\bm{w}) - \bm{x}_{i}^{T}(\bm{\phi} - \bm{w})\right]^{2} + \frac{\lambda_{n}}{n} \sum_{j=1}^{p}\left(1 - \varepsilon^{\abs{\phi_{j}}} \right) \\  &= \frac{1}{n}\sum_{i=1}^{n} \left[ e_{i}^{2} - 2e_{i}\bm{x}_{i}^{T}(\bm{\phi} - \bm{w}) + (\bm{\phi} - \bm{w})^{T}\bm{x}_{i}\bm{x}_{i}^{T}(\bm{\phi} - \bm{w}) \right] + \frac{\lambda_{n}}{n} \sum_{j=1}^{p}\left(1 - \varepsilon^{\abs{\phi_{j}}} \right). \end{align*} \begin{align*} \mathbb{E}(Z_{n}(\phi)) &= \frac{1}{n} \sum_{i=1}^{n}\left(\sigma^{2} - 0 + \left[ (\bm{\phi} - \bm{w})^{T}\bm{x}_{i}\bm{x}_{i}^{T}(\bm{\phi} - \bm{w}) \right] \right) + \frac{\lambda_{n}}{n} \sum_{j=1}^{p}\left(1 - \varepsilon^{\abs{\phi_{j}}} \right)\\ &= \sigma^{2} + (\bm{\phi} - \bm{w})^{T}C_{n}(\bm{\phi} - \bm{w}) + \frac{\lambda_{n}}{n} \sum_{j=1}^{p}\left(1 - \varepsilon^{\abs{\phi_{j}}} \right)\\ &\rightarrow \sigma^{2} + Z(\bm{\phi}), \;\; \mathrm{as\;} n \rightarrow \infty,\;\;\mathrm{since}\; C_{n} \rightarrow C \;\;\mathrm{and}\;\; \lambda_{n}/n \rightarrow \lambda_{0}. \end{align*} Convergence in mean implies pointwise convergence in probability. We claim \begin{equation} \sup_{\phi \in K} \abs{Z_{n}(\phi) - Z(\phi) - \sigma^{2}} \rightarrow_{p} 0, \;\; \mathrm{for \; any\; compact\; set\;} K. \label{Prop1.eq1}\end{equation} Since $Z_{n}$ is nonconvex we cannot use the \textit{Convexity Lemma} (presented in Pollard, 1991) to prove the claim. An alternate route to establish the claim (based on stochastic Ascoli lemma) is by showing \textit{a.s.} equicontinuity for $Z_{n}$. A sufficient condition for stochastic equicontinuity is \textit{a.s.} Lipschitz continuity for $Z_{n}$. Since the OLS loss function is convex (and hence Lipschitz continuous over a compact set $K$) and since we have established Lipschitz property for the Laplace penalty function previously in section 2, the uniform convergence claim in (\ref{Prop1.eq1}) follows. Since we assume $\lambda_{n} \geq 0$ and $\varepsilon \in (0,1)$, we have \[ Z_{n}(\phi) \geq \frac{1}{n} \sum_{i=1}^{n} \left( y_{i} - \bm{x}_{i}^{T}\bm{\phi} \right)^{2} = Z_{n}^{(0)}(\bm{\phi}), \;\; \mathrm{for\;all}\; \bm{\phi}. \] Since $\argminB (Z_{n}^{(0)}) = O_{p}(1)$, we get \begin{equation} \bm{\hat{w}}_{n} = O_{p}(1). \label{Prop1.eq2} \end{equation} Equations (\ref{Prop1.eq1}) and (\ref{Prop1.eq2}) imply that $\bm{\hat{w}}_{n}$ is consistent.
  \hfill\BlackBox

The next proposition is a $\sqrt{n}$-consistency result for the penalized estimators in (\ref{OLS.Lapl}) and (\ref{OLS.atan}). The proof follows closely the proof for Th-KF-3. Again we skip the superscripts $Lapl.$ and $atan$ with the hope that it will be understood from the context.

\noindent \textbf{\underline{Proposition-2}}: If $\lambda_{n}/\sqrt{n} \downarrow \lambda_{0} \geq 0$, then $\sqrt{n}(\hat{\bm{w}}_{n} - \bm{w}) \rightarrow_{d} \mathrm{argmin}(V)$, where $V = V^{\mathrm{Lapl.}}$ in the case of Laplace penalty and $V = V^{\mathrm{atan}}$ in the case of arctan penalty. Here
\begin{align} V^{\mathrm{Lapl.}}(\mathbf{u}) &= -2\mathbf{u}^{T}W + \mathbf{u}^{T}C\mathbf{u} + \lambda_{0}(-\log
(\varepsilon))\sum_{j=1}^{p}\left[u_{j}\mathrm{sign}(w_{j})\varepsilon^{|w_{j}|}I(w_{j}\neq 0) + |u_{j}|I(w_{j} = 0) \right] \\
 V^{\mathrm{atan}}(\mathbf{u}) &= -2\mathbf{u}^{T}W + \mathbf{u}^{T}C\mathbf{u} + \frac{2\gamma\lambda_{0}}{\pi} \sum_{j=1}^{p} \left[\frac{u_{j}\mathrm{sign}(w_{j})}{1 + \gamma^{2}w_{j}^{2}}I(w_{j} \neq 0) + |u_{j}|I(w_{j} = 0) \right] \end{align} and $W$ has a $N(\mathbf{0}, \sigma^{2}C)$ distribution.

 \textbf{Proof:}  Consider \[ V_{n}^{\mathrm{Lapl.}}(\mathbf{u}) = \sum_{i=1}^{n}\left[(e_{i} - \mathbf{u}^{T}\bm{x}_{i}/\sqrt{n})^{2} - e_{i}^{2} \right] + \lambda_{n}\sum_{j=1}^{p}\left[(1-\varepsilon^{|w_{j} + u_{j}/\sqrt{n}|}) - (1-\varepsilon^{|w_{j}|}) \right] \] and \[ V_{n}^{\mathrm{atan}}(\mathbf{u}) = \sum_{i=1}^{n}\left[(e_{i} - \mathbf{u}^{T}\bm{x}_{i}/\sqrt{n})^{2} - e_{i}^{2} \right] + \frac{2\lambda_{n}}{\pi}\sum_{j=1}^{p}\left[\mathrm{arctan}(\gamma|w_{j} + u_{j}/\sqrt{n}|) - \mathrm{arctan}(\gamma|w_{j}|) \right]. \]

 Using Tayor series expansion,
\begin{align}  q_{n}^{(1)}(u_{j}) \equiv \lambda_{n}&\left[\varepsilon^{|w_{j}|}-\varepsilon^{|w_{j} + u_{j}/\sqrt{n}|} \right]  = \begin{cases}
    -\left(\lambda_{n}u_{j}/\sqrt{n}\right)\mathrm{sign}(w_{j})\varepsilon^{|w_{j}|}\log\varepsilon + o(1), & \text{if $w_{j} \neq 0$},\\
    -\left(\lambda_{n}|u_{j}|/\sqrt{n}\right)\log\varepsilon + o(1), & \text{if $w_{j} = 0$} \nonumber \\
  \end{cases}\\ &\rightarrow \lambda_{0}(-\log \varepsilon)\left[u_{j}\mathrm{sign}(w_{j})\varepsilon^{|w_{j}|}I(w_{j} \neq 0) + |u_{j}|I(w_{j} = 0) \right] \label{Prop2.eq1}  \end{align}
and
\begin{align} q_{n}^{(2)}(u_{j}) \equiv \frac{2\lambda_{n}}{\pi} &\left[\mathrm{arctan}(\gamma|w_{j} + u_{j}/\sqrt{n}|) - \mathrm{arctan}(\gamma|w_{j}|) \right] \nonumber \\ &= \begin{cases} \left( 2\lambda_{n}u_{j}\gamma/\sqrt{n}\pi \right) \frac{\mathrm{sign}(w_{j})}{1 + \gamma^{2}w_{j}^{2}} + o(1), & \text{if $w_{j} \neq 0$},\\ \left(2\lambda_{n}|u_{j}|\gamma/\sqrt{n}\pi \right) + o(1), & \text{if $w_{j} = 0$} \nonumber \\
  \end{cases}\\ &\rightarrow \frac{2\gamma\lambda_{0}}{\pi} \left[\frac{u_{j}\mathrm{sign}(w_{j})}{1 + \gamma^{2}w_{j}^{2}}I(w_{j} \neq 0) + |u_{j}|I(w_{j} = 0) \right] \label{Prop2.eq2} \end{align}

Note that the above convergence holds (both in (\ref{Prop2.eq1}) and (\ref{Prop2.eq2})) because of the assumption $\lambda_{n}/\sqrt{n} \rightarrow \lambda_{0}$. On the other hand, if we had assumed $\lambda_{n} = O(n^{\kappa/2}) = o(\sqrt{n}), \; 0 < \kappa < 1$, as in Th-KF-3, then the terms would have converged to zero thereby making the estimates asymptotically same as OLS estimates obtained without penalty. The above pointwise convergence (in (\ref{Prop2.eq1}) and (\ref{Prop2.eq2})), continuity of $q_{n}^{(i)}$ and the fact that $q_{n}^{(i)} \geq q_{n+1}^{(i)}$ for $i = 1,2,$ implies that convergence is uniform over $\mathbf{u}$ in compact sets (see e.g. Theorem 7.13 in. p.150 in Rudin, 1976). It follows that \[ V_{n}^{\mathrm{Lapl.}}(\cdot) \rightarrow_{d} V^{\mathrm{Lapl.}}(\cdot)\; \;\;\; \mathrm{and}\;\;\;\; V_{n}^{\mathrm{atan}}(\cdot) \rightarrow_{d} V^{\mathrm{atan}}(\cdot) \] on the space of functions topologized by uniform convergence on compact sets. Following (Knight and Fu, 2000 and Kim and Pollard, 1990), in order to prove that $\mathrm{argmin}(V_{n}^{\mathrm{Lapl.}}) \rightarrow_{d} \mathrm{argmin}(V^{\mathrm{Lapl.}})$ and $\mathrm{argmin}(V_{n}^{\mathrm{atan}}) \rightarrow_{d} \mathrm{argmin}(V^{\mathrm{atan}})$, it suffices to show that $\mathrm{argmin}(V_{n}^{\mathrm{Lapl.}}) = O_{p}(1)$ and $\mathrm{argmin}(V_{n}^{\mathrm{atan}}) = O_{p}(1)$. Based on Markov's lemma, in turn, it suffices to show that $\mathbb{E}(V_{n}^{\mathrm{Lapl.}})$, $\mathrm{Var}(V_{n}^{\mathrm{Lapl.}})$, $\mathbb{E}(V_{n}^{\mathrm{atan}})$, and $\mathrm{Var}(V_{n}^{\mathrm{atan}})$ are all $O(1)$ (see e.g. remark at the end of Example 2.6 in van der Vaart, 1998, p.10).

\begin{align} \mathbb{E}(V_{n}^{\mathrm{Lapl.}}) &= \frac{1}{n}\sum_{i=1}^{n}\mathbf{u}^{T}(\bm{x}_{i}\bm{x}_{i}^{T})\mathbf{u}
- \lambda_{n}\sum_{j=1}^{p}\left[\varepsilon^{|w_{j} + u_{j}/\sqrt{n}|} - \varepsilon^{|w_{j}|} \right] \nonumber \\ &\rightarrow \mathbb{E}(V^{\mathrm{Lapl.}}), \;\mathrm{a\,constant.}\;\; (\mathrm{i.e.}\; \mathbb{E}(V_{n}^{\mathrm{Lapl.}}) = O(1)). \nonumber \\ \mathrm{Var}(V_{n}^{\mathrm{Lapl.}}) &= \frac{4}{n}\sum_{i=1}^{n}\mathbf{u}^{T}(\bm{x}_{i}\bm{x}_{i}^{T})\mathbf{u} \nonumber \\ &\rightarrow 4\mathbf{u}^{T}C\mathbf{u} = \mathrm{Var}(V^{\mathrm{Lapl.}})\;\; (\mathrm{i.e.}\; \mathrm{Var}(V_{n}^{\mathrm{Lapl.}}) = O(1)). \end{align} Similarly, it is easy to show that \[\mathbb{E}(V_{n}^{\mathrm{atan}}) \rightarrow \mathbb{E}(V^{\mathrm{atan}}), \;\mathrm{a\;constant}\;\mathrm{and}\;\mathrm{Var}(V_{n}^{\mathrm{atan}}) \rightarrow \mathrm{Var}(V^{\mathrm{atan}}), \;\mathrm{a\,constant}. \]

\hfill\BlackBox

Note that although the proof for Proposition-2 followed closely the proof of Th-KF-3 (nonconvex Bridge penalty case), the statement of the result (that is, the conclusion) has some similarities rather to Th-KF-2 (Lasso case). More precisely, the result obtained in Proposition-2 lies somewhere in between those obtained in Th-KF-2 and Th-KF-3. First of all, as mentioned within the proof above, $\lambda_{n} = O(n^{\kappa/2})$, $0 < \kappa < 1$ (as in Th-KF-3; nonconvex Bridge case) would not have given a meaningful result and hence $\lambda_{n} = O(n^{1/2})$ (as in Th-KF-2; Lasso case) was required.

Second of all, Th-KF-3 showed the nonconvex Bridge penalized estimators of non-zero regression parameters to be unbiased asymptotically. However, Proposition-2 above shows some asymptotic bias for estimators of non-zero regression weights based on Laplace or arctan penalization, similar to the case of Lasso penalty presented in Th-KF-2. However the asymptotic biases in Laplace and arctan cases are much smaller than the asymptotic bias obtained for Lasso in Th-KF-2, especially for large $|w_{j}|$. Note that the asymptotic bias for Laplace case is that for the Lasso case multiplied by a factor of $(-\log \varepsilon)\varepsilon^{|w_{j}|}$, for each $j$; similarly, the asymptotic bias for arctan is multiplied by a factor of $2\gamma/(\pi(1 + \gamma^{2}w_{j}^{2})$. Both multiplication factors decrease (at rates $O(e^{|w_{j}|})$ and $O(w_{j}^{2})$, respectively) as $w_{j} \rightarrow \infty$. 

To get a better perspective numerically, for $\varepsilon = 0.0001$, we have $(-\log \varepsilon)\varepsilon^{|w_{j}|}) < 1$ (so that the asymptotic bias is less than that of Lasso), when $|w_{j}| = 0.242$, and we have $(-\log \varepsilon)\varepsilon^{|w_{j}|}) > 1$, when $|w_{j}| = 0.241$. Thus, ``large" $|w_{j}|$ in this case is a value of $0.242$ which for most regression analyses stemming from real-world scenarios, is really a small value. For the arctan case, $2\gamma/\pi(1 + \gamma^{2}w_{j}^{2})$ is less than 1 (i.e. asymptotic bias less than that of Lasso), whenever $w_{j}^{2} > (2\gamma - \pi)/\gamma^{2}\pi$; the RHS of the inequality is maximized when $\gamma = \pi$, with maximum value $2/\pi^{2} \approx 0.203$ (note $\sqrt{2}/\pi \approx 0.45$). When $\gamma = 1$ for example, the asymptotic bias is roughly only half that of Lasso, even with $|w_{j}|$ as small as $0.5$. With the same value of $|w_{j}|$, the asymptotic bias of arctan is roughly $0.025$ as that for Lasso when $\gamma = 100$.

Although Proposition-2 provides interesting insights about Laplace and arctan penalties in comparison to Lasso it does not shed a very flattering light onto these penalties because of the respective asymptotic biases mentioned in the statement. One limitation of the proposition, though, is that it lets $n \rightarrow \infty$ with $p$ fixed. A more realistic setting is perhaps letting $p$ depend on $n$ and letting it grow along with $n$ as well. In this regard, more pertinent results are given in the following Proposition-3; this proposition is very similar to Theorem 1 in Huang, Horowitz and Ma, 2008, for Bridge estimators. For convenience we will use the short-hand HHM2008 to refer to Huang, Horowitz and Ma, 2008.

We introduce some new notation first. Suppose that the true regression weight value $\bm{w}_{0}$ has $k_{n}$ number of non-zero coefficients and $m_{n}$ number of zero coefficients and can be written as $\bm{w}_{0} = (\bm{w}_{10}^{T}, \bm{w}_{20}^{T})^{T}$ where $\bm{w}_{10} (\neq \bm{0})$ is a $k_{n} \times 1$ vector and $\bm{w}_{20} (= \bm{0})$ is a $m_{n} \times 1$ vector. We assume that the $y_{i}$'s are centered and that the covariates are standardized. We also write $\bm{x_{i}} = (\bm{x_{i}}^{(1)^T}, \bm{x_{i}}^{(2)^T})$, where $\bm{x_{i}}^{(1)}$ consists of the first $k_{n}$ covariates corresponding to the non-zero coefficients. Let $C_{1n} = \frac{1}{n}\sum_{i=1}^{n}\bm{x_{i}}^{(1)}\bm{x_{i}}^{(1)T}$ and let $\rho_{1n}$ and $\rho_{2n}$ be the smallest and largest eigenvalues of $C_{n}$ and let $\tau_{1n}$ and $\tau_{2n}$ be the smallest and largest eigenvalues of $C_{1n}$, respectively. Finally, let us denote by $\mathcal{L}_{n}^{Lapl.}$ and $\mathcal{L}_{n}^{atan}$ the least squares loss function penalized by Laplace and arctan penalties, respectively (that is, the objective functions in (\ref{OLS.Lapl}) and (\ref{OLS.atan})).

As in HHM2008, we need the following assumptions (HHM$_{1}$-HHM$_{3}$) for the statement and proof of Proposition-3; note that we do not need all the assumptions listed in their paper.

(HHM$_{1}$) $e_{1}, e_{2}, \ldots$ are independent and identically distributed random variables with mean zero and variance $\sigma^{2}$.\\
(HHM$_{2}$) (a) $\rho_{1n} > 0$ for all $n$; (b) $(p_{n} + \lambda_{n}k_{n})(n\rho_{1n})^{-1} \rightarrow 0.$\\
(HHM$_{3}$) $\lambda_{n}(k_{n}/n)^{1/2} \rightarrow 0$. 

We get a new consistency result (Proposition-3) if the assumptions (HHM$_{1}$-HHM$_{3}$) hold. Note that, unlike Proposition-1, Proposition-3 gives also rates of convergence based on $\rho_{1n}, p_{n}, \lambda_{n}, k_{n}$ and $n$.

\noindent \textbf{\underline{Proposition-3}}: Let $\widehat{\bm{w}}_{n}^{\mathrm{Lapl.}}$ and $\widehat{\bm{w}}_{n}^{\mathrm{atan}}$ denote the minimizers of $\mathcal{L}_{n}^{Lapl.}$ and $\mathcal{L}_{n}^{atan}$ respectively. Suppose that conditions (HHM$_{1}$-HHM$_{3}$) hold. Let $h_{n} = \rho_{1n}^{-1}(p_{n}/n)^{1/2}$ and $h_{n}^{'} = [(p_{n} + \lambda_{n}k_{n})/(n\rho_{1n})]^{1/2}$. Then $||\widehat{\bm{w}}_{n}^{\mathrm{Lapl.}} - \bm{w}_{0}|| = O_{p}(\min \{h_{n}, h_{n}^{'}\})$ and $||\widehat{\bm{w}}_{n}^{\mathrm{atan}} - \bm{w}_{0}|| = O_{p}(\min \{h_{n}, h_{n}^{'}\})$.

The proof of the above proposition requires a lemma from HHM2008, which we state below.

\noindent \textbf{\underline{Lemma 4.1}}: Let $\mathbf{u}$ be a $p_{n} \times 1$ vector. Under condition (HHM$_{1}$), \[ \mathbb{E} \sup_{||u|| < \delta} \abs{\sum_{i=1}^{n} e_{i}\bm{x_{i}}^{T}\mathbf{u} } \leq \delta\sigma n^{1/2}p_{n}^{1/2}. \]

\noindent \textbf{Proof of Proposition-3}: The proof follows very closely the proof of Theorem 1 in HHM2008, and hence we give only a sketch of the proof, highlighting only the steps which are different from their theorem. First task is to show that \begin{equation} ||\widehat{\bm{w}}_{n}^{\mathrm{Lapl.}} - \bm{w}_{0}|| = O_{p}(p_{n} + \lambda_{n}k_{n})/(n\rho_{1n})^{1/2})\; \mathrm{and}\; ||\widehat{\bm{w}}_{n}^{\mathrm{atan}} - \bm{w}_{0}|| = O_{p}(p_{n} + \lambda_{n}k_{n})/(n\rho_{1n})^{1/2}). \label{Prop3.eq1} \end{equation}
Let $\eta_{n}^{Lapl.} = \lambda_{n}\sum_{i=1}^{p_{n}}\left(1 - \varepsilon^{|w_{0j}|} \right)$ and $\eta_{n}^{atan} = \lambda_{n}\sum_{i=1}^{p_{n}}\left(\frac{2}{\pi}\mathrm{arctan}(\gamma|w_{0j}|) \right)$. Then proceeding exactly as in the first few steps in HHM2008's proof, we will get \begin{align} n\mathbb{E}[(\widehat{\bm{w}}_{n}^{\mathrm{Lapl.}} - \bm{w}_{0})^{T}C_{n}(\widehat{\bm{w}}_{n}^{\mathrm{Lapl.}} - \bm{w}_{0})] &\leq 6\sigma^{2}p_{n} + 3\eta_{n}^{Lapl.}  \label{Prop3.eq2} \\ n\mathbb{E}[(\widehat{\bm{w}}_{n}^{\mathrm{atan}} - \bm{w}_{0})^{T}C_{n}(\widehat{\bm{w}}_{n}^{\mathrm{atan}} - \bm{w}_{0})] &\leq 6\sigma^{2}p_{n} + 3\eta_{n}^{atan}.  \label{Prop3.eq3}  \end{align} Since the number of nonzero coefficients is $k_{n}$, both $\eta_{n}^{Lapl.}$ and $\eta_{n}^{atan}$are $O(\lambda_{n}k_{n})$. Noting that $\rho_{1n}$ is the smallest eigenvalue of $C_{n}$, (\ref{Prop3.eq1}) follows from (\ref{Prop3.eq2}) and (\ref{Prop3.eq3}).

Next we show that \begin{equation} ||\widehat{\bm{w}}_{n}^{\mathrm{Lapl.}} - \bm{w}_{0}|| = O_{p}(\rho_{1n}^{-1}(p_{n}/n)^{1/2})\; \mathrm{and}\; ||\widehat{\bm{w}}_{n}^{\mathrm{atan}} - \bm{w}_{0}|| = O_{p}(\rho_{1n}^{-1}(p_{n}/n)^{1/2}). \label{Prop3.eq4} \end{equation} As in HHM2008, we follow proof of Theorem 3.2.5 of van der Vaart and Wellner (1996) to prove (\ref{Prop3.eq4}). The idea is to partition the parameter space (minus $\bm{w}_{0}$) into ``shells'' $S_{j,n} = \{\bm{w}: 2^{j-1} < r_{n}||\bm{w} - \bm{w}_{0}|| < 2^{j}\}$, for each $n$, with $j$ ranging over the integers; here $r_{n} = \rho_{1n}(n/p_{n})^{1/2}$. Using the definition of $\widehat{\bm{w}}_{n}^{\mathrm{Lapl.}}$ that it minimizes $\mathcal{L}_{n}^{Lapl.}$ and the fact that for a given integer $M$, $\{\widehat{\bm{w}}_{n}^{\mathrm{Lapl.}}: r_{n}||\widehat{\bm{w}}_{n}^{\mathrm{Lapl.}} - \bm{w}_{0}|| > 2^{M} \} \subset \cup_{j \geq M}S_{j,n}$, we get \begin{align} \mathrm{P}(r_{n}&||\widehat{\bm{w}}_{n}^{\mathrm{Lapl.}} - \bm{w}_{0}|| > 2^{M}) \nonumber \\ &= \sum_{j \geq M, 2^{j} \leq t r_{n}}\mathrm{P}\left(\inf_{\bm{w}\in S_{j,n}}(\mathcal{L}_{n}^{\mathrm{Lapl.}}(\bm{w}) - \mathcal{L}_{n}^{\mathrm{Lapl.}}(\bm{w_{0}})) \geq 0 \right) + \mathrm{P}(2||\widehat{\bm{w}}_{n}^{\mathrm{Lapl.}} - \bm{w}_{0}|| \geq t), \label{Prop3.eq5} \end{align} for every $t > 0$. Similarly, we get, \begin{align} \mathrm{P}(r_{n}&||\widehat{\bm{w}}_{n}^{\mathrm{atan}} - \bm{w}_{0}|| > 2^{M}) \nonumber \\ &= \sum_{j \geq M, 2^{j} \leq t r_{n}}\mathrm{P}\left(\inf_{\bm{w}\in S_{j,n}}(\mathcal{L}_{n}^{\mathrm{atan}}(\bm{w}) - \mathcal{L}_{n}^{\mathrm{atan}}(\bm{w_{0}})) \geq 0 \right) + \mathrm{P}(2||\widehat{\bm{w}}_{n}^{\mathrm{atan}} - \bm{w}_{0}|| \geq t), \label{Prop3.eq6} \end{align} Since we assume (HHM$_{2}$), (\ref{Prop3.eq1}) implies that $\widehat{\bm{w}}_{n}^{\mathrm{Lapl.}}$ and $\widehat{\bm{w}}_{n}^{\mathrm{atan}}$ are consistent and hence the second terms in the right hand sides of (\ref{Prop3.eq5}) and (\ref{Prop3.eq6}) converge to zero. Now \begin{align} \mathcal{L}_{n}^{\mathrm{Lapl.}}(\bm{w}) - \mathcal{L}_{n}^{\mathrm{Lapl.}}(\bm{w_{0}}) &\geq \sum_{i=1}^{n}[\bm{x}_{i}^{T}(\widehat{\bm{w}}_{n}^{\mathrm{Lapl.}} - \bm{w}_{0})]^{2} - 2\sum_{i=1}^{n}e_{i}\bm{x}_{i}^{T}(\widehat{\bm{w}}_{n}^{\mathrm{Lapl.}} - \bm{w}_{0}) \nonumber \\ &+ \lambda_{n}\sum_{j=1}^{k_{n}}\{\varepsilon^{|w_{0,1j}|} -  \varepsilon^{|w_{1j}|} \} \equiv I_{1n}^{\mathrm{Lapl.}} + I_{2n}^{\mathrm{Lapl.}} + I_{3n}^{\mathrm{Lapl.}}, \nonumber \end{align} and \begin{align} \mathcal{L}_{n}^{\mathrm{atan}}(\bm{w}) - \mathcal{L}_{n}^{\mathrm{atan}}(\bm{w_{0}}) &\geq \sum_{i=1}^{n}[\bm{x}_{i}^{T}(\widehat{\bm{w}}_{n}^{\mathrm{atan}} - \bm{w}_{0})]^{2} - 2\sum_{i=1}^{n}e_{i}\bm{x}_{i}^{T}(\widehat{\bm{w}}_{n}^{\mathrm{atan}} - \bm{w}_{0}) \nonumber \\ &+ \frac{2\lambda_{n}}{\pi}\sum_{j=1}^{k_{n}}\{\mathrm{arctan}(\gamma|w_{1j}|) - \mathrm{arctan}(\gamma|w_{0,1j}|) \} \equiv I_{1n}^{\mathrm{atan}} + I_{2n}^{\mathrm{atan}} + I_{3n}^{\mathrm{atan}}. \nonumber \end{align} On $S_{j,n}$ both the first terms $I_{1n}^{\mathrm{Lapl.}}$ and $I_{1n}^{\mathrm{atan}}$ can be lower-bounded by $n\rho_{1n}2^{2(j-1)}r_{n}^{-2}$. The third term in the Laplace case, \[ I_{3n}^{\mathrm{Lapl.}} = -\lambda_{n}\sum_{j=1}^{k_{n}}\left[\varepsilon^{|w_{1j}|} - \varepsilon^{|w_{0,1j}|} \right] = \lambda_{n}(-\log \varepsilon)\sum_{j=1}^{k_{n}}\mathrm{sign}(w_{0,1j})(w_{1j} - w_{0,1j})\varepsilon^{|w_{0,1j}^{*}|} \nonumber \] for some $w_{0,1j}^{*}$ and $w_{1j}$, so that \begin{align} \abs{I_{3n}^{\mathrm{Lapl.}}} & \leq \lambda_{n}(-\log \varepsilon) \sum_{j=1}^{k_{n}}|w_{1j} - w_{0,1j}|, \; \mathrm{since}\; 0 < \varepsilon < 1 \nonumber \\ &\leq \lambda_{n}(-\log \varepsilon)k_{n}^{1/2}||\bm{w} - \bm{w_{0}}||, \;\mathrm{by}\;\mathrm{Cauchy-}\mathrm{Schwartz\; inequality}. \label{Prop3.eq7} \end{align}
Similarly, in the arctan case, \begin{align} \abs{I_{3n}^{\mathrm{atan}}} &= \frac{2\lambda_{n}}{\pi}\sum_{j=1}^{k_{n}}[ \arctan (\gamma|w_{1j}|) - \arctan (\gamma|w_{0,1j}|)] \nonumber \\ &= \frac{2\lambda_{n}\gamma}{\pi} \sum_{j=1}^{k_{n}} \left[\frac{\mathrm{sign}(w_{0,1j})(w_{1j} - w_{0,1j})}{1 + \gamma^{2}w_{0,1j}^{*2}} \right] \nonumber \end{align} so that \begin{equation} \abs{I_{3n}^{\mathrm{atan}}} \leq \frac{2\gamma\lambda_{n}}{\pi}\left(\sum_{j=1}^{k_{n}}|w_{1j} - w_{0,1j}| \right) \leq \frac{2\gamma\lambda_{n}k_{n}^{1/2}}{\pi}\lVert\bm{w} - \bm{w_{0}}\rVert. \label{Prop3.eq8} \end{equation} The rest of the proof follows as in HHM2008. Note that assumption HHM$_{3}$ and Lemma 4.1 are required to complete the proof as in HHM2008. \hfill\BlackBox

The key difference of our proof above from the proof in HHM2008, is that they needed an upper-bound assumption on the regression weights (assumption (A4) in their paper) to get inequalties similar to (\ref{Prop3.eq7}) or (\ref{Prop3.eq8}). In the Laplace and arctan cases we do not need such an upper bound assumption on the regression weights. This subtle difference is actually conceptually a major difference and speaks to the actual difference between nonconvex Bridge penalties on one hand, and the Laplace and arctan penalties on the other hand, as approximations to the $L_{0}$ norm (see next subsection). In the nonconvex Bridge case, since the penalty function grows to infinity (that is, unbounded), a not-so-realistic boundedness assumption on the regression weights is required. In the Laplace and arctan cases, the penalty approaches 1 from below (and hence bounded). Thus a boundedness assumption on the regression weights is not necessary to prove our Proposition-3, conclusions of which are still same as that of Theorem 1 in HHM2008.

\subsection*{Convergence of the Laplace and arctan approximation functions}

Here we present heuristic justifications for using the Laplace or arctan penalties over the Bridge penalties by considering their respective error in approximating the indicator function involved in the canonical selection procedure (\ref{B2}). \\

\textbf{Lemma 4.2.} Consider the approximation functions for $I(x \neq 0)$, $f(x) = \abs{x}^{\varepsilon}$ and $g(x) = 1 - \varepsilon^{\abs{x}}$ for some fixed $\varepsilon \in (0,1)$. The overall error for $f(x)$ is much larger than that of $g(x)$.

\textbf{Proof:}  We give a proof based on heuristic analysis. Because of symmetry we just focus on the right side of origin on the x-axis for error analysis. For an interval $[a, a+h]$, with $a > 1$, $h > 0$, we have $f(x) = x^{\varepsilon}$, so that the area under the curve in this interval is \[\displaystyle \int_{a}^{a+h} f(x) dx = \frac{ (a+h)^{\varepsilon + 1} - a^{\varepsilon + 1} }{\varepsilon + 1}. \] The area under the curve for the indicator function $I(x \neq 0)$ in this interval is $h$, so that the error in approximation is \[ \frac{(a+h)^{\varepsilon + 1} - a^{\varepsilon + 1} }{\varepsilon + 1} - h \approx \frac{ (\varepsilon + 1)a^{\varepsilon}}{ (\varepsilon + 1) } h - h = (a^{\varepsilon} - 1)h, \] where we used the leading term of the Taylor series approximation to the function $(a+x)^{\varepsilon + 1}$. For the approximation function $g(x)$ the area under the curve in the interval $[a, a+h]$ with $a > 1$ is \[\displaystyle \int_{a}^{a+h} g(x) dx = h - \frac{\varepsilon^{a}\left(\varepsilon^{h} -1 \right)}{\mathrm{log} (\varepsilon) }. \] Using the leading term in the Taylor series approximation, $\varepsilon^{h} - 1$ can be approximated by $\mathrm{log} (\varepsilon) h$ so that the absolute value of the error is approximately $\varepsilon^{a}h$. Thus the absolute value of the error for $f(x)$ in a unit interval $[a, a+1]$ (with $a > 1$) is approximately $a^{\varepsilon} - 1$ and that for $g(x)$ in the same interval is approximately $\varepsilon^{a}$. For a fixed $\varepsilon$, the former can be made arbitrarily large, and the latter arbitrarily small by increasing $a$. The error for $g(x)$ is larger than that for $f(x)$ in the interval $[0,1]$ but the difference in this interval is bounded. \hfill\BlackBox

\textbf{Lemma 4.3.} Consider the approximation functions for $I(x \neq 0)$, $f(x) = \abs{x}^{\varepsilon}$ and $g(x) = (2/\pi)\mathrm{arctan}(\gamma x)$ where $\gamma = \varepsilon^{-1}, \; \varepsilon \in (0,1)$ is fixed. The overall error for $f(x)$ is much larger than that of $g(x)$.

\textbf{Proof:} In this case \begin{align} \displaystyle \int_{a}^{a+h} g(x) dx &= \frac{2}{\pi} \left[ (a+h)\mathrm{arctan}(\gamma(a+h)) - a\,\mathrm{arctan}(a\gamma)\right] \nonumber \\ &\qquad - \frac{1}{\pi\gamma} \left[\log (1 + \gamma^{2}(a+h)^{2}) - \log(1 + \gamma^{2}a^{2}) \right] \nonumber \\ &\approx \frac{2h}{\pi}\left[\mathrm{arctan}(\gamma a) + \frac{\gamma a}{1 + \gamma^{2}a^{2}} \right] - \frac{2h}{\pi}\left[\frac{\gamma a}{1 + \gamma^{2}a^{2}} \right] \nonumber \\ &= \frac{2h}{\pi}\left[\mathrm{arctan}(\gamma a) \right], \nonumber \end{align} where the approximate equality above was obtained using the leading term in the Taylor series expansion of the each of the following functions: \[ f_{1}(x) = (a+x)\mathrm{arctan}[\gamma(a+x)] \; \mathrm{and} \; f_{2}(x) = \frac{1}{2\gamma} \log [1 + \gamma^{2}(a+x)^{2}]. \] Thus the absolute value of the error for $g(x)$ in a unit interval $[a, a+1]$ (with $a > 1$) is approximately \[ \abs{1 - \frac{2}{\pi}\mathrm{arctan}(\gamma a)} \] which can be made arbitrarily small by increasing $a$, since for $x \geq 0$, $\mathrm{arctan}(x) < \pi/2$ and increases to $\pi/2$ as $x \rightarrow \infty$. On the other hand, as shown in the previous lemma, the corresponding error for $f(x)$ can be made arbitrarily large by increasing $a$. Also the error for $g(x)$ and $f(x)$ in the interval $[0,1]$ is bounded. \hfill\BlackBox

\section{Optimization considerations}

Until recently, statistical learning approaches mostly focused on convex optimization primarily because of the many efficient solution methods available in that arena. With the advent of nonconvex and/or nondifferentiable objective functions in statistical learning problems, recently, a lot more attention has been focused on optimization considerations suitable for such problems.

The key inequality typically used in establishing the convergence rate of gradient descent or stochastic gradient algorithms for optimization problems with a strongly convex objective function $F$ is \begin{equation} \langle \nabla F(\mathbf{y}) - \nabla F(\mathbf{x}), \mathbf{y}-\mathbf{x} \rangle \geq l||\mathbf{y}-\mathbf{x}||^{2} \label{OC1} \end{equation} for some positive real number $l$. If $\bm{w}^{*}$ is a minimum of $F$ (so that $\nabla F(\mathbf{\bm{w}}^{*})$ = 0), then (\ref{OC1}) can be re-written as \begin{equation} \langle \nabla F(\bm{w}^{*} + \Delta), \Delta \rangle \geq l||\Delta||^{2}.\label{OC2} \end{equation} If we just assume convexity of $F$, the corresponding key inequality used in the proofs for establishing rates of convergence is \begin{equation} \langle \nabla F(\bm{w}^{*} + \Delta), \Delta) \geq F(\bm{w}^{*} + \Delta) - F(\bm{w}^{*}). \label{OC3}\end{equation} Note that inequalities in (\ref{OC2}) and (\ref{OC3}) are sometimes used as definitions of strong convexity and convexity, respectively. For algorithms that allows non-differentiability of $F$ but still assumes convexity (e.g FOBOS presented in Duchi and Singer, 2009), the gradient in (\ref{OC3}) is replaced with the subgradient.

Since the objective functions that we consider are nonconvex, we cannot apply either of the inequalities ((\ref{OC2}) or (\ref{OC3})) above. We will have to consider alternate optimization frameworks and alternate inequalities to study convergence of algorithms. We briefly review below two types of optimization frameworks, based on composite gradient descent and difference of convex approaches, that have appeared in the literature which could be utilized for optimization with nonconvex penalties considered in this paper. First we present two inequalities (mentioned in LW15 and Agarwal, Negahban and Wainwright, 2012) that will play a role in the discussion below.

Loh and Wainwright (LW15) defined restricted strong convexity (RSC) condition for an empirical loss function $\mathcal{L}_{n}$ as \begin{equation} \langle \nabla \mathcal{L}_{n}(\bm{w}^{*} + \Delta) - \nabla \mathcal{L}_{n}(\bm{w}^{*}), \Delta \rangle \geq \begin{cases} \alpha_{1}||\Delta||_{2}^{2} - \tau_{1} \frac{\log p}{n}||\Delta||_{1}^{2}, & \forall ||\Delta||_{2} \leq 1, \\ \alpha_{2}||\Delta||_{2} - \tau_{2}\sqrt{\frac{\log p}{n}}||\Delta||_{1}, & \forall ||\Delta||_{2} \geq 1, \end{cases} \label{RSC1} \end{equation} where $\alpha_{j} > 0$, $\tau_{j} \geq 0,\; j = 1,2, 3$ (including $\alpha_{3}$ and $\tau_{3}$ to be defined below). There are many examples of nonconvex empirical loss functions that satisfy RSC (see LW15 and also Loh and Wainwright, 2012). For a differentiable $\mathcal{L}_{n}$ the Taylor series error around $\bm{w}'$ is defined as \[ \mathbb{TE}(\bm{w}, \bm{w}') = \mathcal{L}_{n}(\bm{w}) - \mathcal{L}_{n}(\bm{w}') - \langle \nabla \mathcal{L}_{n}(\bm{w}'), \bm{w} - \bm{w}' \rangle. \] Restricted smoothness condition (RSM) defined in previous literature is \[ \mathbb{TE}(\bm{w}, \bm{w}') \leq \alpha_{3}||\bm{w} - \bm{w}' ||_{2}^{2} + \tau_{3} \frac{\log p}{n} ||\bm{w} - \bm{w}'||_{1}^{2}, \; \forall \bm{w}, \bm{w}' \in \mathbb{R}^{p}.\] As shown in Agarwal, Negahban and Wainwright, 2012, the loss function involved in the maximum likelihood estimation of sparse log-linear models with $l_{1}$ constraint is never strongly convex; however, a broad class of such models satisfy both the RSC and RSM conditions.

\subsection{Composite gradient descent (CGD)}

CGD framework was presented in Nesterov, 2007, for optimizing a function which can be written as the sum of two functions, especially for the case where one of the functions in the sum is non-differentiable. The key idea in Nesterov's original formulation of the CGD approach is to approximate the objective function with a composite gradient mapping which replaces the differentiable function with a quadratic approximation of it. The version of CGD algorithm that we consider here is the one described in section 4 in LW15 for an optimization of the form (1); here, the objective function is the sum of two functions, with one of them (i.e. the penalty function) nonconvex, and non-smooth at zero. The key idea in LW15's adaptation of CGD is to add and subtract the term ($\mu/2)||w||_{2}^{2}$ to the functions in (1) to get a new objective function \begin{equation} \left(\mathcal{L}(w) - \frac{c_{\theta}\mu}{2}||w||_{2}^{2} \right) + Q_{\theta}(w),\;\mathrm{where}\; Q_{\theta}(w) = P_{\theta}(w) + \frac{c_{\theta}\mu}{2}||w||_{2}^{2}. \end{equation} $c_{\theta}$ is a constant depending on the parameter vector $\theta$. If we choose $\mu$ same as in property (P5), then $Q_{\theta}(w)$ is a convex penalty function but possibly non-smooth. $\mathcal{L}(w) - \frac{c_{\theta}\mu}{2}||w||_{2}^{2}$ is a differentiable but nonconvex function.

If the penalty function satisfies properties (P1)-(P5) listed in section 3 and if the empirical loss function $\mathcal{L}_{n}$ satisfy the RSM condition and a slightly modified version of the RSC condition (where the RHS in the inequality in (\ref{RSC1}) is modified to a uniform lower bound for all $\bm{w}^{*}$ in a particular set), then Theorem 3 in LW15 guarantees that the optimization error $||\bm{w}^{t}-\bm{w}^{*}||$ will fall below a tolerance level $\delta^{2}$ within $T \asymp \log (1/\delta^{2})/ \log (1/\nu)$ iterations, assuming that the sample size has a particular scaling property (see LW15 for more specific details). Here $t$ represents the iteration counter, $\nu$ is a contraction factor with values in ($0,1$) based on the scaling assumption of sample size mentioned above, and $\delta^{2}$ is of the same order as that of the squared statistical error $\epsilon_{\mathrm{stat}}^{2}$, the distance between a fixed global optimum and the target parameter.

\subsection{Difference-of-convex (DC) learning}

Another framework for nonconvex optimization problems which is more general and less algorithm-specific is the difference-of-convex (DC) programming setting, where the key idea is to write a nonconvex function as the difference of two convex functions. Deterministic (Pham and Souad, 1986) and stochastic DC algorithms (Hoai \textit{et. al.}, 2019) and their convergence analyses are well-established in optimization literature by now; see Le Thi and Pham, 2018, for a survey. Recently, Ahn, Pang and Xin, 2017 studied regularized optimization problems as given in (1) (i.e. bicriteria optimization problems), by exploiting some special structure associated with such problems. (We will be using the short-hand notation APX17 hereafter for Ahn, Pang and Xin, 2017.) Specifically, APX17 considered the cases where a penalty function $P(w)$ can be written as a DC function. \[ P(w) = g(w) - h(w), \;\mathrm{with}\; h(w) = \max_{1 \leq i \leq I} h_{i}(w), \; \mathrm{for}\;\mathrm{some}\;\mathrm{integer}\; I > 0, \] where $g$ is convex but not necessarily differentiable and each $h_{i}$ is convex and differentiable. Based on the above structure for $h$ and the definition of a piecewise smooth function, it follows that $h$ is a piecewise differentiable function. For co-ordinate separate nonconvex penalty functions considered in this paper, each component $p_{\theta}(\cdot)$ can be written as a univariate analog of the DC function above. The cases for SCAD and MCP are illustrated in APX17. For $x \in \mathbb{R}$, in the case of Laplace penalty, \[p_{\varepsilon}(x) = g_{\varepsilon}(x) - h_{\varepsilon}(x), \; \mathrm{where}\;g_{\varepsilon}(x) = (-\log \varepsilon )|x|\;\mathrm{and}\;h_{\varepsilon}(x) = (-\log \varepsilon )|x| - \left(1 - \varepsilon^{|x|} \right) \] and in the case of arctan penalty \[p_{\gamma}(x) = (2/\pi)(g_{\gamma}(x) - h_{\gamma}(x)), \; \mathrm{where}\;g_{\gamma}(x) = \gamma|x|\;\mathrm{and}\;h_{\gamma}(x) = \gamma|x| - \arctan(\gamma |x|). \]

APX17 focused on the d(irectional)-stationary solutions of the optimization problem because such solutions are sharpest kind among all types of stationary solutions; in other words a d-stationary solution must be stationary according to other definitions of stationarity. A few assumptions made in APX17 pertaining to our discussion are given below.

\noindent (A$_{0}$): In addition to the special structure for the penalty function that we assumed above, we assume that the loss function
 $\mathcal{L}(x)$ is a convex function on a closed convex set $\mathcal{X} \subset \mathbb{R}^{p}$.

\noindent (A$_{1}$): The loss function $\mathcal{L}$ is continuously differentiable with a Lipschitz gradient (i.e. of class LC$^{1}$) on $\mathcal{X}$.

\noindent (A$_{2}$): A nonnegative constant $\lambda_{\mathrm{min}}$ exists such that \[ \mathcal{L}(\mathbf{x}) - \mathcal{L}(\mathbf{y}) - \nabla(\mathbf{y})^{T}(\mathbf{x} - \mathbf{y}) \geq (\lambda_{\mathrm{min}}/2)||\mathbf{x} - \mathbf{y}||_{2}^{2}, \; \forall \; \mathbf{x}, \mathbf{y}\; \in\; \mathcal{X}. \]

\noindent (A$_{3}$): Nonnegative constants $L_{h}$ and $\beta$ exist such that for each $i \in \{1, \ldots, I\}$ and for all $\mathbf{x}, \mathbf{y}\; \in\; \mathcal{X}$, with $0/0$ defined to be zero, \[0 \leq h_{i}(\mathbf{x}) - h_{i}(\mathbf{y}) - \nabla h_{i}(\mathbf{y})^{T}(\mathbf{x} - \mathbf{y}) \leq \left(\frac{L_{h}}{2} + \frac{\beta}{||\mathbf{y}||_{2}} \right)||\mathbf{x} - \mathbf{y}||_{2}^{2}  \]

The constant $\lambda_{\mathrm{min}}$ appearing in (A$_{2}$) expresses the convexity of the loss function $\mathcal{L}$, and the strong convexity when $\lambda_{\mathrm{min}} >$ 0. If the convex functions $h_{i}$ appearing in (A$_{3}$) are of class LC$^{1}$, then we may take $\beta = 0$. When the assumptions (A$_{0}$-A$_{3}$) hold, in the $\beta$ = 0 case, one of the key propositions proved in APX17 (Proposition 3.1) states that a d-stationary point of $Z_{\lambda}(w) = \mathcal{L}(w) + \lambda P(w)$ on $\mathcal{X}$ is a unique minimizer of $Z_{\lambda}$ on a slightly more restrictive set $\mathcal{X}_{*} \subset \mathcal{X}$; here, $0 < \lambda \leq  \lambda_{\mathrm{min}}/L_{h}$. In order for the above-mentioned proposition from APX17 to hold for Laplace and arctan penalties, we verify below that $h_{\varepsilon}(x) = (-\log \varepsilon )|x| - (1 - \varepsilon^{|x|})$ and $h_{\gamma}(x) = \gamma|x| - \arctan(\gamma |x|)$ are indeed both of the class LC$^{1}$.

\noindent \underline{\textit{Laplace case}}: We will show the derivative of $h_{\varepsilon}$ is Lipschitz with Lipschitz constant ($\log \varepsilon$)$^{2}$.
\[ h_{\varepsilon}'(x) = (-\log \varepsilon) \mathrm{sign} (x) (1 - \varepsilon^{|x|}). \]
\noindent When $x \geq 0, y \geq 0$:
  \begin{align} |h_{\varepsilon}'(x) - h_{\varepsilon}'(y)| &= (-\log \varepsilon)|\varepsilon^{x} - \varepsilon^{y}| \nonumber \\ &= (\log \varepsilon)^{2}|(x-y)\varepsilon^{u}|, \; u \in (\min\{x,y\}, \max\{x,y\}) \nonumber \\ &\leq (\log \varepsilon)^{2}|x-y|, \; \mathrm{since}\; |\varepsilon^{u}| \leq 1. \nonumber \end{align}
\noindent When $x \geq 0, y \leq 0$ ($z = -y$, say):
  \begin{align} |h_{\varepsilon}'(x) - h_{\varepsilon}'(y)| &= |\log \varepsilon||(\varepsilon^{x} - 1) + (\varepsilon^{z} - 1)| \nonumber \\ &= |\log \varepsilon|| x(\log \varepsilon)\varepsilon^{u_{1}} + z(\log \varepsilon)\varepsilon^{u_{2}}|, \; u_{1} \in (0, x), \; u_{2} \in (0, z), \nonumber \\ &= (\log \varepsilon)^{2}\left(x\varepsilon^{u_{1}} + z\varepsilon^{u_{2}}\right) \nonumber \\ &\leq (\log \varepsilon)^{2}(x + z), \;\mathrm{since}\; x \geq 0, z \geq 0, 0 \leq \varepsilon^{u_{1}} \leq 1, 0 \leq \varepsilon^{u_{2}} \leq 1 \nonumber \\ &= (\log \varepsilon)^{2}|x + z|, \;\mathrm{since}\; x + z \geq 0 \nonumber \\ &= (\log \varepsilon)^{2}|x - y|, \;\mathrm{since}\; z = -y. \nonumber \end{align} Remaining cases $x \leq 0, y \geq 0$ and $x \leq 0, y \leq 0$ can be proved similarly.

\noindent \underline{\textit{arctan case}}: We will show the derivative of $h_{\gamma}$ is Lipschitz with Lipschitz constant $\gamma^{2}$. \[ h_{\gamma}'(x) = \gamma \mathrm{sign}(x) \left(1 - \frac{1}{1 + \gamma^{2}x^{2}}\right) = \gamma \mathrm{sign}(x) \frac{\gamma^{2}x^{2}}{1 + \gamma^{2}x^{2}}. \]
\noindent When $x \geq 0, y \geq 0$:
\[ h_{\gamma}'(x) - h_{\gamma}'(y) = \gamma \left[\frac{\gamma^{2}x^{2} - \gamma^{2}y^{2}}{(1 + \gamma^{2}x^{2})(1 + \gamma^{2}x^{2})}\right] = \gamma^{2}(x-y)\left[\frac{\gamma x + \gamma y}{(1 + \gamma^{2}x^{2})(1 + \gamma^{2}x^{2})} \right] \] Hence \begin{align} |h_{\gamma}'(x) - h_{\gamma}'(y)| &= \gamma^{2}|x-y|\left[\frac{\gamma x}{(1 + \gamma^{2}x^{2})(1 + \gamma^{2}y^{2})} + \frac{\gamma y}{(1 + \gamma^{2}x^{2})(1 + \gamma^{2}y^{2})} \right] \nonumber \\ &\leq \gamma^{2}|x-y|\left[\frac{1}{2} + \frac{1}{2} \right] = \gamma^{2}|x-y|. \nonumber \end{align}
\noindent When $x \geq 0, y \leq 0$:  \[h_{\gamma}'(x) - h_{\gamma}'(y) = \gamma \left[\frac{\gamma^{2}x^{2} }{1 + \gamma^{2}x^{2}} + \frac{\gamma^{2}y^{2} }{1 + \gamma^{2}y^{2}} \right]  \] Hence \begin{align} |h_{\gamma}'(x) - h_{\gamma}'(y)| &= \gamma^{2}\abs{ |x|\left(\frac{\gamma|x|}{1 + \gamma^{2}|x|^{2}} \right) + |y|\left(\frac{\gamma|y|}{1 + \gamma^{2}|y|^{2}} \right) } \nonumber \\ &\leq \frac{\gamma^{2}}{2}\abs{|x| + |y|} = \frac{\gamma^{2}}{2}|x-y| \leq \gamma^{2}|x-y|. \nonumber \end{align} Proof for other cases follows similarly.

\section{Experimental results}

  We assessed the performance of regularized DNNs with the nonconvex penalty functions presented in this paper, by applying them on seven datasets (MNIST, FMNIST, RCV1, CIFAR-10, CIFAR-100, SVHN and ImageNet). We used convolutional neural networks with nonconvex regularization on four of the seven datasets: CIFAR-10, CIFAR-100, SVHN and ImageNet.

  \subsection*{Results based on Deep Neural Network analysis}

  Details of the DNN analysis and description of the three datasets are given below. The optimal weights of the fitted deep neural networks (DNN) were estimated by minimizing the total cross entropy loss function. We used batch gradient descent algorithm with early stopping. To avoid the vanishing/exploding gradients problem, the weights were initialized to values obtained from a normal distribution with mean zero and variance $4/(n_{i}+n_{(i-1)})$ where $n_{i}$ is the number of neurons in the $i^{th}$ layer (Glorot and Bengio, 2010; He et al., 2015). Rectified linear units (ReLU) function was used as the activation function.

  The training data was randomly split into multiple batches. During each epoch, the gradient descent algorithm was sequentially applied to each of these batches resulting in new weights estimates. At the end of each epoch, the total validation loss was calculated using the validation set. When twenty consecutive epochs failed to improve the total validation loss, the iteration was stopped. The maximum number of epochs was set at 250. The weights estimate that resulted in the lowest total validation loss was selected as the final estimate. Since there was a random aspect to the way the training sets were split into batches, the whole process was repeated three times with seed values 1, 2, and 3. The reported test error rates are the median of the three test error rates obtained using each of these seed values.

\begin{figure}[H]
\begin{center}
\includegraphics[height=3in,width=4in,angle=0]{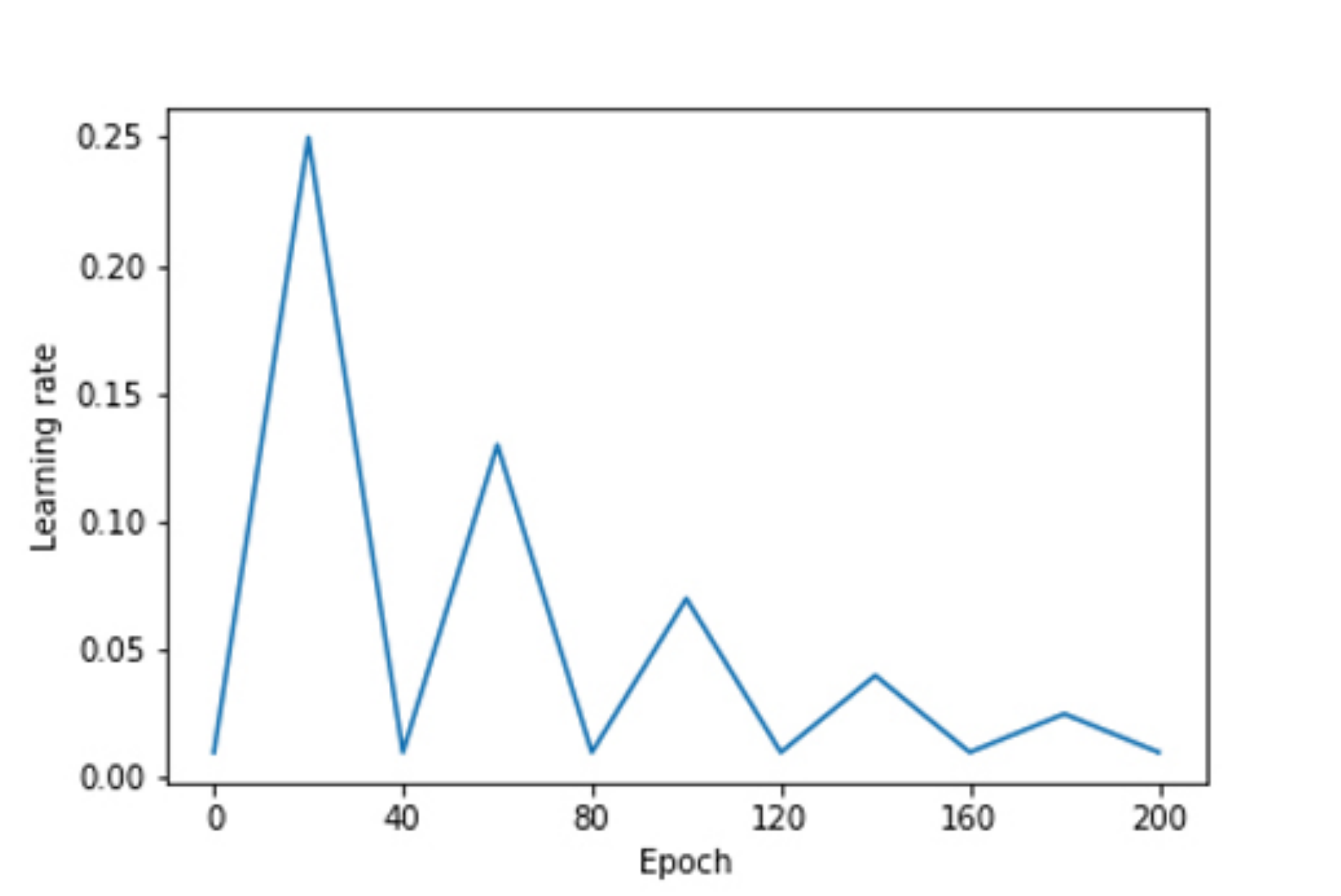}
\caption{Learning rate plot}
\end{center}
\end{figure}

   A triangular learning rate schedule was used because it produced the lowest test error rates (Smith, 2017). The learning rates varied from a minimum of 0.01 to a maximum of 0.25 (see Figure 5). For all penalty functions the optimal $\lambda$ was found by fitting models with logarithmically equidistant values in a grid. We used Python ver. 3.6.7rc2 and TensorFlow ver.1.12.0 for the calculations. The models were fit with no regularization, $L_{1}$ and $L_{2}$ regularizations and the non-convex regularization methods.  The results based on new non-convex penalty functions were comparable to or better than $L_{1}$ and $L_{2}$  regularization in all the datasets. A general overview of the datasets and the DNN model specifications is given in Table 1. The models were intentionally overparameterized to better contrast the effects of various types of regularization methods.

\begin{center}
    \begin{tabular}{ | l || c | c | c | c | c | c | c | }
    \hline
    \multicolumn{8}{|c|}{Table 1. Overview of dataset and DNN model specifications}   \\ \hline
    Dataset & Domain & Dimensionality & Classes	& DNN & 	Training&	Validation & Test   \\
        & & & 	& Specifications & 	Set & Set &	Set  \\ \hline \hline
    MNIST   & Visual & 784 ($28 \times 28$  & 10 & 5 layers,	& 48000	& 2000	& 10000 \\
            &        &  greyscale) &  & 1024 nodes & & &  \\ \hline
    FMNIST  & Visual & 784 ($28 \times 28$  & 10 & 5 layers,	& 45000	& 5000	& 10000 \\
            &        &  greyscale) &  & 1024 nodes & & &  \\ \hline

    Reuters	& Text	& 47236 & 	50	& 5 layers,
	& 13355	& 2000	& 49565 \\
            & 	& &	& 512 nodes	& &	&  \\ \hline
    \end{tabular}
\end{center}

For nonconvex penalties there is an extra parameter that we have to take into consideration: $a$ for SCAD, $b$ for MCP, $\varepsilon$ for Laplace and $\gamma$ for Arctan. In previous literature (Fan and Li, 2001), $a = 3.7$ has been suggested as an optimal parameter for SCAD based on Bayes risk criterion. Under different settings, $b  = 1.5, 5$ and $20$ have been considered as optimal values for the MCP parameter (Breheny and Huang, 2011). $a = 3.7$ for SCAD and $b  = 1.5, 5$ and $20$ are the parameter values that we considered for all DNN analyses in this paper. After trial and error runs with one dataset (MNIST), we picked $\gamma =$ 1 and 100 for Arctan and $\varepsilon = 10^{-7}$ for Laplace penalty.

All the results from DNN data analyses are summarized in Table 2 below. In Table 2 we present best test error rate across all the $\lambda$ values that we considered, for each method and each dataset. Standard errors are also presented to quantify statistical uncertainty. Test error rates for each $\lambda$ value are plotted in figures 6-8 corresponding to each dataset.  More detailed results based on all seed values are given in the tables in Appendix B. Appendix C provides $t$-test based p-values for pairwise comparisons between methods for each dataset. The samples for each group in the $t$-test consisted of the test error rates obtained by varying the seed values for the best test error rates. For comparisons of penalty based methods with no regularization, Bonferroni corrected p-values are also provided in the tables in Appendix C. A p-value threshold of 0.05 (5$\%$ significance level) was used to conclude whether any pairwise comparisons showed a difference above statistical uncertainty.

\begin{center}
    \begin{tabular}{ | l || c | c | c |  }
    \hline
    \multicolumn{4}{|c|}{Table 2. Median test error rates at optimal $\lambda$ (DNN) } \\ \hline
    Penalty function & \multicolumn{3}{|c|}{Dataset} \\ \hline
            & MNIST & FMNIST & RCV1 \\ \hline\hline
    None & 1.87 (0.11) & 11.94 (0.13) & 14.66 (0.32)\\ \hline
   $\mathrm{L}_{1}$ (Lasso) & 1.24 (0.05) & 10.06 (0.21) & 12.97 (0.15) \\ \hline
   $\mathrm{L}_{2}$ (Ridge) & \color{blue} 1.23  \color{black} (0.01) & 10.15 (0.02) & 13.77 (0.33) \\ \hline
     SCAD (a = 3.7) & 1.80 (0.07) & 11.45 (0.23) & 13.96 (0.03) \\ \hline
      MCP (b = 1.5) & 1.60 (0.22) & 11.39 (0.15) & 13.33 (0.22) \\ \hline
      MCP (b = 5)   & 1.67 (0.21) & 11.39 (0.12) & 14.44 (0.45) \\ \hline
      MCP (b = 20)  & 1.65 (0.16) & 11.33 (0.04) & 14.36 (0.37) \\ \hline
      Laplace ($\varepsilon = 10^{-7}$)  & \color{blue} 1.23 \color{black} (0.05) & 9.98 (0.25) & \color{blue}12.94 \color{black} (0.21) \\ \hline
      arctan ($\gamma$ = 1)    & 1.26 (0.04)& \color{blue} 9.87\color{black} (0.13) & 13.41 (0.12) \\ \hline
      arctan ($\gamma$ = 100)  & 1.25 (0.03) & 9.93 (0.23) & 13.81 (0.32) \\ \hline
      \multicolumn{4}{|c|}{Standard errors are given in parentheses} \\ \hline
    \end{tabular}
\end{center}

\underline{MNIST}:

Modified National Institute of Standards and Technology (MNIST) dataset is a widely used toy dataset of 60,000 grey-scale images of hand-written digits. Each image has $28 \times 28 = 784$ pixels. The intensity measures of these 784 pixels form the input variables of the model. The dataset was split into 48,000 training set, 2000 validation set, and 10,000 test set.

\begin{figure}[H]
\begin{center}
\hspace*{-1cm}
\includegraphics[height=3.5in,width=7in,angle=0]{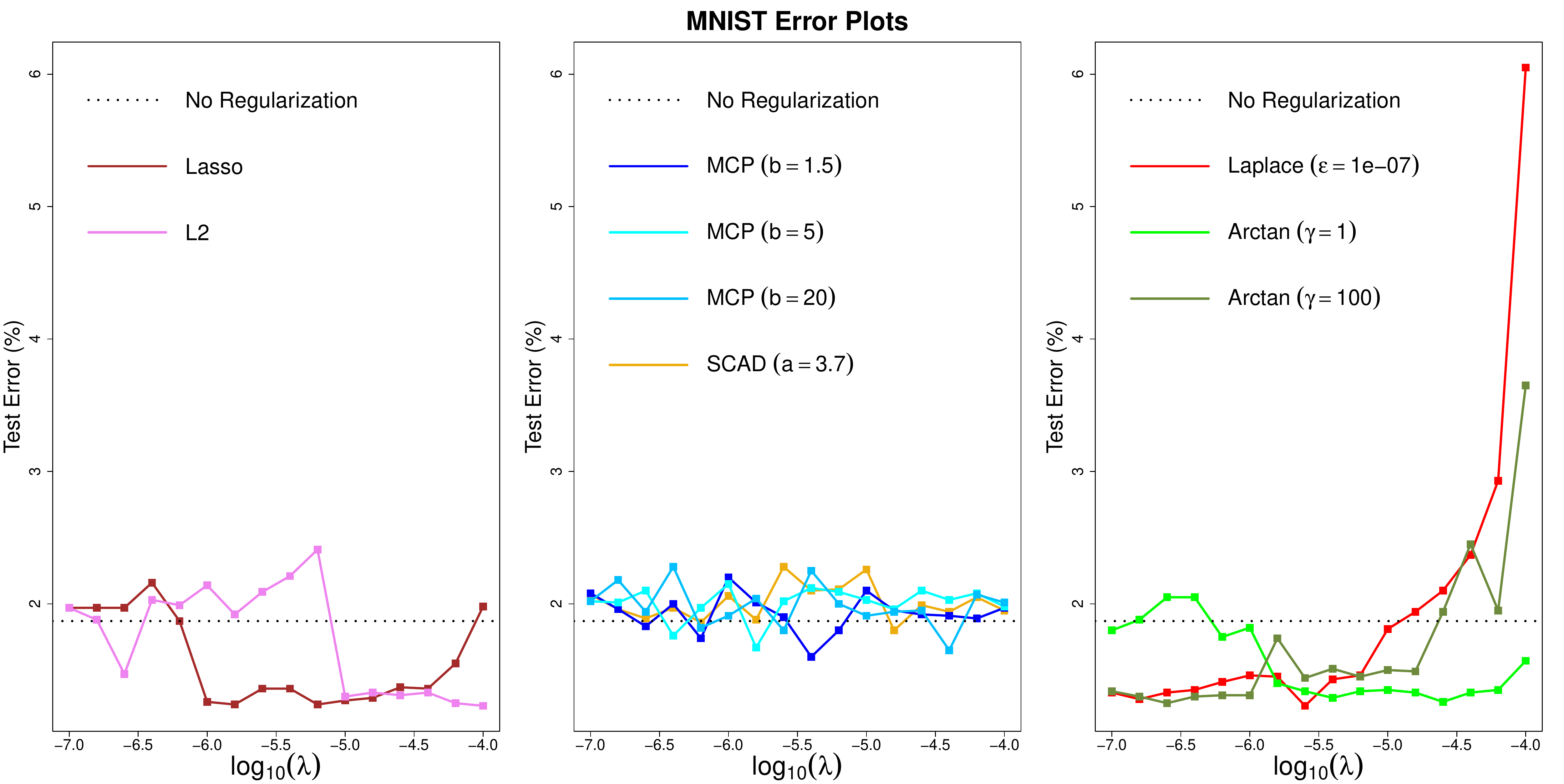}
\caption{Test error rates, from MNIST DNN analysis, corresponding to a grid of $\log_{10}(\lambda)$ values. Horizontal line in each panel corresponds to the error rate obtained without regularization. Left-most panel contains the error rates based on convex penalties; the middle and right panels contain error rates based on nonconvex penalties.}
\end{center}
\end{figure}

The test error rate obtained with no regularization was 1.87$\%$. With $L_{1}$ regularization, the test error reduced to 1.24$\%$ and with $L_{2}$ regularization the test error was even lower: 1.23$\%$. The test error rate of 1.23$\%$, which was the lowest among all the methods considered, was also obtained by the Laplace penalty. Arctan method gave test error rates of 1.26$\%$ and 1.25$\%$, with $\gamma =$ 1 and 100 respectively, which were comparable to error rates obtained with $L_{1}$ and $L_{2}$ regularization. SCAD and MCP methods improved the error rates compared with no regularization, but the results were not as good as those obtained with other penalty functions.

Pairwise comparisons table (Table C.1) shows that regularization with L$_{1}$, L$_{2}$, Laplace or arctan penalties significantly reduced the test error rates compared with no regularization, even with Bonferroni adjustment. The improvement seen with SCAD or MCP over no regularization was not statistically significant. The pairwise differences among L$_{1}$, L$_{2}$, Laplace and arctan penalties were only nominal and all of them differed significantly from SCAD ($a$ = 3.7) but not MCP penalty.

An interesting feature of the results based on Laplace and arctan penalties is their large fluctuation with $\lambda$. Although Laplace gave the best rate (1.23$\%$, same as $L_{2}$) for a particular $\lambda$, there was another $\lambda$ value for which the test error was near 6$\%$. This phenomenon was observed for error rates based on Laplace and arctan penalties from all other data analyses as well.

\underline{FMNIST}:

Fashion MNIST dataset consists of 60,000 grey-scale images of various types of clothing such as shirts, pants, and caps. There are 10 classes in total. The images have $28 \times 28 = 784$ pixels whose intensity measures were used as the input variables of the model. The 60,000 images were split into 45,000 training set, 5000 validation set, and 10,000 test set. FMNIST is very similar to MNIST because of similar characteristics except that the FMNIST test error rates tend to be much higher than MNIST test error rates.

\begin{figure}[H]
\begin{center}
\hspace*{-1cm}
\includegraphics[height=3.5in,width=7in,angle=0]{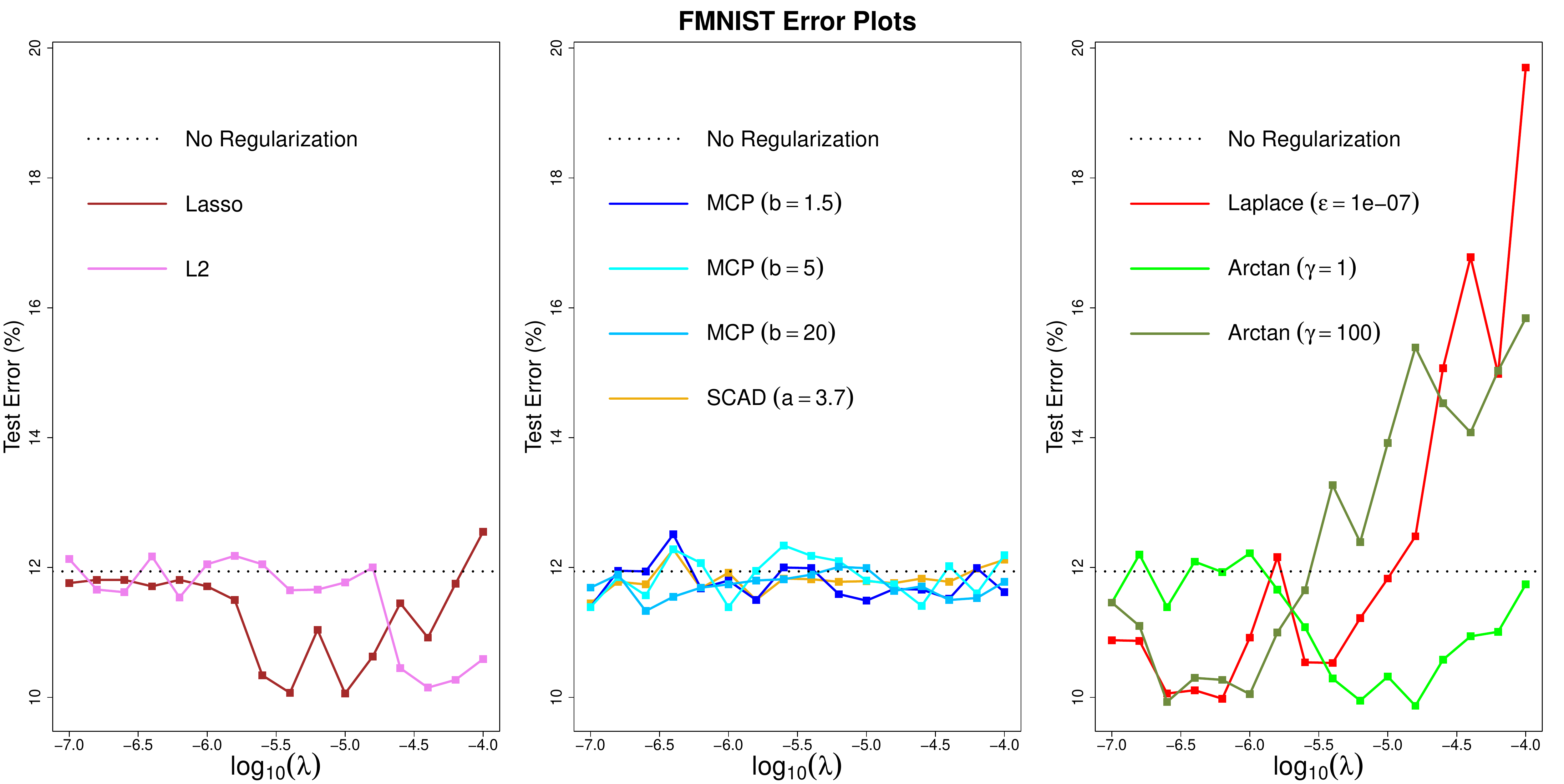}
\caption{Test error rates, from FMNIST DNN analysis, corresponding to a grid of $\log_{10}(\lambda)$ values. Horizontal line in each panel corresponds to the error rate obtained without regularization. Left-most panel contains the error rates based on convex penalties; the middle and right panels contain error rates based on nonconvex penalties.}
\end{center}
\end{figure}

The test error rate obtained with no regularization was 11.94$\%$. With $L_{2}$ regularization, the test error reduced to 10.15$\%$, and with $L_{1}$ the test error rate was 10.06$\%$. SCAD and MCP improved upon the error rate obtained without any regularization, but not lower than those obtained with $L_{2}$ or $L_{1}$. The new nonconvex penalties presented in this paper achieved lower rates than $L_{1}$ and $L_{2}$, with arctan ($\gamma$ = 1) giving the lowest error rate: Laplace, 9.98$\%$, arctan ($\gamma$ = 1), 9.87$\%$ and arctan ($\gamma$ = 100), 9.93$\%$.

Pairwise comparions (Table C.2) showed a statistically significant improvement for all penalties, except SCAD, compared to no regularization. After Bonferroni correction, only $L_{2}$, MCP ($b$ = 20) and arctan ($\gamma$ = 1) showed statistically significant improvement compared to no regularization. SCAD showed statistically significant difference from all other penalties. Note again that we used only one parameter value, $a$ = 3.7, for SCAD based on recommendations from existing literature. However, there could be other parameter values for which the performance of SCAD may be comparable with that of other penalties. The improvement in performance of Laplace and arctan penalty over SCAD and MCP was substantial at a statistical significance level, but was only nominal compared to $L_{2}$ or $L_{1}$.

\underline{RCV1}:

Reuters Corpus Volume I (RCV1) is a collection of 804,414 newswire articles labelled as belonging to one or more of 103 news categories (Lewis \textit{et. al.}, 2004). In our analysis, only single-labelled data points were used and all the multi-labelled data points were excluded resulting in a dataset consisting of news wire articles from 50 news categories. The cosine-normalized, log TF-IDF values of 47,236 words appearing in these news articles were used as the input variables for the model. The training set consisted of 13,355 data points, the validation set consisted of 2000 data points and the test set consisted of 49,565 data points.

\begin{figure}[H]
\begin{center}
\hspace*{-1cm}
\includegraphics[height=3.5in,width=7in,angle=0]{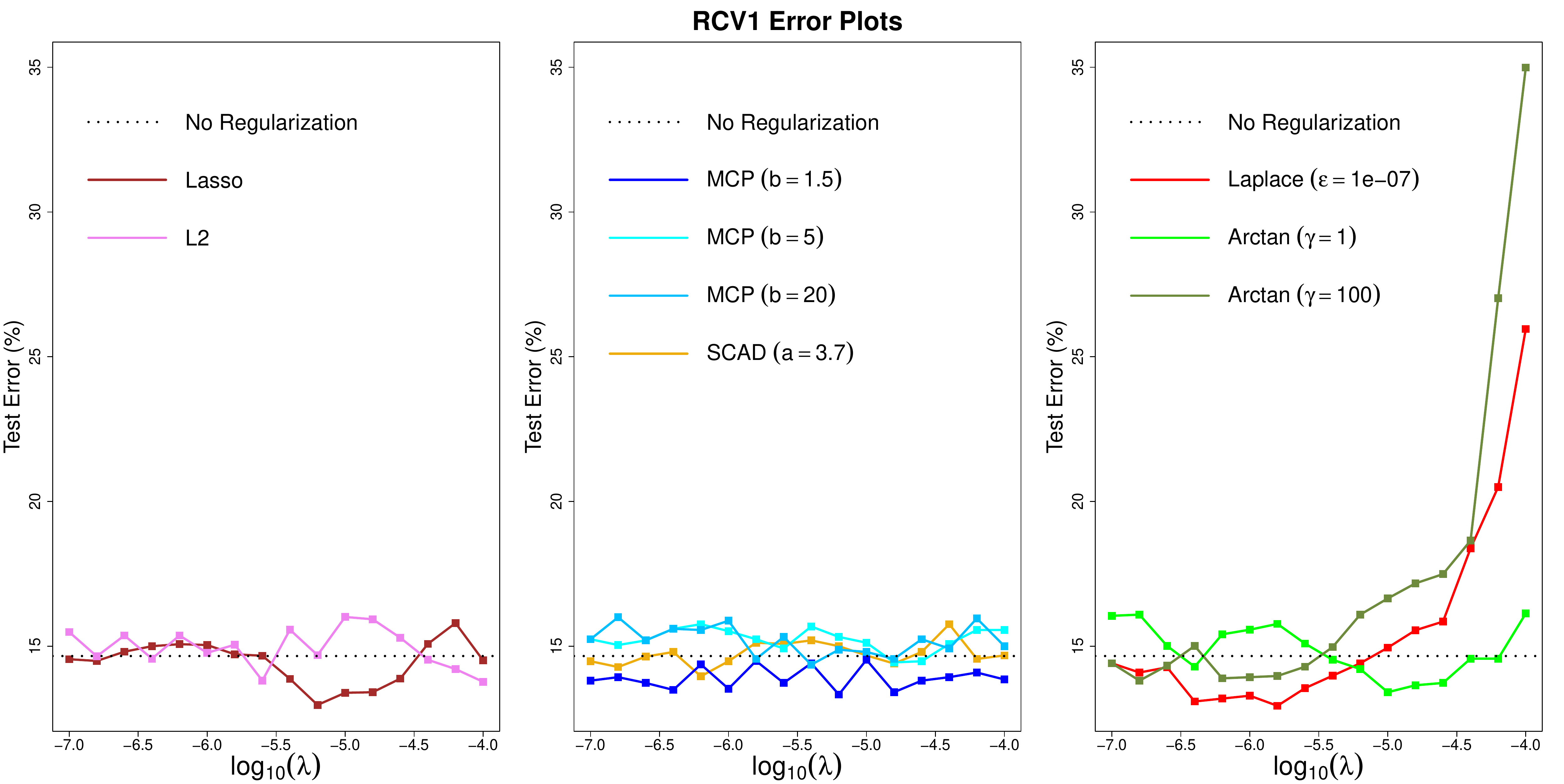}
\caption{Test error rates, from RCV1 (Reuters) DNN analysis, corresponding to a grid of $\log_{10}(\lambda)$ values. Horizontal line in each panel corresponds to the error rate obtained without regularization. Left-most panel contains the error rates based on convex penalties; the middle and right panels contain error rates based on nonconvex penalties.}
\end{center}
\end{figure}

The test error rate obtained with no regularization was 14.66$\%$. With Lasso regularization, the test error reduced to 12.97$\%$. Laplace method gave the lowest test error of 12.94$\%$. The test error rate with all other methods were below 14.66$\%$ but above 13$\%$. MCP with $b =$ 1.5 fared better than arctan penalty for this dataset. Pairwise comparisons (Table C.3) showed L$_{1}$, L$_{2}$, SCAD, MCP ($b$ = 1.5), Laplace and arctan penalties having a statistically significant improvement compared to no regularization, but only the p-values for L$_{1}$, Laplace and arctan ($\gamma$ = 1) survived Bonferroni correction. Although the Laplace penalty gave the best rate, statistically it was different only with the rate based on MCP ($b$ = 20) penalty.

\subsection*{Results based on Convolutional Neural Networks (CNN) analysis}

The CNN architecture that we used consisted of three convolutional ``blocks'' followed by a hidden ``block''. Each of the three convolutional blocks consisted of a convolutional layer with 96, 128 and 256 filters respectively, kernel size of 5, stride value of 1, and “same” padding, followed by batch normalization with momentum value for the moving average set to 0.9. This was further followed by a ReLU activation layer and finally a max-pooling layer with kernel-size of 3, stride value of 2, and ``same'' padding. In the hidden block, the data was first flattened and then passed through two fully connected hidden layers of 4096 nodes with ReLU activation function. Finally, the signals were classified into various categories by calculating the sparse softmax cross entropy values between logits and labels. The CNN model specification is further described in Figure 9.

\begin{figure}[H]
\begin{center}
\hspace*{-2.2cm}
\includegraphics[height=2.7in,width=8.1in,angle=0]{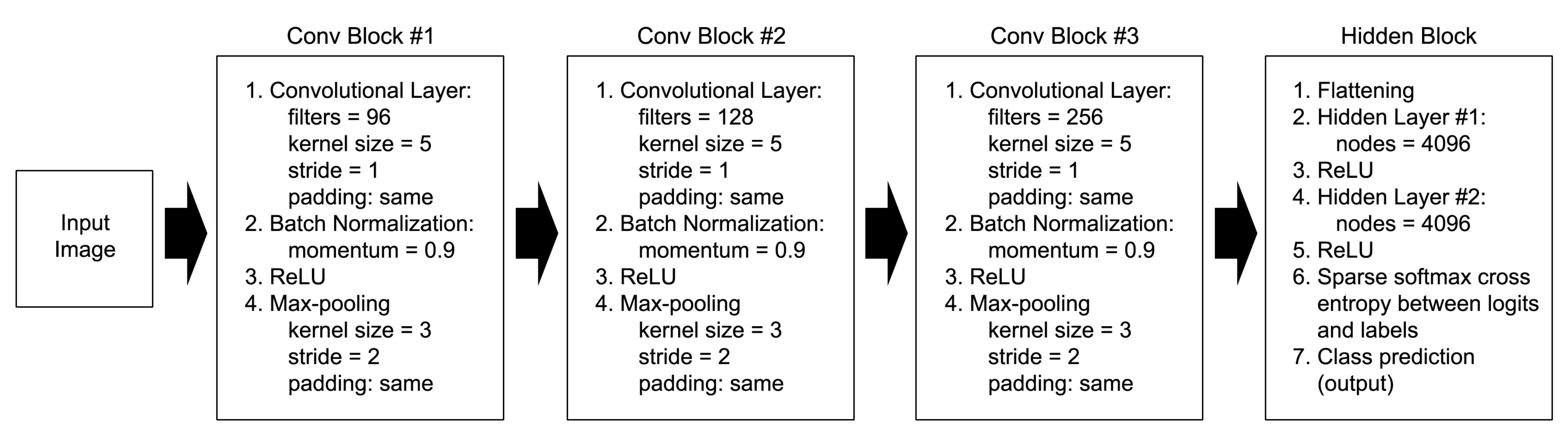}
\caption{Specifications for Convolutional Neural Network architecture}
\end{center}
\end{figure}

We assessed the performance of regularized CNNs with penalty functions discussed in this paper by applying them on four datasets (CIFAR10, CIFAR100, SVHN, ImageNet). More details of these datasets are given in Table 3. Since SCAD and MCP did not perform as well as the other penalties in the DNN analyses, we restricted our focus to only Laplace and arctan as nonconvex penalty-candidates for CNN analyses.

\begin{center}
    \begin{tabular}{ | l || c | c | c | c | c | c | }
    \hline
    \multicolumn{7}{|c|}{Table 3. Overview of datasets used for Convolutional Neural Networks}   \\ \hline
    Dataset & Domain & Dimensionality & Classes	& 	Training &	Validation & Test   \\
        & & & 	& Set & Set &	Set  \\ \hline \hline
    CIFAR-10  & Visual & 3072 ($32 \times 32 \times 3$  & 10 & 48000	& 2000	& 10000 \\
            &        &  color) &  & & & \\ \hline
    CIFAR-100  & Visual & 3072 ($32 \times 32 \times 3$  & 100 & 48000	& 2000	& 10000 \\
            &        &  color) &  & & &  \\ \hline
    SVHN & Visual & 3072 ($32 \times 32 \times 3$  & 10 & 71257	& 2000	& 26032 \\
            &        &  color) &  & & & \\ \hline
    Reduced size& Visual & 12288 ($64 \times 64 \times 3$  & 200 & 90000 & 10000	& 10000  \\
    ImageNet&        &  color) &  & &  &\\ \hline
    \end{tabular}
\end{center}

\underline{CIFAR-10 and CIFAR-100}:

CIFAR10 and CIFAR100 are datasets of 50000 32 $\times$ 32 color training images and 10000 test images. The images are classified into 10 categories for CIFAR-10 and 100 categories for CIFAR-100. The test error rates obtained  for each $\lambda$ value are plotted in Figure 10 for CIFAR-10 and Figure 11 for CIFAR100.

\begin{figure}[H]
\begin{center}
\hspace*{-1cm}
\includegraphics[height=3in,width= 6in,angle=0]{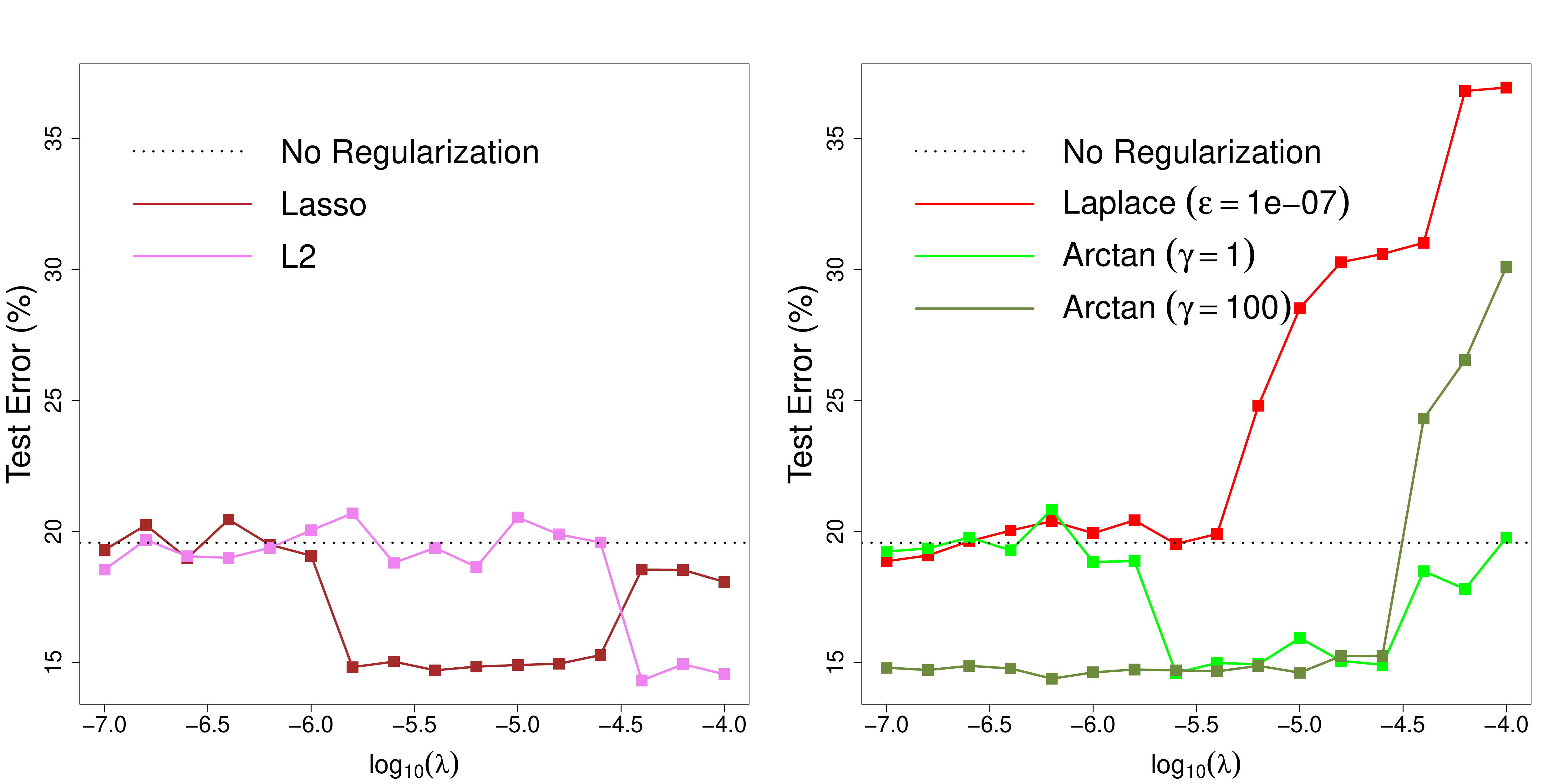}
\caption{Test error rates, from CIFAR-10 CNN analysis, corresponding to a grid of $\log_{10}(\lambda)$ values. Horizontal line in each panel corresponds to the error rate obtained without regularization. Left panel contains the error rates based on convex penalties; right panel contains error rates based on the new nonconvex penalties.}
\end{center}
\end{figure}

\begin{figure}[H]
\begin{center}
\hspace*{-1cm}
\includegraphics[height=3in,width= 6in,angle=0]{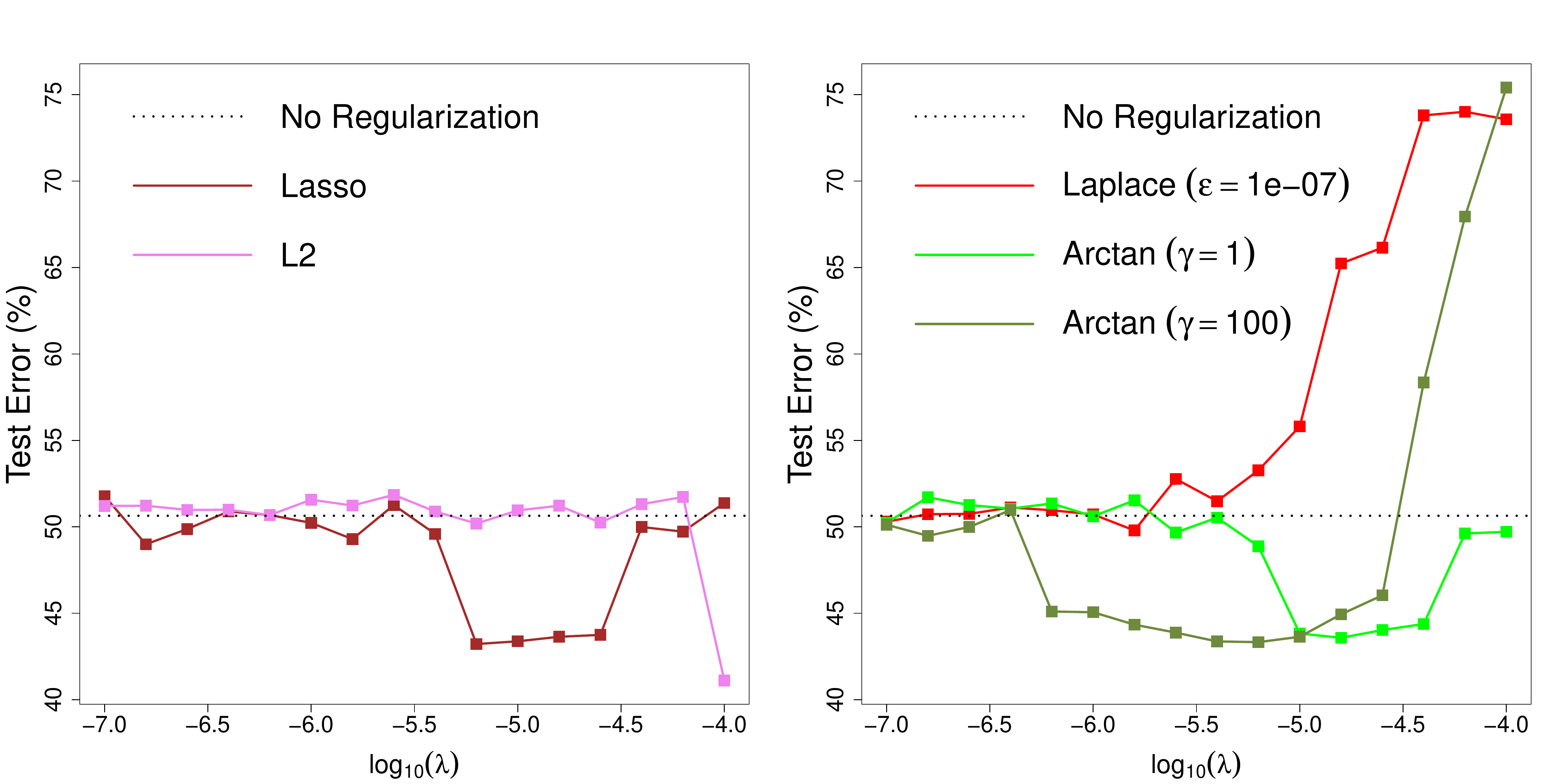}
\caption{Test error rates, from CIFAR-100 CNN analysis, corresponding to a grid of $\log_{10}(\lambda)$ values. Horizontal line in each panel corresponds to the error rate obtained without regularization. Left panel contains the error rates based on convex penalties; right panel contains error rates based on the new nonconvex penalties.}
\end{center}
\end{figure}

For CIFAR-10 dataset, the best test error rate was obtained with L$_{2}$ regularization (14.32$\%$) followed by arctan (14.39$\%$ with $\gamma$ = 100 and 14.61$\%$ with $\gamma$ = 1) and L$_{1}$ (14.71$\%$). Pairwise comparisons showed only nominal differences among L$_{1}$, L$_{2}$ and arctan penalties. However, the best test error rates based on L$_{1}$, L$_{2}$ and arctan differed statistically significantly from those obtained without regularization and with Laplace regularization.

For CIFAR-100 data analysis, the pattern of results were similar to those obtained for CIFAR-10 analysis. In this case, though, Lasso performed slightly better than arctan penalty. The best rate was exhibited by L$_{2}$ (41.11$\%$), followed by Lasso (43.22$\%$) and arctan  (43.33$\%$ with $\gamma$ = 100 and 43.58$\%$ with $\gamma$ = 1). The results based on the above penalties were better than those based on Laplace penalty or no regularization, at a statistically significant level. One aspect seen in favor of arctan penalties for this particular dataset is the smaller variance exhibited, which in turn led to p-values less than 0.0001 even after Bonferroni correction.

\underline{Street View House Numbers (SVHN)}:

The SVHN dataset is a collection of 32 $\times$ 32 color images of digits 0 to 9 obtained from house numbers in Google Street View images. SVHN is a bigger dataset than MNIST: 71257 training, 2000 validation, 26032 test.The images are in a natural setting unlike in MNIST, and hence the classification problem is a bit harder.

\begin{figure}[H]
\begin{center}
\hspace*{-1cm}
\includegraphics[height=3in,width= 6in,angle=0]{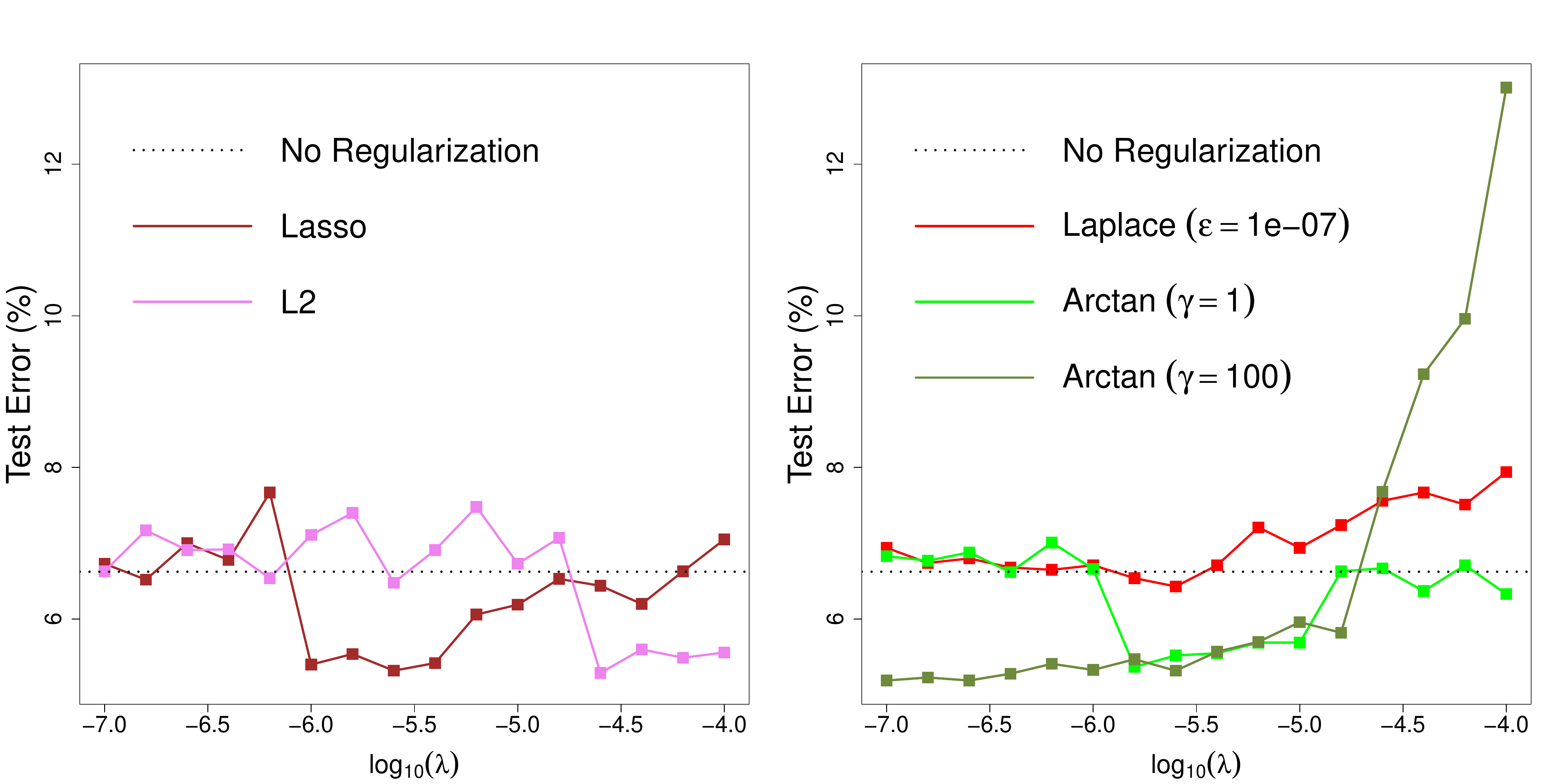}
\caption{Test error rates, from SVHN CNN analysis, corresponding to a grid of $\log_{10}(\lambda)$ values. Horizontal line in each panel corresponds to the error rate obtained without regularization. Left panel contains the error rates based on convex penalties; right panel contains error rates based on the new nonconvex penalties.}
\end{center}
\end{figure}

Figure 12 plots the test error rates for SVHN analyses for each $\lambda$ value. The best error rate was obtained by arctan regularization with $\gamma$ = 100 (5.19$\%$) followed by L$_{2}$ (5.29$\%$), L$_{1}$ (5.32$\%$) and arctan with $\gamma$ = 1 (5.37$\%$). The rates of all the above penalties differed at a statistical significance level from those obtained with no regularization and those obtained with Laplace penalty (Table C.6). The pairwise differences among arctan, L$_{1}$ and L$_{2}$ were only nominal.

\underline{ImageNet}:

The original ImageNet dataset for ILSVRC-2014 challenge had a total of 476688 color images in the training and validation sets (http://www.image-net.org/challenges/LSVRC/2014/). These images were of varying sizes. Since our analysis is mainly for illustration purposes, we used a subset of the full ILSVRC-2014 dataset with 200 classes. The color images in this subset were resized from their original sizes to 64 $\times$ 64 pixels. The smaller subset helped us to reduce substantially the computational time and resources required to train the models. However, the subset that we used is still large and challenging enough to obtain meaningful comparisons. In a sense, reducing the image counts and resolution made the classification problem even more challenging in terms of accuracy compared to the original ILSVRC-2014 classification problem. For each class, there were 450 training images, 50 validation images, and 50 test images. The final image counts for training, validation, and test datasets in our subset were 90000, 10000 and 10000, respectively. The class IDs of the images that we used are available as a text file in our web-appendix. In order to further reduce the time for overall computations, we considered only one seed for all analyses for this particular dataset.

Figure 13 plots the test error rates for ImageNet analyses for each $\lambda$ value. The best error rate was obtained by arctan regularization with $\gamma$ = 1 (40.98$\%$) followed by L$_{1}$ (41.32$\%$) and L$_{2}$ (41.55$\%$) penalties. Since we conducted this analysis only for one seed value, pairwise comparisons were not possible.

\begin{figure}[H]
\begin{center}
\hspace*{-1cm}
\includegraphics[height=3in,width= 6in,angle=0]{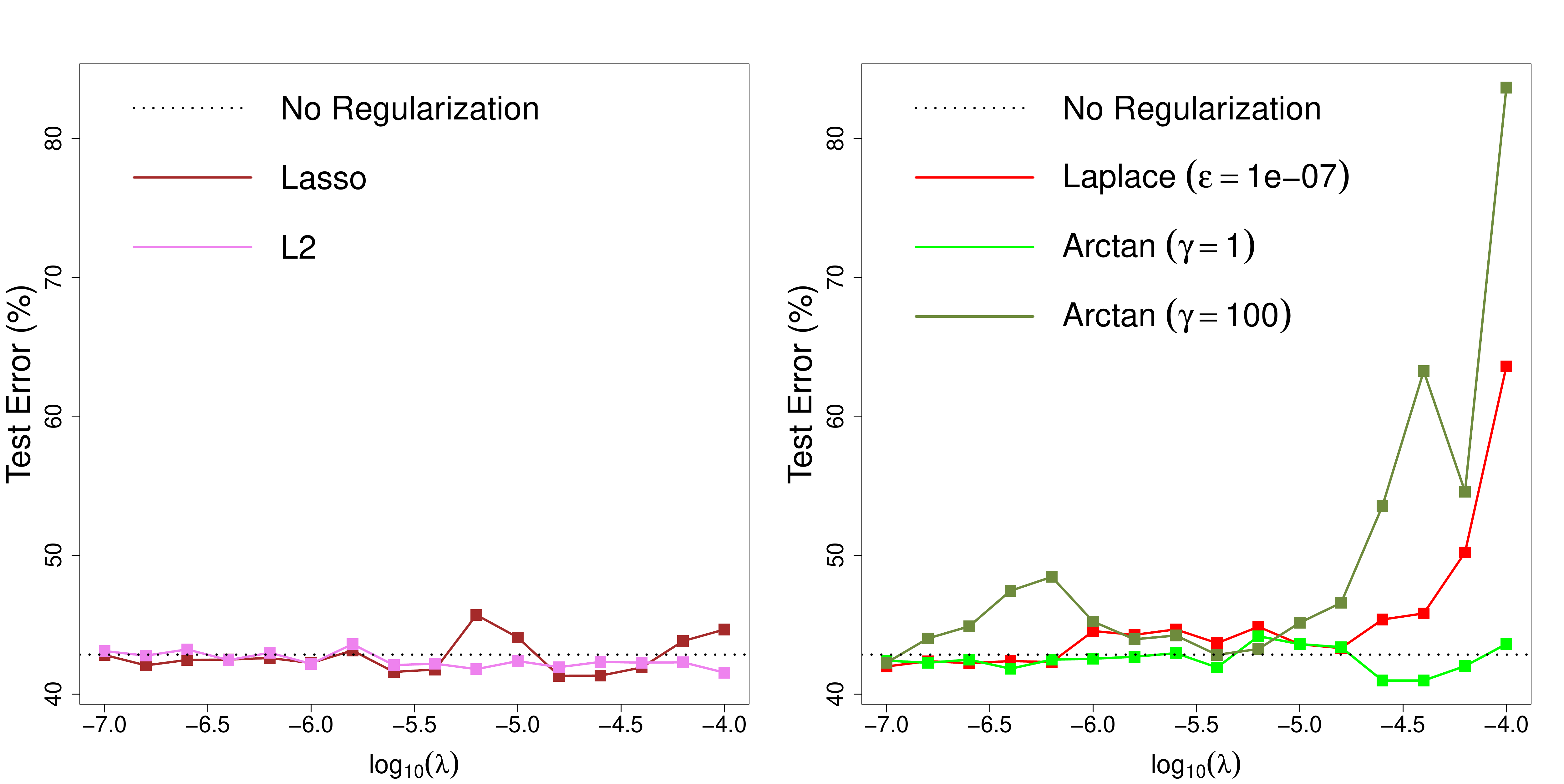}
\caption{Test error rates, from ImageNet CNN analysis, corresponding to a grid of $\log_{10}(\lambda)$ values. Horizontal line in each panel corresponds to the error rate obtained without regularization. Left panel contains the error rates based on convex penalties; right panel contains error rates based on the new nonconvex penalties.}
\end{center}
\end{figure}

The best test error rates based on the optimal $\lambda$ value for each method are summarized in Table 4 below. The rate for the best method for each dataset is highlighted in blue

\begin{center}
    \begin{tabular}{ | l || c | c | c | c |  }
    \hline
    \multicolumn{5}{|c|}{Table 4. Median test error rates at optimal $\lambda$ (CNN)} \\ \hline
    Penalty function & \multicolumn{4}{|c|}{Dataset} \\ \hline
            & CIFAR-10 & CIFAR-100 & SVHN & ImageNet \\ \hline\hline
    None & 19.58 (0.24) & 50.64 (0.25) & 6.63 (0.10) & 42.85 (0.19)\\ \hline
   $\mathrm{L}_{1}$ (Lasso) & 14.71 (0.15)  & 43.22 (0.36) & 5.32 (0.06) & 41.32 (NA) \\ \hline
   $\mathrm{L}_{2}$ (Ridge) & \color{blue} 14.32 \color{black} (0.25) & \color{blue} 41.11 \color{black} (0.61)  & 5.29 (0.11) & 41.55 (NA) \\ \hline
      Laplace ($\varepsilon = 10^{-7}$)  & 18.87 (0.36) & 49.79 (0.69) & 6.43 (0.14) & 41.99 (NA) \\ \hline
      arctan ($\gamma$ = 1) & 14.61 (0.06) & 43.58 (0.04) & 5.37 (0.05) & \color{blue} 40.98 \color{black} (NA) \\ \hline
      arctan ($\gamma$ = 100)  & 14.39 (0.17) & 43.33 (0.26) & \color{blue} 5.19 \color{black} (0.09)   & 42.26 (NA) \\ \hline
      \multicolumn{5}{|c|}{Standard errors are given in parenthesis} \\ \hline
    \end{tabular}
\end{center}

\section{Conclusions and Discussion}

Nonconvex regularizers were originally considered in statistical literature after observing certain limitations of the convex regularizers from the class of Bridge functions. Yet, nonconvex regularizers never gained as much popularity as their convex counterparts in DNN applications, perhaps because of certain perceived computational and optimization limitations - that is, in the presence of local optima which are not global optima for the case of nonconvex functions, iterative methods such as gradient or coordinate descent may terminate undesirably at a local optimum. However, recent theoretical work (Loh and Wainwright, 2015, 2017) that established regularity conditions under which both local and global minimum lie within a small neighborhood of the true minimum have made nonconvex regularizers worth considering for DNN. The new theory eliminates the need for specially designed optimization algorithms for most nonconvex regularizers as it implies that standard first-order optimization methods will converge to points within statistical error of the truth. In other words, nonconvex regularizers that satisfy such regularity conditions enjoy guarantees for both statistical accuracy and optimization efficiency.

In this paper we considered, for DNN regularization, nonconvex penalty functions that satisfy the regularity conditions mentioned above. We studied theoretical properties of solutions based on two new nonconvex penalties (Laplace and arctan). We also studied the performance of the penalty functions for regularization of DNN applied on three large datasets (MNIST, FMNIST and RCV1) and for regularization of CNN applied on four large datasets (CIFAR-10, CIFAR-100, SVHN and ImageNet). In the first three cases test error rates for Laplace and arctan penalty functions, and in the last four cases test error rates for arctan penalty were comparable to those obtained by the convex penalties. In five out of the seven datasets considered, at least one of the new nonconvex penalties gave a nominally better test error rate than convex regularization. However the differences in test error rates between the best-performing nonconvex penalty for each dataset and convex penalties were within statistical uncertainty.

Another popular and very efficient approach for DNN regularization is \textit{Dropout} (Srivastava et al., 2014) in which units along with their connections are randomly dropped. One way to understand the good performance of \textit{Dropout} is within the statistical framework of model averaging. A more recently proposed method for DNN regularization is \textit{Shakeout} (Kang, Li and Tao, 2017) in which a unit's contribution to the next layer is randomly enhanced or reversed. By choosing the enhance factor to be 1 and the reverse factor to be 0, \textit{Dropout} can be considered as a special case of \textit{Shakeout}. When the data is scarce, it has been observed that \textit{Shakeout} outperforms \textit{Dropout} (Kang, Li and Tao, 2017). Although the operating characteristics and implementation of \textit{Dropout} and \textit{Shakeout} are different from the methods mentioned in this paper, there are certain connections among the underlying paradigms. By marginalizing the noise, equivalency between \textit{Dropout} and adaptive $L_{2}$ regularization has been established in the literature (Srivastava et al., 2014; Wager, Wang and Liang, 2013). Theoretically, \textit{Shakeout} regularizer can be seen as a linear combination of $L_{0}$, $L_{1}$ and $L_{2}$ penalty functions (Kang, Li and Tao, 2017). Plotting the regularization effect as a function of a single weight (e.g. see Figure 2 in Kang, Li and Tao, 2017) makes the comparison even more clearer. \textit{Dropout} and \textit{Shakeout} are similar to nonconvex regularizers except in a neighborhood near zero. Near zero, the \textit{Dropout} regularizer is similar to $L_{2}$ penalty function (i.e. quadratic), while the \textit{Shakeout} regularizer is sharp and discontinuous.

Nonconvex penalties introduce yet another parameter into the objective function. Ideally a grid of values in the domain of this extra parameter should be considered. However, this could lead to substantial increase in computational time. Cross-validation strategy that may work for small scale datasets becomes computationally intensive in DNN and CNN applications. Based on Bayes risk criterion and simulations, $a = 3.7$ was suggested as an optimal value for the extra parameter in SCAD (Fan and Li, 2001). For MCP parameter $b$, a value of 1.5 was found optimal in one setting while as values 5 and 20 were optimal (compared to Lasso) in another setting in (Breheny and Huang, 2011). The method, based on path of solutions, suggested in (Breheny and Huang, 2011) could be considered for determining the optimal value of the extra parameter (say $k$) in any non-convex penalty. In this method, first an interval for $\lambda$ is determined in which the objective function is ``locally convex''. For a given $k$, if the chosen solution lies in this region, then $k$ can be increased to make the penalty more convex, and if it lies outside the region one can lower $k$ without fear of nonconvexity. Iterate this process a few times to obtain a $k$ that strikes a balance between parsimony and convexity of the objective function. Other parameter selection methods within the deep learning literature such as grid search, random search (Bergstra and Bengio, 2012), Bayesian optimization methods (Snoek, Larochelle and Adams, 2015) and gradient-based optimization (Maclaurin, Duvenaud and Adams, 2015) could also be used. For DNN and CNN data analyses presented in this paper, we considered only one parameter value ($\varepsilon$ = 1e-07) for the Laplace penalty and two parameter values ($\gamma$ = 1 and 100) for arctan  penalty. The performance of the regularization methods even with this limited range of values for the corresponding parameters was very good for the datasets that we considered.

\section*{Acknowledgements} We gratefully acknowledge the constructive feedback from an AE and three anonymous reviewers which led to substantial improvement of the manuscript.


\section*{References}

Agarwal, A., Negahban, S., and Wainwright, M.J. (2012). Fast global convergence of gradient methods for high-dimensional statistical recovery. Annals of Statistics, 40(5):2452-2482.

Ahn M., Pang J.-S., and Xin J. (2017). Difference-of-Convex Learning: Directional Stationarity, Optimality, and Sparsity, SIAM J. Optim., 27(3), 1637-1665.

Bergstra, J., and Bengio, Y. (2012). Random search for hyper-parameter optimization. Journal of Machine Learning Research, 13: 281-305.

Breheny, P., and Huang, J. (2011). Coordinate descent algorithms for nonconvex penalized regression, with applications to biological feature selection. The Annals of Applied Statistics, 5(1): 232-253.

Breiman, L. (1996) Heuristics of instability and stabilization in model selection. The Annals of Statistics, 24 (6): 235-2383.

Chen, L., Gu, Y. (2014). The convergence guarantees of a non-convex approach for sparse recovery. IEEE Transactions on Signal Processing, 62(15): 3754-3767.

Duchi, J. and Singer, Y. (2009). Efficient Online and Batch Learning Using Forward Backward Splitting. JMLR, 10: 2899-2934.

Fan, J. and Li, R. (2001). Variable selection via nonconcave penalized likelihood and its oracle properties. Journal of the American Statistical Association, 96: 1348–1360.

Frank, I. and Friedman, J. (1993). A statistical view of some chemometrics regression tools. Technometrics, 35: 109–148.

Fu, W. J. (1998). Penalized regressions: the Bridge versus the Lasso. J. Comput. Graph. Statist, 7: 397–416.

Geman, D and Yang, C. (1995). Nonlinear image recovery with half-quadratic regularization.  IEEE Transactions on Image Processing, 4(7): 932-946.


Glorot, X., and Bengio, Y. (2010). Understanding the difficulty of training deep feedforward neural networks. PMLR 9: 249-256.

Hastie, T., Tibshirani, R., and Friedman, J. H. (2009). The elements of statistical learning: data mining, inference, and prediction. 2nd ed. New York: Springer.

He, K.,  Zhang, X., Ren, S., Sun, J. (2015) Delving Deep into Rectifiers: Surpassing Human-Level Performance on ImageNet Classification. arXiv eprint 1502.01852.

Hoerl, A.E. and Kennard, R. (1970). Ridge regression: Biased estimation for nonorthogonal problems. Technometrics, 12: 55-67.

Huang, J., Horowitz, J.L., Ma, S. (2008). Asymptotic properties of bridge estimators in sparse high-dimensional regression models. The Annals of Statistics, 36 (2): 587-613.

Kang, G., Li, J., Tao, D. (2017). Shakeout: A new approach to regularized deep neural network training. IEEE Trans. Pattern Anal. Mach. Int., 40(5): 1245-1258.

Kim, J.and Pollard, D.(1990).  Cube root asymptotics. The Annals of Statistics. 18: 191–219.

Knight, K. and Fu, W. (2000). Asymptotics for lasso-type estimators. Ann. Statist. 28(5): 1356-1378.

Le Thi H. A., Huynh V. N. and Pham D. T. (2019). Stochastic difference-of-convexalgorithms for solving nonconvex optimization problems.arXiv:1911.04334.

Le Thi H. A. and Pham D. T. (2018). DC programming and DCA : thirty years of developments, Mathematical Programming 69(1), 5-68 (2018).

Lewis, D. D., Yang, Y., Rose, T. G.,  Li, F. (2004). RCV1: A new benchmark collection for text categorization research. The Journal of Machine Learning Research, 5: 361-397.

Loh, P. and Wainwright, M.J. (2012). High-dimensional regression with noisy and missing data: Provable guarantees with non-convexity. Annals of Statistics, 40(3):1637-1664.

Loh, P. and Wainwright, M.J. (2015). Regularized M-estimators with nonconvexity: Statistical and algorithmic theory for local optima. Journal of Machine Learning Research, 16: 559-616.

Loh, P. and Wainwright, M.J. (2017) Support recovery without incoherence: A case for nonconvex regularization. Annals of Statistics 45(6): 2455-2482.

Lu, C., Tang, J., Yan, S. and Lin, Z. (2014). Generalized Nonconvex Nonsmooth Low-Rank Minimization. In Proceedings of the 2014 IEEE Conference on Computer Vision and Pattern Recognition (CVPR '14). IEEE Computer Society, Washington, DC, USA, 4130-4137.

Maclaurin, D., Duvenaud, D., and Adams, R.P. (2015). Gradient-based hyperparameter optimization through reversible learning. Proceedings of the 32nd international conference of machine learning.

Meier, L., van de Geer, S., Buehlmann, P (2008). The group Lasso for logistic regression. JRSS, Series B, 70, 53-71.

Nesterov, Y. (2007). Gradient methods for minimizing composite objective function. CORE Discussion Papers 2007076, Universit Catholique de Louvain, Center for Operations Research and Econometrics (CORE), 2007. URL http://EconPapers.repec.org/RePEc:cor:louvco:2007076.

Pham D. T., Souad E.B. (1986) Algorithms for solving a class of nonconvex optimization problems.Methods of subgradients. In: J.B. Hiriart-Urruty (ed.) Fermat Days 85: Mathematics forOptimization,North-Holland Mathematics Studies,129, 249-271.

Pollard, D. (1991).  Asymptotics for least absolute deviation regression estimators. Econometric Theory 7: 186–199.

Rudin, W. (1976) Principles of Mathematical Analysis. 3rd Edition, McGraw-Hill, New York.

Rumelhart, D. E., Hinton, G. E., and Williams, R. J. (1986). Learning representations by back-propagating errors. Nature, 323: 533-536.

Rumelhart, D. E., Hinton, G. E., and Williams, R. J. (1986). Learning internal representations by error propagation. In Rumelhart, D. E. and McClelland, J. L., editors, Parallel Distributed Processing: Explorations in the Microstructure of Cognition. Volume 1: Foundations Volume 1: Foundations, MIT Press, Cambridge, MA.

Smith, L.N. (2017). Cyclical Learning Rates for Training Neural Networks. arXiv eprint: 1506.01186v6.

Snoek, J., Larochelle, H and Adams, R.P. (2012). Practical Bayesian optimization of machine learning algorithms. Advances in Neural Information processing systems 25, pp. 2951-2959.

Srivastava, N., Hinton, G., Krizhevsky, A., Sutskever, I., Salakhutdinov, R. (2014). Dropout: A simple way to prevent neural network from overfitting. Journal of Machine Learning Research, 15: 1929-1958.

Tarigan, B. and van de Geer, S. (2006). Classifiers of support vector machine type with L1 complexity regularization. Bernoulli, 12: 1045–1076.

Tibshirani, R. (1996). Regression shrinkage and selection via the lasso. J. Royal. Statist. Soc B., 58(1): 267-288.

Trzasko, J and Manduca, A. (2009). Highly undersampled magnetic resonance image reconstruction via homotopic L0-minimization.  IEEE Transactions on Medical Imaging, 28(1): 106-121.

van der Vaart, A.W. (1998). Asymptotic Statistics. Cambridge University Press, New York.

van der Vaart, A.W. and Wellner, J.A. (1996). Weak convergence and Empirical Processes: With Applications to Statistics. Springer, New York.

Vial, J.-P. (1982). Strong convexity of sets and functions.  Journal of Mathematical Economics, 9 (1-2): 187-205.

Wager, S., Wang, S., and Liang, P.S. (2013). Dropout training as adaptive regularization. Advances in Neural Information Processing Systems 26, pp. 351-359.

Yin, P., Lou, Y., He, Q., Xin, J. (2015). Minimization of l$_{1-2}$ for compressed sensing. SIAM J. Sci. Comput. 37(1), 536–563.

Zhang, C.H. (2010). Nearly unbiased variable selection under minimax concave penalty. Annals of Statistics, 38(2): 894–942.

Zhang, H., Ahn, J., Lin, X., Park, C. (2005) Gene selection using support vectormachines with non-convex penalty. Bioinformatics 22:88–95.

Zhang, C.-H., Zhang, T. A general theory of concave regularization for high-dimensional sparse estimation problems. Statistical Science. (2012). 27(4): 576-593.

\appendix

\section*{Appendix A: Statistical Consistency for Logistic Loss}
  Statistical consistency results for the weight estimates based on penalty functions in (\ref{B3}) and (\ref{B4}) can be obtained by modifying slightly existing theoretical results (Tarigan and van de Geer, 2006; Meier, van de Geer and Buehlmann, 2008) in the literature. We focus on only Laplace and Arctan penalties. Consider the class of logistic classifiers, \[ \mathcal{F} = \{ \eta_{\beta}: \mathbb{R}^{p} \rightarrow \mathbb{R}\; |\; \eta_{\beta}(\mathbf{x}) =  \mathbf{x}^{T}\mathbf{\beta},\; \mathbf{x} \in \mathcal{X} \subseteq \mathbb{R}^{p},\;  \mathbf{\beta} \in \mathcal{B} \subseteq \mathbb{R}^{p} \}. \] Classification is done based on the sign of of the function $f_{\beta}: \mathbb{R}^{p} \rightarrow \mathbb{R}$ defined as \[ f_{\beta}(\mathbf{x}) = \frac{1}{1 + \exp (-\eta_{\beta}(\mathbf{x}))} - \frac{1}{2}. \] Here \[  \pi_{\beta}(\mathbf{x}) = \frac{1}{1 + \exp (-\eta_{\beta}(\mathbf{x}))} = \frac{\exp (\eta_{\beta}(\mathbf{x}))}{1 + \exp (\eta_{\beta}(\mathbf{x}))} \] denotes the class probability. If there are $K > 2$ classes, then the class probability for the $k^{th}$ class may be modeled as \[\displaystyle \frac{\exp (\eta_{\beta_{k}}(\mathbf{x}))}{\sum_{m=1}^{K}\exp(\eta_{\beta_{m}}(\mathbf{x}))}, \] but for simplicity, we just focus on binary classification.

  We assume that $\mathcal{X}$ is endowed with a probability measure $\nu$ and let $|| \cdot ||_{p,\nu}$ be the $L_{p}(\nu)$ norm ($1 \leq p < \infty$). Denote $\Sigma = \mathbb{E}_{\nu}(\mathbf{x}^{T}\mathbf{x})$. Design matrix $ \mathbf{X} = [\mathbf{x}_{1}^{T}, \ldots, \mathbf{x}_{n}^{T} ]^{T}$ consists of $n$ copies of $\mathbf{x}$. The empirical logistic (also known as cross-entropy) loss is \[ R_{n}(\eta_{\beta}) = \frac{1}{n} \sum_{i=1}^{n} l(y_{i}, \eta_{\beta}(\mathbf{x}_{i})), \; \mathrm{where}\;l(y_{i}, \eta_{\beta}(\mathbf{x}_{i})) = - [y_{i}\pi_{\beta}(\mathbf{x}_{i}) + (1 - y_{i})(1 - \pi_{\beta}(\mathbf{x}_{i}))] \] and theoretical loss \[ R(\eta_{\beta}) = \mathbb{E}(R_{n}(\eta_{\beta})). \] Let \[ \eta^{*} = \argminB_{\mathrm{all} \; \eta} R(\eta). \]

  We assume the following three conditions given in Tarigan and van de Geer, 2006.

  (C1): There exists constants $\sigma > 0$ and $\kappa \geq 1$, such that for all $\eta \in \mathcal{F}$, \[ R(\eta) - R(\eta^{*}) \geq \frac{||\eta - \eta^{*}||_{1,\nu}^{\kappa}}{\sigma^{\kappa}}. \]

  (C2): The smallest eigenvalue $\rho^{2}$ of $\Sigma$ is non-zero.

  (C3): $ \max\limits_{1\leq k\leq p}|| \mathbf{x}e_{k} ||_{\infty} \leq \sqrt{\frac{n}{\log n} }, \; \max\limits_{1\leq k\leq p}|| \mathbf{x}e_{k} ||_{2} < \infty, \; p < n^{D}$ for some $D$. Here $e_{k}$ denotes the unit vector with 1 as the $k^{th}$ element and 0's elsewhere.\\

The following theorem holds for $p_{\lambda}(\beta)$, where $p_{\lambda}(\beta)$ equals either the Laplace penalty function given in (3) or the arctan penalty function given in (4).\\

\textbf{Theorem.} Assume conditions C1 to C3 hold and that $|| \eta_{\beta} ||_{\infty} \leq K, \; \forall \eta_{\beta} \in \mathcal{F}$. Then for universal constants $c, c_{1}$, \[ \mathbb{P}(R(\hat{\eta}_{n}) - R(\eta^{*}) > \varepsilon_{n}) \leq \frac{c_{1}}{n^{2}}, \] where \[ \hat{\eta}_{n} = \argminB_{\eta_{\beta} \in \mathcal{F}} \{R_{n}(\eta_{\beta}) + p_{\hat{\lambda}_{n}}(\beta)) \},\,\hat{\lambda}_{n} = c(\hat{C}_{n} \vee 4)DK^{2}\sqrt{\frac{\log(n)}{n}}, \,\hat{C}_{n} = \frac{1}{n}\max\limits_{1\leq k\leq p}||\mathbf{x}e_{k} ||_{2}^{2}; \] \[ \varepsilon_{n} = (1 + 4\delta)\inf \left\{ R(\eta_{\beta}) - R(\eta^{*}) + V_{n}(N(\beta)) + 2\lambda_{n}K\sqrt{\frac{\log(n)}{n}}: \eta_{\beta} \in \mathcal{F} \right\},\, \delta \in (0, 0.5]; \] \begin{equation} V_{n}(N) = 2\delta^{-1/(2\kappa - 1)}(18\sigma\lambda_{n}^{2}L N DK/\rho^{2})^{\kappa/(2\kappa-1)}, \; \kappa \geq 1;\; N(\beta) = \#(\mathrm{nonzero\;elements\;in}\; \beta). \label{app1} \end{equation} The constant $L$ in (\ref{app1}) depends on the penalty function: $L = (\log(\varepsilon))^{2}$ for the Laplace penalty and $L = (2\gamma/\pi)^{2}$ for the arctan penalty.

\textbf{Proof:} We give only a sketch of the proof, as the proof is the same as the lengthy proof given in Tarigan and van de Geer, 2006, with only minor differences. First of all note that the only difference in the statement of the above theorem from the statement of the Theorem 1 in Tarigan and van de Geer, 2006 is the constant $L$ inserted in (\ref{app1}).

Although the loss function used in Tarigan and van de Geer, 2006, was Hinge-loss function, the steps in their proof holds true for logistic loss also (-actually it becomes easier-) as pointed out in Meier, van de Geer and Buehlmann, 2008. Thus, the only difference in the steps in the proof that we need to focus are those corresponding to the penalty functions. Their theorem is stated for the Lasso penalty. The triangle inequality satisfied by the Lasso penalty is used in certain steps of the proof. But, since both Laplace and arctan penalties are subadditive ( - concave in the positive real line, with $p_{\lambda}(0) = 0$ - ), those steps hold true for these two penalties as well.

The only other step we need to focus is Lemma 5.2 in their proof, where the key inequality used is \begin{equation} \displaystyle  \left( \sum_{t = 1}^{T} \abs{\beta_{t}} \right)^{2} \leq T\sum_{t = 1}^{T} \beta_{t}^{2}. \label{app2} \end{equation} Instead, we use the inequalities \begin{equation}  \displaystyle  \left( \sum_{t = 1}^{T} \left( 1 - \varepsilon^{\abs{\beta_{t}}} \right) \right)^{2} \leq LT\sum_{t = 1}^{T} \beta_{t}^{2},\; \mathrm{where}\; L = (\log(\varepsilon))^{2} \label{app3} \end{equation}  and \begin{equation}  \displaystyle  \left( \sum_{t = 1}^{T} \left( \frac{2}{\pi} \mathrm{arctan} (\gamma\abs{\beta_{t}}) \right) \right)^{2} \leq LT\sum_{t = 1}^{T} \beta_{t}^{2},\; \mathrm{where}\; L = \left(\frac{2\gamma}{\pi} \right)^{2}. \label{app4}\end{equation} Inequality in (\ref{app3}) follows from the inequality in (\ref{app2}) and the fact that \[ 1 - \varepsilon^{x} \leq -(\log \varepsilon)x,\; \mathrm{for}\; x \geq 0, \] which follows easily by considering the function $f(x) = -(\log \varepsilon)x - 1 + \varepsilon^{x}$ and noting that $f(0) = 0$ and \[f'(x) = -\log \varepsilon + \varepsilon^{x}\log \varepsilon = (-\log \varepsilon)(1 - \varepsilon^{x}) \geq 0, \; \mathrm{for}\; x \geq 0\; \mathrm{and}\; \varepsilon \in (0,1).\] Inequality in (\ref{app4}) follows from the inequality in (\ref{app2}) and the right inequality in Lemma 3.2. \hfill\BlackBox

\section*{Appendix B: Detailed Tables of Test Error Rates (Supplementary Material)}


DNN analysis was repeated for multiple seed values. The test error rates presented in section 6 was the median of the test error rates from all seed values. Detailed results (that is, test error rates for each grid point and seed value) used for compiling the summarized table in section 6 are presented below.

\scriptsize
\subsection*{MNIST}

\begin{center}
\begin{tabular}{ | l || c | c | c | c | c | c | c | c | c | c | c | }
    \hline
    \multicolumn{12}{|c|}{MNIST results with no regularization}   \\ \hline
    Seed & 1 & 2 & 3 & 4 & 5 & 6 & 7 & 8 & 9 & 10 & Median \\ \hline
    Error & 1.69	& 2.54	& 1.76	& 2.38	& 1.70	& 1.98	& 2.34	& 1.72	& 2.56 & 	1.76 & \color{blue} 1.87\color{black} \\ \hline
    \end{tabular}
\end{center}

\begin{center}
    \begin{tabular}{ | c | c | c | c | c |   }
    \hline
    \multicolumn{5}{|c|}{MNIST results with Lasso regularization}   \\ \hline
    $\log(\lambda)$ & Seed $=$ 1 & Seed $=$ 2 & Seed $=$ 3 & Median \\ \hline
-4.00 &	1.98 &  1.98 &	1.69 &	1.98 \\ \hline
-4.20 &	1.55 &	1.55 &	1.52 &	1.55 \\ \hline
-4.40 &	1.45 &	1.36 &	1.32 &	1.36 \\ \hline
-4.60 &	1.37 &	1.34 &	1.41 &	1.37 \\ \hline
-4.80 &	1.28 &	1.40 &	1.29 &	1.29 \\ \hline
-5.00 &	1.44 &	1.27 &	1.24 &	1.27 \\ \hline
-5.20 &	1.16 &	1.34 &	1.24 &	\color{blue} 1.24\color{black} \\ \hline
-5.40 &	1.45 &	1.36 &	1.35 &	1.36 \\ \hline
-5.60 &	1.36 &	1.42 &	1.34 &	1.36 \\ \hline
-5.80 &	1.24 &	1.32 &	1.21 &	1.24 \\ \hline
-6.00 &	1.26 &	1.31 &	1.24 &	1.26 \\ \hline
-6.20 &	1.72 &	2.09 &	1.87 &	1.87 \\ \hline
-6.40 &	2.16 &	1.78 &	2.31 &	2.16 \\ \hline
-6.60 &	2.14 &	1.97 &	1.41 &	1.97 \\ \hline
-6.80 &	1.97 &	1.84 &	2.45 &	1.97 \\ \hline
-7.00 &	1.84 &	1.97 &	2.03 &	1.97 \\ \hline
    \end{tabular}
\end{center}

\begin{center}
    \begin{tabular}{ | c | c | c | c | c |   }
    \hline
    \multicolumn{5}{|c|}{MNIST results with $L_{2}$ regularization}   \\ \hline
    $\log(\lambda)$ & Seed $=$ 1 & Seed $=$ 2 & Seed $=$ 3 & Median \\ \hline
-4.00 &	 1.26 &	1.23 &	1.22 & \color{blue} 1.23\color{black}\\ \hline
-4.20 &	 1.36 &	1.25 &	1.22 &	1.25\\ \hline
-4.40 &	 1.45 &	1.33 &	1.29 &	1.33\\ \hline
-4.60 &	 1.31 &	1.28 &	1.34 &	1.31\\ \hline
-4.80 &	 1.33 &	1.35 &	1.26 &	1.33\\ \hline
-5.00 &	 1.90 &	1.30 &	1.27 &	1.30\\ \hline
-5.20 &	 1.70 &	2.45 &	2.41 &	2.41\\ \hline
-5.40 &	 2.21 &	2.16 &	2.43 &	2.21\\ \hline
-5.60 &	 2.09 &	2.23 &	2.02 &	2.09\\ \hline
-5.80 &	 2.43 &	1.71 &	1.92 &	1.92\\ \hline
-6.00 &	 2.14 &	2.07 &	2.15 &	2.14\\ \hline
-6.20 &	 1.99 &	1.96 &	2.38 &	1.99\\ \hline
-6.40 &	 1.83 &	2.05 &	2.03 &	2.03\\ \hline
-6.60 &	 2.02 &	1.38 &	1.47 &	1.47\\ \hline
-6.80 &	 1.88 &	2.40 &	1.83 &	1.88\\ \hline
-7.00 &	 2.19 &	1.40 &	1.97 &	1.97\\ \hline
    \end{tabular}
\end{center}

\begin{center}
    \begin{tabular}{ | c | c | c | c | c |    }
    \hline
    \multicolumn{5}{|c|}{MNIST results with Laplace ($\varepsilon$ = 1e-07) regularization}   \\ \hline
     $\log (\lambda)$ & Seed = 1 & Seed = 2 & Seed = 3 & Median \\ \hline
-4.00 &	6.05 &	5.78 &	6.43 &	6.05 \\ \hline
-4.20 &	3.07 &	2.70 &	2.93 &	2.93 \\ \hline
-4.40 &	2.29 &	2.37 &	2.41 &	2.37 \\ \hline
-4.60 &	2.33 &	2.10 &	1.93 &	2.10 \\ \hline
-4.80 &	2.04 &	1.77 &	1.94 &	1.94 \\ \hline
-5.00 &	1.81 &	1.86 &	1.79 &	1.81 \\ \hline
-5.20 &	1.46 &	1.53 &	1.39 &	1.46 \\ \hline
-5.40 &	1.35 &	1.43 &	1.45 &	1.43 \\ \hline
-5.60 &	1.18 &	1.36 &	1.23 &	\color{blue} 1.23\color{blue} \\ \hline
-5.80 &	1.47 &	1.42 &	1.45 &	1.45 \\ \hline
-6.00 &	1.46 &	1.46 &	1.35 &	1.46 \\ \hline
-6.20 &	1.45 &	1.40 &	1.41 &	1.41 \\ \hline
-6.40 &	1.33 &	1.35 &  1.35 &	1.35 \\ \hline
-6.60 &	1.33 &	1.36 &  1.30 &	1.33 \\ \hline
-6.80 &	1.25 &	1.28 &  1.34 &	1.28 \\ \hline
-7.00 &	1.19 &	1.40 &  1.33 &	1.33 \\ \hline

    \end{tabular}
\end{center}

\begin{center}
    \begin{tabular}{ | c | c | c | c | c |  }
    \hline
    \multicolumn{5}{|c|}{MNIST results with Arctan ($\gamma$ = 1) regularization}   \\ \hline
    $\log(\lambda)$ & Seed $=$ 1 & Seed $=$ 2 & Seed $=$ 3 & Median \\ \hline
-4.00 &	1.70 &	1.57 &	1.55 &	1.57  \\ \hline
-4.20 &	1.35 &	1.41 &	1.22 &	1.35  \\ \hline
-4.40 &	1.33 &	1.41 &	1.33 &	1.33  \\ \hline
-4.60 &	1.26 &	1.31 &	1.16 &	\color{blue} 1.26\color{black} \\ \hline
-4.80 &	1.25 &	1.33 &	1.38 &	1.33  \\ \hline
-5.00 &	1.35 &	1.35 &	1.24 &	1.35  \\ \hline
-5.20 &	1.34 &	1.38 &	1.32 &	1.34  \\ \hline
-5.40 &	1.24 &	1.42 &	1.29 &	1.29  \\ \hline
-5.60 &	1.35 &	1.34 &	1.22 &	1.34  \\ \hline
-5.80 &	1.45 &	1.40 &	1.31 &	1.40  \\ \hline
-6.00 &	1.82 &	1.80 &	2.22 & 	1.82  \\ \hline
-6.20 &	2.12 &	1.75 &	1.69 & 	1.75  \\ \hline
-6.40 &	2.16 &	2.05 &	1.61 & 	2.05  \\ \hline
-6.60 &	2.29 &	1.74 &	2.05 & 	2.05  \\ \hline
-6.80 &	1.85 &	1.97 &	1.88 &	1.88  \\ \hline
-7.00 &	2.01 &	1.79 &	1.80 & 	1.80  \\ \hline
    \end{tabular}
\end{center}

\begin{center}
    \begin{tabular}{ | c | c | c | c | c |  }
    \hline
    \multicolumn{5}{|c|}{MNIST results with Arctan ($\gamma$ = 100) regularization}   \\ \hline
    $\log(\lambda)$ & Seed $=$ 1 & Seed $=$ 2 & Seed $=$ 3 & Median \\ \hline
-4.00 &	 3.43 & 3.65 & 5.71	& 3.65		 \\ \hline
-4.20 &	 1.95 & 3.62 & 1.75	& 1.95		 \\ \hline
-4.40 &	 2.57 & 1.66 & 2.45	& 2.45		 \\ \hline
-4.60 &	 1.94 & 2.64 & 1.36	& 1.94		 \\ \hline
-4.80 &	 1.51 & 1.43 & 1.49	& 1.49		 \\ \hline
-5.00 &	 1.47 & 1.50 & 2.11	& 1.50		 \\ \hline
-5.20 &	 1.41 & 1.45 & 2.02	& 1.45		 \\ \hline
-5.40 &	 1.51 & 1.51 & 1.50	& 1.51		 \\ \hline
-5.60 &	 1.44 & 1.39 & 1.57	& 1.44		 \\ \hline
-5.80 &	 1.88 & 1.74 & 1.57	& 1.74		 \\ \hline
-6.00 &	 1.30 & 1.34 & 1.31	& 1.31		 \\ \hline
-6.20 &	 1.35 & 1.31 & 1.31	& 1.31		 \\ \hline
-6.40 &	 1.38 & 1.30 & 1.29	& 1.30		 \\ \hline
-6.60 &	 1.33 & 1.25 & 1.25	& \color{blue} 1.25\color{black}	\\ \hline
-6.80 &	 1.39 & 1.30 & 1.27	& 1.30		 \\ \hline
-7.00 &	 1.34 & 1.27 & 1.34	& 1.34		 \\ \hline
    \end{tabular}
\end{center}

\begin{center}
    \begin{tabular}{ | c | c | c | c | c |   }
    \hline
    \multicolumn{5}{|c|}{MNIST results with SCAD ($a$ = 3.7) regularization}   \\ \hline
    $\log(\lambda)$ & Seed $=$ 1 & Seed $=$ 2 & Seed $=$ 3 & Median \\ \hline					
-4.00 &  1.95	& 2.29 & 1.92 & 1.95 \\ \hline					
-4.20 &	 2.07	& 2.05 & 1.78 & 2.05 \\ \hline
-4.40 &	 1.94	& 2.29 & 1.92 & 1.94 \\ \hline
-4.60 &	 2.08	& 1.99 & 1.67 & 1.99 \\ \hline
-4.80 &	 1.70	& 1.80 & 1.95 & \color{blue}1.80\color{black} \\ \hline
-5.00 &	 2.34	& 2.26 & 2.08 & 2.26 \\ \hline
-5.20 &	 2.28	& 2.11 & 1.94 & 2.11 \\ \hline
-5.40 &	 2.29	& 2.01 & 2.10 & 2.10 \\ \hline
-5.60 &	 2.28	& 2.39 & 1.83 & 2.28 \\ \hline
-5.80 &	 2.02	& 1.88 & 1.58 & 1.88 \\ \hline
-6.00 &	 2.32	& 2.06 & 1.94 & 2.06 \\ \hline
-6.20 &	 2.12	& 1.74 & 1.86 & 1.86 \\ \hline
-6.40 &	 2.20	& 1.97 & 1.94 & 1.97 \\ \hline
-6.60 &	 1.89	& 1.61 & 2.12 & 1.89 \\ \hline
-6.80 &	 2.00	& 1.44 & 1.96 & 1.96 \\ \hline
-7.00 &	 2.07	& 2.08 & 2.00 & 2.07 \\ \hline
    \end{tabular}
\end{center}

\begin{center}
    \begin{tabular}{ | c | c | c | c | c |   }
    \hline
    \multicolumn{5}{|c|}{MNIST results with MCP ($b$ = 1.5) regularization}   \\ \hline
    $\log(\lambda)$ & Seed $=$ 1 & Seed $=$ 2 & Seed $=$ 3 & Median \\ \hline					
-4.00 & 1.72 & 1.97	& 2.09 & 1.97  \\ \hline					
-4.20 &	2.19 & 1.89	& 1.84 & 1.89  \\ \hline
-4.40 &	2.16 & 1.88	& 1.91 & 1.91  \\ \hline
-4.60 &	1.94 & 1.92	& 1.85 & 1.92  \\ \hline
-4.80 &	1.82 & 2.38	& 1.95 & 1.95  \\ \hline
-5.00 &	2.10 & 1.93	& 2.34 & 2.10  \\ \hline
-5.20 &	1.76 & 2.28	& 1.80 & 1.80  \\ \hline
-5.40 &	2.09 & 1.33	& 1.60 & \color{blue}1.60\color{black}  \\ \hline
-5.60 &	1.80 & 2.12	& 1.90 & 1.90  \\ \hline
-5.80 &	2.01 & 1.83	& 2.12 & 2.01  \\ \hline
-6.00 &	2.24 & 2.20	& 1.65 & 2.20  \\ \hline
-6.20 &	2.16 & 1.46	& 1.74 & 1.74  \\ \hline
-6.40 &	2.00 & 1.67	& 2.15 & 2.00  \\ \hline
-6.60 &	2.02 & 1.78	& 1.83 & 1.83  \\ \hline
-6.80 &	2.02 & 1.96	& 1.89 & 1.96  \\ \hline
-7.00 &	2.02 & 2.08	& 2.56 & 2.08  \\ \hline
    \end{tabular}
\end{center}

\begin{center}
    \begin{tabular}{ | c | c | c | c | c |   }
    \hline
    \multicolumn{5}{|c|}{MNIST results with MCP ($b$ = 5) regularization}   \\ \hline
    $\log(\lambda)$ & Seed $=$ 1 & Seed $=$ 2 & Seed $=$ 3 & Median \\ \hline					
-4.00 & 2.05 &  1.65 &	1.98 &	1.98  \\ \hline					
-4.20 &	1.96 &	2.40 &	2.08 &	2.08  \\ \hline
-4.40 &	2.03 &	1.75 &	2.11 &	2.03  \\ \hline
-4.60 &	2.16 &	2.10 &	1.86 &	2.10  \\ \hline
-4.80 &	1.82 &	1.96 &	2.14 &	1.96  \\ \hline
-5.00 &	2.01 &	2.05 &	2.03 &	2.03  \\ \hline
-5.20 &	2.01 &	2.24 &	2.09 &	2.09  \\ \hline
-5.40 &	2.19 &	2.12 &	1.96 &	2.12  \\ \hline
-5.60 &	2.17 &	1.94 &	2.02 &	2.02  \\ \hline
-5.80 &	2.29 &	1.67 &	1.64 &	\color{blue} 1.67 \color{black}  \\ \hline
-6.00 &	2.13 &	2.15 &	2.40 &	2.15  \\ \hline
-6.20 &	1.97 &	2.00 &	1.80 &	1.97  \\ \hline
-6.40 &	1.71 &	1.76 &	2.09 &	1.76  \\ \hline
-6.60 &	2.03 &	2.13 &	2.10 &	2.10  \\ \hline
-6.80 &	2.01 &	2.02 &	1.78 &	2.01  \\ \hline
-7.00 &	2.02 &	2.08 &	2.00 &	2.02  \\ \hline
    \end{tabular}
\end{center}

\begin{center}
    \begin{tabular}{ | c | c | c | c | c |   }
    \hline
    \multicolumn{5}{|c|}{MNIST results with MCP ($b$ = 20) regularization}   \\ \hline
    $\log(\lambda)$ & Seed $=$ 1 & Seed $=$ 2 & Seed $=$ 3 & Median \\ \hline					
-4.00 & 2.37 &	2.01 &	1.95 &	2.01  \\ \hline					
-4.20 &	2.06 &	2.38 &	2.07 &	2.07  \\ \hline
-4.40 &	2.11 &	1.64 &	1.65 &	\color{blue} 1.65 \color{black}  \\ \hline
-4.60 &	1.95 &	1.64 &	1.98 &	1.95  \\ \hline
-4.80 &	1.65 &	1.94 &	2.07 &	1.94  \\ \hline
-5.00 &	1.65 &	2.22 &	1.91 &	1.91  \\ \hline
-5.20 &	2.01 &	1.74 &	2.00 &	2.00  \\ \hline
-5.40 &	2.25 &	2.72 &	2.06 &	2.25  \\ \hline
-5.60 &	1.80 &	2.39 &	1.72 &	1.80  \\ \hline
-5.80 &	2.31 &	2.04 &	1.88 &	2.04  \\ \hline
-6.00 &	1.91 &	2.13 &	1.89 &	1.91  \\ \hline
-6.20 &	1.81 &	2.10 &	1.82 &	1.82  \\ \hline
-6.40 &	2.28 &	2.40 &	2.02 &	2.28  \\ \hline
-6.60 &	1.94 &	2.36 &	1.87 &	1.94  \\ \hline
-6.80 &	1.80 &	2.18 &	2.52 &	2.18  \\ \hline
-7.00 &	2.02 &	1.67 &	2.37 &	2.02  \\ \hline
    \end{tabular}
\end{center}

\subsection*{FMNIST}

\begin{center}
    \begin{tabular}{ | l || c | c | c | c | c | c | c | c | c | c | c | }
    \hline
    \multicolumn{12}{|c|}{FMNIST results with no regularization}   \\ \hline
    Seed & 1 & 2 & 3 & 4 & 5 & 6 & 7 & 8 & 9 & 10 & Median \\ \hline
    Error & 11.91 & 11.75 & 11.70 & 12.80 & 11.97 & 12.47 & 11.91 & 12.28 & 12.38 & 11.40 & \color{blue} 11.94\color{black} \\ \hline
    \end{tabular}
\end{center}

\begin{center}
    \begin{tabular}{ | c | c | c | c | c |   }
    \hline
    \multicolumn{5}{|c|}{FMNIST results with Lasso regularization}   \\ \hline
    $\log(\lambda)$ & Seed $=$ 1 & Seed $=$ 2 & Seed $=$ 3 & Median \\ \hline
-4.00 &	12.23 & 12.55 & 14.30 & 12.55 \\ \hline
-4.20 &	11.90 & 11.44 & 11.75 & 11.75 \\ \hline
-4.40 &	14.09 & 10.79 & 10.92 & 10.92 \\ \hline
-4.60 &	11.45 & 10.55 & 13.68 & 11.45 \\ \hline
-4.80 &	10.63 & 13.42 & 10.58 & 10.63 \\ \hline
-5.00 &	10.06 & 10.65 &  9.97 & \color{blue} 10.06\color{black} \\ \hline
-5.20 &	11.04 & 10.86 & 11.15 & 11.04 \\ \hline
-5.40 &	10.53 & 10.07 &  9.91 & 10.07 \\ \hline
-5.60 &	10.46 & 10.29 & 10.34 & 10.34 \\ \hline
-5.80 &	11.09 & 11.66 & 11.50 & 11.50 \\ \hline
-6.00 &	11.97 & 11.48 & 11.71 & 11.71 \\ \hline
-6.20 &	11.88 & 11.81 & 11.70 & 11.81 \\ \hline
-6.40 &	11.57 & 11.78 & 11.71 & 11.71 \\ \hline
-6.60 &	12.00 & 11.79 & 11.81 & 11.81 \\ \hline
-6.80 &	12.10 & 11.81 & 11.35 & 11.81 \\ \hline
-7.00 &	12.04 & 11.59 & 11.76 & 11.76 \\ \hline
    \end{tabular}
\end{center}

\begin{center}
    \begin{tabular}{ | c | c | c | c | c |   }
    \hline
    \multicolumn{5}{|c|}{FMNIST results with $L_{2}$ regularization}   \\ \hline
    $\log(\lambda)$ & Seed $=$ 1 & Seed $=$ 2 & Seed $=$ 3 & Median \\ \hline
-4.00 &	10.59 & 10.32 & 11.45 & 10.59 \\ \hline
-4.20 &	 9.95 & 10.27 & 10.76 & 10.27 \\ \hline
-4.40 &	10.15 & 10.20 & 10.15 & \color{blue} 10.15\color{black} \\ \hline
-4.60 &	10.45 & 10.65 & 10.15 & 10.45 \\ \hline
-4.80 &	12.00 & 12.13 & 11.73 & 12.00 \\ \hline
-5.00 &	11.21 & 12.44 & 11.77 & 11.77 \\ \hline
-5.20 &	11.66 & 11.75 & 11.53 & 11.66 \\ \hline
-5.40 &	12.20 & 11.56 & 11.65 & 11.65 \\ \hline
-5.60 &	12.01 & 12.47 & 12.05 & 12.05 \\ \hline
-5.80 &	12.18 & 11.49 & 12.44 & 12.18 \\ \hline
-6.00 &	12.05 & 11.40 & 12.22 & 12.05 \\ \hline
-6.20 &	12.25 & 11.54 & 11.54 & 11.54 \\ \hline
-6.40 &	12.17 & 12.32 & 11.77 & 12.17 \\ \hline
-6.60 &	11.36 & 11.78 & 11.62 & 11.62 \\ \hline
-6.80 &	11.93 & 11.66 & 11.38 & 11.66 \\ \hline
-7.00 &	12.13 & 12.44 & 11.61 & 12.13 \\ \hline
    \end{tabular}
\end{center}

\begin{center}
    \begin{tabular}{ | c | c | c | c | c |    }
    \hline
    \multicolumn{5}{|c|}{FMNIST results with Laplace ($\varepsilon$ = 1e-07) regularization}   \\ \hline
     $\log (\lambda)$ & Seed = 1 & Seed = 2 & Seed = 3 & Median \\ \hline
-4.00 &	18.37 &	19.90 & 19.70 & 19.70 \\ \hline
-4.20 &	14.39 & 14.98 & 16.91 & 14.98 \\ \hline
-4.40 &	16.89 & 16.78 & 14.02 & 16.78 \\ \hline
-4.60 &	15.51 & 15.07 & 13.01 & 15.07 \\ \hline
-4.80 &	12.51 & 12.48 & 12.29 & 12.48 \\ \hline
-5.00 &	11.82 & 11.83 & 11.93 & 11.83 \\ \hline
-5.20 &	13.88 & 11.22 & 11.05 & 11.22 \\ \hline
-5.40 &	13.22 & 10.25 & 10.53 & 10.53 \\ \hline
-5.60 &	10.54 & 12.81 & 10.19 & 10.54 \\ \hline
-5.80 &	12.52 &  9.99 & 12.16 & 12.16 \\ \hline
-6.00 &	12.04 & 10.21 & 10.92 & 10.92 \\ \hline
-6.20 &  9.68 &  9.98 & 10.52 &  \color{blue} 9.98\color{black} \\ \hline
-6.40 &	11.23 & 10.11 & 10.01 & 10.11 \\ \hline
-6.60 &	10.06 & 10.00 & 10.14 & 10.06 \\ \hline
-6.80 &	11.49 & 10.24 & 10.87 & 10.87 \\ \hline
-7.00 &	10.69 & 11.43 & 10.88 & 10.88 \\ \hline

    \end{tabular}
\end{center}

\begin{center}
    \begin{tabular}{ | c | c | c | c | c |  }
    \hline
    \multicolumn{5}{|c|}{FMNIST results with Arctan ($\gamma$ = 1) regularization}   \\ \hline
    $\log(\lambda)$ & Seed $=$ 1 & Seed $=$ 2 & Seed $=$ 3 & Median \\ \hline
-4.00 &	 14.07 & 11.60 & 11.74 & 11.74		 \\ \hline
-4.20 &	 11.44 & 10.98 & 11.01 & 11.01		 \\ \hline
-4.40 &	 10.94 & 10.35 & 11.19 & 10.94		 \\ \hline
-4.60 &	 10.62 &  9.88 & 10.58 & 10.58		 \\ \hline
-4.80 &	  9.84 &  9.87 & 10.24 &  \color{blue} 9.87\color{black} \\ \hline
-5.00 &	 10.47 & 10.21 & 10.32 & 10.32		 \\ \hline
-5.20 &	  9.95 &  9.91 & 11.14 &  9.95		 \\ \hline
-5.40 &	 10.51 & 10.29 &  9.82 & 10.29		 \\ \hline
-5.60 &	 11.08 & 10.44 & 12.55 & 11.08		 \\ \hline
-5.80 &	 11.50 & 11.66 & 11.71 & 11.66		 \\ \hline
-6.00 &	 12.22 & 13.05 & 11.43 & 12.22		 \\ \hline
-6.20 &	 12.09 & 11.69 & 11.93 & 11.93		 \\ \hline
-6.40 &	 12.09 & 12.67 & 11.27 & 12.09		 \\ \hline
-6.60 &	 11.15 & 11.39 & 11.53 & 11.39	     \\ \hline
-6.80 &	 12.20 & 12.88 & 11.22 & 12.20		 \\ \hline
-7.00 &	 11.46 & 11.47 & 11.18 & 11.46		 \\ \hline
    \end{tabular}
\end{center}

\begin{center}
    \begin{tabular}{ | c | c | c | c | c |  }
    \hline
    \multicolumn{5}{|c|}{FMNIST results with Arctan ($\gamma$ = 100) regularization}   \\ \hline
    $\log(\lambda)$ & Seed $=$ 1 & Seed $=$ 2 & Seed $=$ 3 & Median \\ \hline
-4.00 &	 15.54 & 15.84 & 18.20 & 15.84		 \\ \hline
-4.20 &	 15.03 & 16.42 & 14.62 & 15.03		 \\ \hline
-4.40 &	 14.08 & 16.67 & 13.38 & 14.08		 \\ \hline
-4.60 &	 14.53 & 15.96 & 13.75 & 14.53		 \\ \hline
-4.80 &	 17.16 & 15.39 & 12.85 & 15.39		 \\ \hline
-5.00 &	 11.77 & 14.65 & 13.92 & 13.92		 \\ \hline
-5.20 &	 11.56 & 12.39 & 13.84 & 12.39		 \\ \hline
-5.40 &	  9.68 & 14.27 & 13.27 & 13.27		 \\ \hline
-5.60 &	  9.69 & 12.23 & 11.65 & 11.65		 \\ \hline
-5.80 &	  9.77 & 11.00 & 11.90 & 11.00		 \\ \hline
-6.00 &	  9.80 & 10.13 & 10.05 & 10.05		 \\ \hline
-6.20 &	 12.05 &  9.73 & 10.27 & 10.27		 \\ \hline
-6.40 &	 10.30 & 10.33 & 10.21 & 10.30		 \\ \hline
-6.60 &	  9.93 & 10.61 &  9.93 &  \color{blue} 9.93\color{black} \\ \hline
-6.80 &	 11.03 & 11.10 & 12.74 & 11.10		 \\ \hline
-7.00 &	 11.47 & 11.46 & 11.00 & 11.46		 \\ \hline
    \end{tabular}
\end{center}

\begin{center}
    \begin{tabular}{ | c | c | c | c | c |   }
    \hline
    \multicolumn{5}{|c|}{FMNIST results with SCAD ($a$ = 3.7) regularization}   \\ \hline
    $\log(\lambda)$ & Seed $=$ 1 & Seed $=$ 2 & Seed $=$ 3 & Median \\ \hline					
-4.00 & 12.12 & 12.55 & 11.48 & 12.12  \\ \hline					
-4.20 &	11.98 & 12.54 & 11.75 & 11.98  \\ \hline
-4.40 &	11.96 & 11.78 & 11.28 & 11.78  \\ \hline
-4.60 &	11.83 & 11.31 & 11.85 & 11.83  \\ \hline
-4.80 &	12.14 & 11.74 & 11.76 & 11.76  \\ \hline
-5.00 &	12.15 & 11.79 & 11.65 & 11.79  \\ \hline
-5.20 &	11.56 & 12.53 & 11.78 & 11.78  \\ \hline
-5.40 &	11.90 & 11.37 & 11.82 & 11.82  \\ \hline
-5.60 &	11.83 & 12.75 & 11.79 & 11.83  \\ \hline
-5.80 &	11.72 & 11.49 & 11.50 & 11.50  \\ \hline
-6.00 &	11.92 & 12.59 & 11.86 & 11.92  \\ \hline
-6.20 &	11.80 & 11.68 & 11.52 & 11.68  \\ \hline
-6.40 &	12.18 & 13.08 & 12.28 & 12.28  \\ \hline
-6.60 &	12.19 & 11.74 & 11.60 & 11.74  \\ \hline
-6.80 &	12.06 & 11.68 & 11.78 & 11.78  \\ \hline
-7.00 &	11.07 & 11.87 & 11.45 & \color{blue} 11.45\color{black}  \\ \hline
    \end{tabular}
\end{center}

\begin{center}
    \begin{tabular}{ | c | c | c | c | c |   }
    \hline
    \multicolumn{5}{|c|}{FMNIST results with MCP ($b$ = 1.5) regularization}   \\ \hline
    $\log(\lambda)$ & Seed $=$ 1 & Seed $=$ 2 & Seed $=$ 3 & Median \\ \hline					
-4.00 & 12.15   & 11.62	& 11.17	& 11.62  \\ \hline					
-4.20 &	11.99	& 12.77	& 11.38	& 11.99  \\ \hline
-4.40 &	11.52	& 11.39	& 11.95	& 11.52  \\ \hline
-4.60 &	11.66	& 11.34	& 12.15	& 11.66  \\ \hline
-4.80 &	11.05	& 11.67	& 11.81	& 11.67  \\ \hline
-5.00 &	11.39	& 12.29	& 11.49	& 11.49  \\ \hline
-5.20 &	12.25	& 11.41	& 11.59	& 11.59  \\ \hline
-5.40 &	12.24	& 11.82	& 11.99	& 11.99  \\ \hline
-5.60 &	12.06	& 11.76	& 12.00	& 12.00  \\ \hline
-5.80 &	11.44	& 11.77	& 11.50	& 11.50  \\ \hline
-6.00 &	11.18	& 12.79	& 11.80	& 11.80  \\ \hline
-6.20 &	11.44	& 11.68	& 12.27	& 11.68  \\ \hline
-6.40 &	12.51	& 12.95	& 11.90	& 12.51  \\ \hline
-6.60 &	12.03	& 11.84	& 11.94	& 11.94  \\ \hline
-6.80 &	11.95	& 11.68	& 12.14	& 11.95  \\ \hline
-7.00 &	10.96	& 11.39	& 11.45	& \color{blue} 11.39\color{black}  \\ \hline
    \end{tabular}
\end{center}

\begin{center}
    \begin{tabular}{ | c | c | c | c | c |   }
    \hline
    \multicolumn{5}{|c|}{FMNIST results with MCP ($b$ = 5) regularization}   \\ \hline
    $\log(\lambda)$ & Seed $=$ 1 & Seed $=$ 2 & Seed $=$ 3 & Median \\ \hline					
-4.00 & 12.19	& 12.78	& 11.38 & 12.19  \\ \hline					
-4.20 &	11.19	& 12.67	& 11.60 & 11.60  \\ \hline
-4.40 &	12.02	& 11.72	& 12.33	& 12.02  \\ \hline
-4.60 &	12.03	& 11.41	& 11.37	& 11.41  \\ \hline
-4.80 &	11.79	& 11.64	& 11.74	& 11.74  \\ \hline
-5.00 &	11.80	& 11.75	& 11.82	& 11.80  \\ \hline
-5.20 &	12.10	& 12.35	& 12.02	& 12.10  \\ \hline
-5.40 &	11.77	& 13.21	& 12.18	& 12.18  \\ \hline
-5.60 &	12.34	& 12.86	& 10.67	& 12.34  \\ \hline
-5.80 &	12.12	& 11.65	& 11.95	& 11.95  \\ \hline
-6.00 &	11.24	& 11.39	& 11.40	& 11.39  \\ \hline
-6.20 &	12.07	& 12.42	& 11.81	& 12.07  \\ \hline
-6.40 &	12.28	& 12.87	& 11.78	& 12.28  \\ \hline
-6.60 &	11.32	& 11.57	& 12.18	& 11.57  \\ \hline
-6.80 &	11.93	& 11.34	& 11.85	& 11.85  \\ \hline
-7.00 &	11.07	& 11.39	& 11.45	& \color{blue} 11.39\color{black}  \\ \hline
    \end{tabular}
\end{center}

\begin{center}
    \begin{tabular}{ | c | c | c | c | c |   }
    \hline
    \multicolumn{5}{|c|}{FMNIST results with MCP ($b$ = 20) regularization}   \\ \hline
    $\log(\lambda)$ & Seed $=$ 1 & Seed $=$ 2 & Seed $=$ 3 & Median \\ \hline					
-4.00 &  11.86 &	11.60 & 11.78 & 11.78 \\ \hline					
-4.20 &	 12.19 &	11.53 & 10.92 & 11.53 \\ \hline
-4.40 &	 11.65 &	11.30 & 11.50 & 11.50 \\ \hline
-4.60 &	 11.60 &	11.75 & 11.71 & 11.71 \\ \hline
-4.80 &	 11.37 &	13.05 & 11.64 & 11.64 \\ \hline
-5.00 &	 12.06 &	11.93 & 11.99 & 11.99 \\ \hline
-5.20 &	 12.25 &	11.51 & 12.01 & 12.01 \\ \hline
-5.40 &	 11.76 &	12.50 & 11.89 & 11.89 \\ \hline
-5.60 &	 12.14 &	11.82 & 11.54 & 11.82 \\ \hline
-5.80 &	 11.80 &	11.46 & 12.12 & 11.80 \\ \hline
-6.00 &	 11.74 &	11.87 & 11.64 & 11.74 \\ \hline
-6.20 &	 11.37 &	11.86 & 11.69 & 11.69 \\ \hline
-6.40 &	 11.55 &	11.24 & 11.67 & 11.55 \\ \hline
-6.60 &	 11.26 &	11.40 & 11.33 & \color{blue} 11.33\color{black} \\ \hline
-6.80 &	 11.23 &	11.89 & 12.13 & 11.89 \\ \hline
-7.00 &	 11.80 &    11.39 & 11.69 & 11.69 \\ \hline
    \end{tabular}
\end{center}

\subsection*{Reuters}

\begin{center}
    \begin{tabular}{ | l || c | c | c | c | c | c | c | c | c | c | c | }
    \hline
    \multicolumn{12}{|c|}{RCV1 results with no regularization}   \\ \hline
    Seed & 1 & 2 & 3 & 4 & 5 & 6 & 7 & 8 & 9 & 10 & Median \\ \hline
    Error & 13.81 & 14.62 & 14.37 & 14.38 & 16.87 & 15.63 & 15.85 & 16.26 & 14.69 & 14.11 & \color{blue} 14.66\color{black}\\ \hline
    \end{tabular}
\end{center}

\begin{center}
    \begin{tabular}{ | c | c | c | c | c |   }
    \hline
    \multicolumn{5}{|c|}{RCV1 results with Lasso regularization}   \\ \hline
    $\log(\lambda)$ & Seed $=$ 1 & Seed $=$ 2 & Seed $=$ 3 & Median \\ \hline
-4.00 &	14.51 & 14.38 & 14.68 & 14.51 \\ \hline
-4.20 &	15.86 & 14.79 & 15.80 & 15.80 \\ \hline
-4.40 &	15.08 & 15.49 & 14.87 & 15.08 \\ \hline
-4.60 &	13.88 & 13.82 & 14.54 & 13.88 \\ \hline
-4.80 &	13.80 & 13.36 & 13.41 & 13.41 \\ \hline
-5.00 &	13.39 & 13.72 & 12.90 & 13.39 \\ \hline
-5.20 &	12.79 & 12.97 & 13.29 & \color{blue} 12.97\color{black} \\ \hline
-5.40 &	14.26 & 12.99 & 13.87 & 13.87 \\ \hline
-5.60 &	14.67 & 13.46 & 14.69 & 14.67 \\ \hline
-5.80 &	14.98 & 14.72 & 13.95 & 14.72 \\ \hline
-6.00 &	15.04 & 14.68 & 15.12 & 15.04 \\ \hline
-6.20 &	14.49 & 15.07 & 15.28 & 15.07 \\ \hline
-6.40 &	14.17 & 15.00 & 15.14 & 15.00 \\ \hline
-6.60 &	15.10 & 14.38 & 14.81 & 14.81 \\ \hline
-6.80 &	14.49 & 14.85 & 14.15 & 14.49 \\ \hline
-7.00 &	14.77 & 14.55 & 13.79 & 14.55 \\ \hline
    \end{tabular}
\end{center}

\begin{center}
    \begin{tabular}{ | c | c | c | c | c |   }
    \hline
    \multicolumn{5}{|c|}{RCV1 results with $L_{2}$ regularization}   \\ \hline
    $\log(\lambda)$ & Seed $=$ 1 & Seed $=$ 2 & Seed $=$ 3 & Median \\ \hline
-4.00 &	13.09 &	13.77 & 14.21 & \color{blue} 13.77\color{black} \\ \hline
-4.20 &	14.21 &	14.33 &	14.05 &	14.21 \\ \hline
-4.40 &	15.81 &	14.21 &	14.53 &	14.53 \\ \hline
-4.60 &	14.33 &	15.41 &	15.29 &	15.29 \\ \hline
-4.80 &	15.93 &	15.67 &	16.05 &	15.93 \\ \hline
-5.00 &	14.25 &	16.17 &	16.01 &	16.01 \\ \hline
-5.20 &	15.89 &	13.89 &	14.69 &	14.69 \\ \hline
-5.40 &	15.57 &	15.77 &	15.57 &	15.57 \\ \hline
-5.60 &	14.89 &	13.53 &	13.81 &	13.81 \\ \hline
-5.80 &	15.73 &	15.05 &	14.77 &	15.05 \\ \hline
-6.00 &	14.77 &	15.17 &	14.61 &	14.77 \\ \hline
-6.20 &	14.61 &	15.37 &	15.61 &	15.37 \\ \hline
-6.40 &	13.41 &	15.33 &	14.57 &	14.57 \\ \hline
-6.60 &	15.37 &	15.69 &	14.33 &	15.37 \\ \hline
-6.80 &	13.53 &	15.81 &	14.65 &	14.65 \\ \hline
-7.00 &	15.25 &	15.85 &	15.49 &	15.49 \\ \hline
    \end{tabular}
\end{center}

\begin{center}
    \begin{tabular}{ | c | c | c | c | c |    }
    \hline
    \multicolumn{5}{|c|}{RCV1 results with Laplace ($\varepsilon$ = 1e-07) regularization}   \\ \hline
     $\log (\lambda)$ & Seed = 1 & Seed = 2 & Seed = 3 & Median \\ \hline
-4.00 &	 25.96	& 26.27	& 25.63	& 25.96 \\ \hline
-4.20 &	 20.50	& 20.29	& 25.54	& 20.50 \\ \hline
-4.40 &	 17.82	& 18.38	& 18.46	& 18.38 \\ \hline
-4.60 &	 15.64	& 15.92	& 15.85	& 15.85 \\ \hline
-4.80 &	 15.22	& 15.99	& 15.55	& 15.55 \\ \hline
-5.00 &	 14.95	& 14.18	& 16.53	& 14.95 \\ \hline
-5.20 &	 13.89	& 14.41	& 14.79	& 14.41 \\ \hline
-5.40 &	 13.32	& 13.98	& 14.64	& 13.98 \\ \hline
-5.60 &	 13.16	& 13.55	& 13.81	& 13.55 \\ \hline
-5.80 &	 12.87	& 13.54	& 12.94	& \color{blue} 12.94\color{black} \\ \hline
-6.00 &	 13.29	& 14.95	& 13.09	& 13.29 \\ \hline
-6.20 &  13.19	& 13.57	& 12.97	& 13.19 \\ \hline
-6.40 &	 13.58	& 12.57	& 13.09	& 13.09 \\ \hline
-6.60 &	 15.39	& 14.27	& 13.30	& 14.27 \\ \hline
-6.80 &	 14.09	& 15.21	& 13.88	& 14.09 \\ \hline
-7.00 &	 14.41	& 14.62	& 14.04	& 14.41 \\ \hline

    \end{tabular}
\end{center}

\begin{center}
    \begin{tabular}{ | c | c | c | c | c |  }
    \hline
    \multicolumn{5}{|c|}{RCV1 results with Arctan ($\gamma$ = 1) regularization}   \\ \hline
    $\log(\lambda)$ & Seed $=$ 1 & Seed $=$ 2 & Seed $=$ 3 & Median \\ \hline
-4.00 &	 16.25	& 14.49	& 16.13	 & 16.13		 \\ \hline
-4.20 &	 14.57	& 14.53	& 15.29	 & 14.57		 \\ \hline
-4.40 &	 14.57	& 13.69	& 14.89	 & 14.57		 \\ \hline
-4.60 &	 13.73	& 13.57	& 13.85	 & 13.73		 \\ \hline
-4.80 &	 13.73	& 13.65	& 13.53	 & 13.65        \\ \hline
-5.00 &	 13.13	& 13.41	& 13.53	 & \color{blue} 13.41\color{black}\\ \hline
-5.20 &	 15.25	& 13.81	& 14.21	 & 14.21		 \\ \hline
-5.40 &	 14.69	& 14.53	& 13.93	 & 14.53		 \\ \hline
-5.60 &	 15.09	& 15.05	& 16.13	 & 15.09		 \\ \hline
-5.80 &	 14.41	& 15.77	& 15.85	 & 15.77		 \\ \hline
-6.00 &	 15.89	& 15.17	& 15.57	 & 15.57		 \\ \hline
-6.20 &	 15.41	& 14.69	& 15.53	 & 15.41		 \\ \hline
-6.40 &	 14.29	& 14.21	& 17.73	 & 14.29		 \\ \hline
-6.60 &	 16.65	& 14.93	& 15.01	 & 15.01	     \\ \hline
-6.80 &	 16.09	& 16.41	& 15.13	 & 16.09		 \\ \hline
-7.00 &	 16.05	& 17.25	& 15.21	 & 16.05		 \\ \hline
    \end{tabular}
\end{center}

\begin{center}
    \begin{tabular}{ | c | c | c | c | c |  }
    \hline
    \multicolumn{5}{|c|}{RCV1 results with Arctan ($\gamma$ = 100) regularization}   \\ \hline
    $\log(\lambda)$ & Seed $=$ 1 & Seed $=$ 2 & Seed $=$ 3 & Median \\ \hline
-4.00 &	 34.99	& 26.94	& 52.80	& 34.99		 \\ \hline
-4.20 &	 27.02	& 21.66	& 27.94	& 27.02		 \\ \hline
-4.40 &	 17.61	& 18.65	& 20.58	& 18.65		 \\ \hline
-4.60 &	 16.93	& 17.49	& 17.93	& 17.49		 \\ \hline
-4.80 &	 17.17	& 16.93	& 18.21	& 17.17      \\ \hline
-5.00 &	 16.65	& 15.65	& 16.69	& 16.65		 \\ \hline
-5.20 &	 16.09	& 14.85	& 16.41	& 16.09 	 \\ \hline
-5.40 &	 15.13	& 14.77	& 14.97	& 14.97		 \\ \hline
-5.60 &	 14.73	& 14.29	& 13.61	& 14.29		 \\ \hline
-5.80 &	 13.97	& 13.57	& 14.45	& 13.97		 \\ \hline
-6.00 &	 13.93	& 13.57	& 14.25	& 13.93		 \\ \hline
-6.20 &	 13.89	& 13.49	& 14.33	& 13.89		 \\ \hline
-6.40 &	 15.01	& 13.89	& 15.77	& 15.01		 \\ \hline
-6.60 &	 15.61	& 14.33	& 14.09	& 14.33	     \\ \hline
-6.80 &	 14.69	& 13.81	& 13.65	& \color{blue} 13.81\color{black}\\ \hline
-7.00 &	 15.93	& 13.69	& 14.41	& 14.41		 \\ \hline
    \end{tabular}
\end{center}

\begin{center}
    \begin{tabular}{ | c | c | c | c | c |   }
    \hline
    \multicolumn{5}{|c|}{RCV1 results with SCAD ($a$ = 3.7) regularization}   \\ \hline
    $\log(\lambda)$ & Seed $=$ 1 & Seed $=$ 2 & Seed $=$ 3 & Median \\ \hline					
-4.00 & 14.68 &	14.32 &	15.08 &	14.68  \\ \hline					
-4.20 &	14.56 &	14.04 &	14.64 &	14.56  \\ \hline
-4.40 &	15.44 &	15.76 &	16.00 &	15.76  \\ \hline
-4.60 &	15.20 &	14.00 &	14.80 &	14.80  \\ \hline
-4.80 &	14.24 &	14.40 &	16.28 &	14.40  \\ \hline
-5.00 &	14.68 &	14.32 &	15.92 &	14.68  \\ \hline
-5.20 &	15.84 &	12.56 &	15.00 &	15.00  \\ \hline
-5.40 &	15.20 &	14.52 &	15.84 &	15.20  \\ \hline
-5.60 &	15.60 &	13.76 &	15.08 &	15.08  \\ \hline
-5.80 &	15.12 &	14.92 &	15.28 &	15.12  \\ \hline
-6.00 &	14.72 &	13.44 &	14.48 &	14.48  \\ \hline
-6.20 &	13.96 &	13.88 &	13.96 &	\color{blue} 13.96 \color{black} \\ \hline
-6.40 &	14.60 &	14.80 &	15.12 &	14.80  \\ \hline
-6.60 &	14.76 &	14.64 &	14.60 &	14.64  \\ \hline
-6.80 &	14.60 &	13.64 &	14.28 & 14.28  \\ \hline
-7.00 &	14.48 &	13.92 &	14.56 & 14.56  \\ \hline
    \end{tabular}
\end{center}

\begin{center}
    \begin{tabular}{ | c | c | c | c | c |   }
    \hline
    \multicolumn{5}{|c|}{RCV1 results with MCP ($b$ = 1.5) regularization}   \\ \hline
    $\log(\lambda)$ & Seed $=$ 1 & Seed $=$ 2 & Seed $=$ 3 & Median \\ \hline					
-4.00 &  14.33	& 13.93	& 13.85	& 13.85 \\ \hline					
-4.20 &	 14.13	& 12.97	& 14.09	& 14.09 \\ \hline
-4.40 &	 14.33	& 12.53	& 13.93	& 13.93 \\ \hline
-4.60 &	 13.65	& 14.81	& 13.81	& 13.81 \\ \hline
-4.80 &	 13.85	& 13.41	& 13.29	& 13.41 \\ \hline
-5.00 &	 14.53	& 15.05	& 13.69	& 14.53 \\ \hline
-5.20 &	 13.33	& 13.29	& 13.97	& \color{blue} 13.33\color{black} \\ \hline
-5.40 &	 14.77	& 13.69	& 14.41	& 14.41 \\ \hline
-5.60 &	 13.33	& 13.77	& 13.73	& 13.73 \\ \hline
-5.80 &	 14.17	& 14.49	& 14.69	& 14.49 \\ \hline
-6.00 &	 14.93	& 13.53	& 13.29	& 13.53 \\ \hline
-6.20 &	 14.89	& 14.37	& 13.85	& 14.37 \\ \hline
-6.40 &	 13.69	& 12.57	& 13.49	& 13.49 \\ \hline
-6.60 &	 14.97	& 13.57	& 13.73	& 13.73 \\ \hline
-6.80 &	 13.81	& 13.93	& 15.41	& 13.93 \\ \hline
-7.00 &	 13.81	& 13.49	& 14.25	& 13.81 \\ \hline
    \end{tabular}
\end{center}

\begin{center}
    \begin{tabular}{ | c | c | c | c | c |   }
    \hline
    \multicolumn{5}{|c|}{RCV1 results with MCP ($b$ = 5) regularization}   \\ \hline
    $\log(\lambda)$ & Seed $=$ 1 & Seed $=$ 2 & Seed $=$ 3 & Median \\ \hline					
-4.00 & 15.96 &	14.92 &	15.56 &	15.56  \\ \hline					
-4.20 &	15.56 &	14.16 &	16.92 &	15.56  \\ \hline
-4.40 &	15.08 &	14.40 &	15.40 &	15.08  \\ \hline
-4.60 &	14.48 &	13.96 &	15.20 &	14.48  \\ \hline
-4.80 &	15.60 &	14.44 &	14.12 &	\color{blue} 14.44\color{black}  \\ \hline
-5.00 &	15.12 &	15.52 &	15.12 &	15.12  \\ \hline
-5.20 &	15.32 &	16.12 &	15.16 &	15.32  \\ \hline
-5.40 &	16.24 &	15.32 &	15.68 &	15.68  \\ \hline
-5.60 &	15.24 &	14.92 &	14.68 &	14.92  \\ \hline
-5.80 &	15.28 &	14.60 &	15.24 &	15.24  \\ \hline
-6.00 &	15.88 &	15.52 &	15.44 &	15.52  \\ \hline
-6.20 &	16.20 &	13.80 &	15.76 &	15.76  \\ \hline
-6.40 &	15.60 &	15.44 &	15.64 &	15.60  \\ \hline
-6.60 &	15.20 &	14.52 &	15.24 &	15.20  \\ \hline
-6.80 &	15.04 &	13.04 &	16.24 &	15.04  \\ \hline
-7.00 &	15.04 &	15.64 &	15.24 &	15.24  \\ \hline
    \end{tabular}
\end{center}

\begin{center}
    \begin{tabular}{ | c | c | c | c | c |   }
    \hline
    \multicolumn{5}{|c|}{RCV1 results with MCP ($b$ = 20) regularization}   \\ \hline
    $\log(\lambda)$ & Seed $=$ 1 & Seed $=$ 2 & Seed $=$ 3 & Median \\ \hline					
-4.00 & 15.08 &	14.76 & 15.00 &	15.00  \\ \hline					
-4.20 &	15.96 &	14.92 &	16.44 &	15.96  \\ \hline
-4.40 &	14.92 &	13.60 &	16.92 &	14.92  \\ \hline
-4.60 &	15.20 &	15.24 &	16.24 &	15.24  \\ \hline
-4.80 &	14.56 &	14.08 &	14.72 &	14.56  \\ \hline
-5.00 &	14.80 &	14.36 &	15.56 &	14.80  \\ \hline
-5.20 &	14.88 &	14.60 &	16.84 &	14.88  \\ \hline
-5.40 &	14.36 &	14.28 &	15.44 &	\color{blue} 14.36\color{black}  \\ \hline
-5.60 &	16.00 &	14.96 &	15.32 &	15.32  \\ \hline
-5.80 &	15.40 &	13.76 &	14.56 &	14.56  \\ \hline
-6.00 &	16.20 &	13.88 &	15.88 &	15.88  \\ \hline
-6.20 &	15.56 &	15.40 &	16.56 &	15.56  \\ \hline
-6.40 &	15.60 &	13.80 &	15.88 &	15.60  \\ \hline
-6.60 &	15.20 &	14.48 &	15.24 &	15.20  \\ \hline
-6.80 &	15.04 &	16.00 &	16.24 &	16.00  \\ \hline
-7.00 &	15.04 &	15.64 &	15.24 &	15.24  \\ \hline
    \end{tabular}
\end{center}

\subsection*{CIFAR-10}

\begin{center}
    \begin{tabular}{ | l || c | c | c | c | c | c | c | c | c | c | c | }
    \hline
    \multicolumn{12}{|c|}{CIFAR-10 results with no regularization}   \\ \hline
    Seed & 1 & 2 & 3 & 4 & 5 & 6 & 7 & 8 & 9 & 10 & Median \\ \hline
    Error & 20.75 &	19.89	& 17.99 &	19.64 &	18.55	& 19.67	& 19.25 &	19.52 &	18.86 &	19.63 & \color{blue}19.58\color{black} \\ \hline
    \end{tabular}
\end{center}

\begin{center}
    \begin{tabular}{ | c | c | c | c | c |   }
    \hline
    \multicolumn{5}{|c|}{CIFAR-10 results with Lasso regularization}   \\ \hline
    $\log(\lambda)$ & Seed $=$ 1 & Seed $=$ 2 & Seed $=$ 3 & Median \\ \hline
-4.00 &	19.10 &	18.08 &	17.99 &	18.08 \\ \hline
-4.20 &	19.45 &	17.93 &	18.54 &	18.54 \\ \hline
-4.40 &	18.88 &	18.55 &	18.22 &	18.55 \\ \hline
-4.60 &	15.29 &	17.83 &	14.85 &	15.29 \\ \hline
-4.80 &	15.45 &	14.92 &	14.96 &	14.96 \\ \hline
-5.00 &	14.76 &	15.14 &	14.91 &	14.91 \\ \hline
-5.20 &	14.85 &	15.15 &	14.51 &	14.85 \\ \hline
-5.40 &	14.71 &	14.48 &	14.99 &	\color{blue}14.71 \color{black} \\ \hline
-5.60 &	14.93 &	15.07 &	15.04 &	15.04 \\ \hline
-5.80 &	14.78 &	15.11 &	14.83 &	14.83 \\ \hline
-6.00 &	19.03 &	19.08 &	21.69 &	19.08 \\ \hline
-6.20 &	19.33 &	19.60 &	19.50 &	19.50 \\ \hline
-6.40 &	20.46 &	20.59 &	18.57 &	20.46 \\ \hline
-6.60 &	18.61 &	18.98 &	19.10 &	18.98 \\ \hline
-6.80 &	20.43 &	19.57 &	20.25 &	20.25 \\ \hline
-7.00 &	19.59 &	19.30 &	18.19 &	19.30 \\ \hline
    \end{tabular}
\end{center}

\begin{center}
    \begin{tabular}{ | c | c | c | c | c |   }
    \hline
    \multicolumn{5}{|c|}{CIFAR-10 results with $L_{2}$ regularization}   \\ \hline
    $\log(\lambda)$ & Seed $=$ 1 & Seed $=$ 2 & Seed $=$ 3 & Median \\ \hline
-4.00 &	14.57 &	14.37 &	14.56 &	14.56 \\ \hline
-4.20 &	16.10 &	14.94 &	14.37 &	14.94 \\ \hline
-4.40 &	15.04 &	14.32 &	14.29 &	\color{blue}14.32\color{black} \\ \hline
-4.60 &	20.41 &	19.59 &	18.87 &	19.59 \\ \hline
-4.80 &	19.50 &	19.89 &	21.35 &	19.89 \\ \hline
-5.00 &	20.54 &	19.01 &	21.77 &	20.54 \\ \hline
-5.20 &	18.21 &	18.65 &	19.63 &	18.65 \\ \hline
-5.40 &	19.73 &	19.38 &	18.53 &	19.38 \\ \hline
-5.60 &	18.81 &	20.71 &	18.45 &	18.81 \\ \hline
-5.80 &	20.70 &	20.84 &	20.48 &	20.70 \\ \hline
-6.00 &	20.05 &	19.64 &	20.08 &	20.05 \\ \hline
-6.20 &	19.16 &	19.38 &	20.02 &	19.38 \\ \hline
-6.40 &	19.64 &	19.00 &	18.16 &	19.00 \\ \hline
-6.60 &	18.65 &	19.24 &	19.06 &	19.06 \\ \hline
-6.80 &	20.36 &	19.50 &	19.69 & 19.69 \\ \hline
-7.00 &	19.12 & 18.55 & 18.54 &	18.55 \\ \hline		
    \end{tabular}
\end{center}

\begin{center}
    \begin{tabular}{ | c | c | c | c | c |   }
    \hline
    \multicolumn{5}{|c|}{CIFAR-10 results with Laplace ($\varepsilon$ = 1e-07) regularization}   \\ \hline
    $\log(\lambda)$ & Seed $=$ 1 & Seed $=$ 2 & Seed $=$ 3 & Median \\ \hline
-4.00 &	36.67 &	36.94 &	38.14 &	36.94 \\ \hline
-4.20 &	35.25 &	36.81 &	38.33 &	36.81 \\ \hline
-4.40 &	31.02 &	32.12 &	30.46 &	31.02 \\ \hline
-4.60 &	30.59 &	31.78 &	30.56 &	30.59 \\ \hline
-4.80 &	30.28 &	27.70 &	30.47 &	30.28 \\ \hline
-5.00 &	26.48 &	28.52 &	28.81 &	28.52 \\ \hline
-5.20 &	26.51 &	22.58 &	24.81 &	24.81 \\ \hline
-5.40 &	19.91 &	22.60 &	19.22 &	19.91 \\ \hline
-5.60 &	19.51 &	19.53 &	21.48 &	19.53 \\ \hline
-5.80 &	20.43 &	20.95 &	19.28 &	20.43 \\ \hline
-6.00 &	20.13 &	19.94 &	18.48 &	19.94 \\ \hline
-6.20 &	20.40 &	20.66 &	20.03 &	20.40 \\ \hline
-6.40 &	18.42 &	20.09 &	20.04 &	20.04 \\ \hline
-6.60 &	19.63 &	19.45 &	19.72 &	19.63 \\ \hline
-6.80 &	19.09 &	19.09 &	19.29 &	19.09 \\ \hline
-7.00 &	18.74 &	18.87 &	19.89 &	\color{blue}18.87\color{black} \\ \hline
    \end{tabular}
\end{center}

\begin{center}
    \begin{tabular}{ | c | c | c | c | c |   }
    \hline
    \multicolumn{5}{|c|}{CIFAR-10 results with Arctan ($\gamma$ = 1) regularization}   \\ \hline
    $\log(\lambda)$ & Seed $=$ 1 & Seed $=$ 2 & Seed $=$ 3 & Median \\ \hline
-4.00 &	19.77 &	18.75 &	19.78 &	19.77	 \\ \hline
-4.20 &	17.81 &	18.35 &	17.75 &	17.81	 \\ \hline
-4.40 &	18.49 &	18.63 &	18.16 &	18.49	 \\ \hline
-4.60 &	14.91 &	15.89 &	14.92 &	14.92	 \\ \hline
-4.80 &	15.07 &	15.10 &	15.06 &	15.07	 \\ \hline
-5.00 &	16.22 &	15.94 &	14.72 &	15.94	 \\ \hline
-5.20 &	14.87 &	15.09 &	14.94 &	14.94	 \\ \hline
-5.40 &	15.03 &	14.99 &	14.75 &	14.99	 \\ \hline
-5.60 &	14.76 &	14.56 &	14.61 &	\color{blue}14.61\color{black}	 \\ \hline
-5.80 &	20.09 &	18.88 &	18.34 &	18.88	 \\ \hline
-6.00 &	18.53 &	18.84 &	19.78 &	18.84	 \\ \hline
-6.20 &	18.66 &	20.84 &	21.76 &	20.84	 \\ \hline
-6.40 &	19.04 &	19.75 &	19.29 &	19.29	 \\ \hline
-6.60 &	19.04 &	19.78 &	21.47 &	19.78	 \\ \hline
-6.80 &	19.96 &	19.36 &	19.17 &	19.36	 \\ \hline
-7.00 &	20.78 &	18.55 &	19.24 &	19.24	 \\ \hline
    \end{tabular}
\end{center}

\begin{center}
    \begin{tabular}{ | c | c | c | c | c |   }
    \hline
    \multicolumn{5}{|c|}{CIFAR-10 results with Arctan ($\gamma$ = 100) regularization}   \\ \hline
    $\log(\lambda)$ & Seed $=$ 1 & Seed $=$ 2 & Seed $=$ 3 & Median \\ \hline
-4.00 &	29.90 &	30.10 &	31.01 &	30.10	 \\ \hline
-4.20 &	25.24 &	26.54 &	30.00 &	26.54	 \\ \hline
-4.40 &	24.76 &	22.21 &	24.31 &	24.31	 \\ \hline
-4.60 &	15.26 &	15.52 &	14.91 &	15.26	 \\ \hline
-4.80 &	15.07 &	15.25 &	15.44 &	15.25	 \\ \hline
-5.00 &	14.86 &	14.62 &	14.39 &	14.62	 \\ \hline
-5.20 &	14.94 &	14.78 &	14.88 &	14.88	 \\ \hline
-5.40 &	14.65 &	14.83 &	14.67 &	14.67	 \\ \hline
-5.60 &	14.56 &	14.71 &	14.85 &	14.71	 \\ \hline
-5.80 &	14.74 &	14.57 &	14.75 &	14.74	 \\ \hline
-6.00 &	14.51 &	14.63 &	14.97 &	14.63	 \\ \hline
-6.20 &	14.39 &	14.89 &	14.39 &	\color{blue}14.39\color{black}	 \\ \hline
-6.40 &	14.72 &	15.04 &	14.78 &	14.78	 \\ \hline
-6.60 &	14.88 &	14.88 &	14.58 &	14.88	 \\ \hline
-6.80 &	14.72 &	14.66 &	14.86 &	14.72	 \\ \hline
-7.00 &	14.80 &	14.94 &	14.81 &	14.81	 \\ \hline
    \end{tabular}
\end{center}

\subsection*{CIFAR-100}

\begin{center}
    \begin{tabular}{ | l || c | c | c | c | c | c | c | c | c | c | c | }
    \hline
    \multicolumn{12}{|c|}{CIFAR-100 results with no regularization}   \\ \hline
    Seed & 1 & 2 & 3 & 4 & 5 & 6 & 7 & 8 & 9 & 10 & Median \\ \hline
    Error & 51.35 &	52.12 &	50.71 &	50.74 &	49.16 &	51.44 &	50.24 &	50.57 &	50.45 &	50.55 & \color{blue}50.64\color{black} \\ \hline
    \end{tabular}
\end{center}

\begin{center}
    \begin{tabular}{ | c | c | c | c | c |   }
    \hline
    \multicolumn{5}{|c|}{CIFAR-100 results with Lasso regularization}   \\ \hline
    $\log(\lambda)$ & Seed $=$ 1 & Seed $=$ 2 & Seed $=$ 3 & Median \\ \hline
-4.00 &	51.38 &	50.17 &	51.80 &	51.38   \\ \hline
-4.20 &	50.37 &	49.53 &	49.72 &	49.72   \\ \hline
-4.40 &	49.99 &	51.09 &	48.68 &	49.99   \\ \hline
-4.60 &	43.75 &	45.31 &	43.20 &	43.75   \\ \hline
-4.80 &	43.93 &	42.82 &	43.64 &	43.64   \\ \hline
-5.00 &	43.44 &	43.38 &	43.13 &	43.38   \\ \hline
-5.20 &	44.16 &	43.22 &	42.98 &	\color{blue}43.22\color{black}   \\ \hline
-5.40 &	49.59 &	50.76 &	48.78 &	49.59   \\ \hline
-5.60 &	52.01 &	51.26 &	50.75 &	51.26   \\ \hline
-5.80 &	50.31 &	49.29 &	49.25 &	49.29   \\ \hline
-6.00 &	50.23 &	48.17 &	50.62 &	50.23   \\ \hline
-6.20 &	50.68 &	52.20 &	50.25 &	50.68   \\ \hline
-6.40 &	50.32 &	52.16 &	50.89 &	50.89   \\ \hline
-6.60 &	49.54 &	52.57 &	49.87 &	49.87   \\ \hline
-6.80 &	48.99 &	52.03 &	48.39 &	48.99   \\ \hline
-7.00 &	51.78 &	52.67 &	50.98 &	51.78   \\ \hline
    \end{tabular}
\end{center}

\begin{center}
    \begin{tabular}{ | c | c | c | c | c |   }
    \hline
    \multicolumn{5}{|c|}{CIFAR-100 results with $L_{2}$ regularization}   \\ \hline
    $\log(\lambda)$ & Seed $=$ 1 & Seed $=$ 2 & Seed $=$ 3 & Median \\ \hline
-4.00 &	40.78 &	41.11 &	42.76 &	\color{blue}41.11\color{black} 	 \\ \hline
-4.20 &	51.73 &	51.82 &	48.68 &	51.73 	 \\ \hline
-4.40 &	50.16 &	52.11 &	51.31 &	51.31 	 \\ \hline
-4.60 &	50.25 &	49.42 &	50.21 &	50.25 	 \\ \hline
-4.80 &	50.86 &	52.12 &	51.22 &	51.22 	 \\ \hline
-5.00 &	50.33 &	52.82 &	50.95 &	50.95 	 \\ \hline
-5.20 &	50.33 &	50.19 &	49.35 &	50.19 	 \\ \hline
-5.40 &	50.87 &	51.85 &	50.89 &	50.89 	 \\ \hline
-5.60 &	51.85 &	52.40 &	51.30 &	51.85 	 \\ \hline
-5.80 &	50.67 &	51.75 &	51.23 &	51.23 	 \\ \hline
-6.00 &	51.57 &	52.97 &	50.54 &	51.57 	 \\ \hline
-6.20 &	50.68 &	51.81 &	50.66 &	50.68 	 \\ \hline
-6.40 &	50.83 &	51.84 &	50.99 &	50.99 	 \\ \hline
-6.60 &	50.98 &	52.49 &	49.76 &	50.98 	 \\ \hline
-6.80 &	50.54 &	52.19 &	51.22 &	51.22 	 \\ \hline
-7.00 &	50.82 &	52.33 &	51.21 &	51.21 	 \\ \hline
    \end{tabular}
\end{center}

\begin{center}
    \begin{tabular}{ | c | c | c | c | c |   }
    \hline
    \multicolumn{5}{|c|}{CIFAR-100 results with Laplace ($\varepsilon$ = 1e-07) regularization}   \\ \hline
    $\log(\lambda)$ & Seed $=$ 1 & Seed $=$ 2 & Seed $=$ 3 & Median \\ \hline
-4.00 &	73.57 &	73.51 &	74.02 &	73.57 	 \\ \hline
-4.20 &	73.72 &	74.01 &	74.55 &	74.01 	 \\ \hline
-4.40 &	73.80 &	74.63 &	64.32 &	73.80 	 \\ \hline
-4.60 &	68.15 &	66.15 &	64.73 &	66.15 	 \\ \hline
-4.80 &	66.69 &	65.23 &	64.68 &	65.23 	 \\ \hline
-5.00 &	53.47 &	55.81 &	56.32 &	55.81 	 \\ \hline
-5.20 &	53.27 &	52.79 &	54.23 &	53.27 	 \\ \hline
-5.40 &	50.98 &	51.57 &	51.48 &	51.48 	 \\ \hline
-5.60 &	51.93 &	52.77 &	53.38 &	52.77 	 \\ \hline
-5.80 &	48.91 &	49.79 &	51.27 &	\color{blue}49.79\color{black} 	 \\ \hline
-6.00 &	50.21 &	52.83 &	50.72 &	50.72 	 \\ \hline
-6.20 &	50.96 &	51.81 &	50.65 &	50.96 	 \\ \hline
-6.40 &	51.12 &	52.15 &	50.87 &	51.12 	 \\ \hline
-6.60 &	50.75 &	52.74 &	48.97 &	50.75 	 \\ \hline
-6.80 &	51.37 &	50.73 &	49.88 &	50.73 	 \\ \hline
-7.00 &	51.06 &	50.30 &	49.89 &	50.30 	 \\ \hline
    \end{tabular}
\end{center}

\begin{center}
    \begin{tabular}{ | c | c | c | c | c |   }
    \hline
    \multicolumn{5}{|c|}{CIFAR-100 results with Arctan ($\gamma$ = 1) regularization}   \\ \hline
    $\log(\lambda)$ & Seed $=$ 1 & Seed $=$ 2 & Seed $=$ 3 & Median \\ \hline
-4.00 &	50.68 &	49.71 &	49.60 &	49.71 	 \\ \hline
-4.20 &	49.62 &	50.05 &	48.58 &	49.62 	 \\ \hline
-4.40 &	44.38 &	44.23 &	49.30 &	44.38 	 \\ \hline
-4.60 &	43.84 &	44.27 &	44.03 &	44.03 	 \\ \hline
-4.80 &	43.48 &	43.58 &	43.62 &	\color{blue}43.58\color{black} 	 \\ \hline
-5.00 &	43.83 &	43.99 &	43.61 &	43.83 	 \\ \hline
-5.20 &	48.77 &	48.87 &	48.94 &	48.87 	 \\ \hline
-5.40 &	50.42 &	50.53 &	50.71 &	50.53 	 \\ \hline
-5.60 &	48.24 &	52.84 &	49.67 &	49.67 	 \\ \hline
-5.80 &	50.27 &	51.88 &	51.53 &	51.53 	 \\ \hline
-6.00 &	50.47 &	52.44 &	50.61 &	50.61 	 \\ \hline
-6.20 &	50.78 &	52.14 &	51.35 &	51.35 	 \\ \hline
-6.40 &	51.05 &	52.29 &	50.00 &	51.05 	 \\ \hline
-6.60 &	51.20 &	52.49 &	51.26 &	51.26 	 \\ \hline
-6.80 &	50.50 &	51.71 &	51.92 &	51.71 	 \\ \hline
-7.00 &	49.92 & 50.23 & 51.04 & 50.23 	 \\ \hline
    \end{tabular}
\end{center}

\begin{center}
    \begin{tabular}{ | c | c | c | c | c |   }
    \hline
    \multicolumn{5}{|c|}{CIFAR-100 results with Arctan ($\gamma$ = 100) regularization}   \\ \hline
    $\log(\lambda)$ & Seed $=$ 1 & Seed $=$ 2 & Seed $=$ 3 & Median \\ \hline
-4.00 &	74.88 &	75.41 &	77.23 &	75.41 	 \\ \hline
-4.20 &	67.95 &	69.79 &	66.62 &	67.95 	 \\ \hline
-4.40 &	59.89 &	58.35 &	58.27 &	58.35 	 \\ \hline
-4.60 &	45.50 &	46.49 &	46.04 &	46.04 	 \\ \hline
-4.80 &	44.78 &	44.94 &	44.95 &	44.94 	 \\ \hline
-5.00 &	44.53 &	43.64 &	43.47 &	43.64 	 \\ \hline
-5.20 &	43.33 &	43.25 &	44.06 &	\color{blue}43.33\color{black} 	 \\ \hline
-5.40 &	43.37 &	43.41 &	43.24 &	43.37 	 \\ \hline
-5.60 &	43.09 &	43.95 &	43.88 &	43.88 	 \\ \hline
-5.80 &	44.34 &	45.06 &	44.13 &	44.34 	 \\ \hline
-6.00 &	44.18 &	46.16 &	45.06 &	45.06 	 \\ \hline
-6.20 &	45.10 &	44.86 &	46.11 &	45.10 	 \\ \hline
-6.40 &	53.41 &	50.97 &	50.29 &	50.97 	 \\ \hline
-6.60 &	49.99 &	50.30 &	48.56 &	49.99 	 \\ \hline
-6.80 &	49.58 &	49.48 &	48.96 &	49.48 	 \\ \hline
-7.00 &	49.11 &	50.12 &	50.39 &	50.12 	 \\ \hline
    \end{tabular}
\end{center}

\subsection*{SVHN}

\begin{center}
    \begin{tabular}{ | l || c | c | c | c | c | c | c | c | c | c | c | }
    \hline
    \multicolumn{12}{|c|}{SVHN results with no regularization}   \\ \hline
    Seed & 1 & 2 & 3 & 4 & 5 & 6 & 7 & 8 & 9 & 10 & Median \\ \hline
    Error & 7.29 &	6.58 &	6.47 &	7.33 &	6.55 &	6.66 &	6.78 &	6.55 &	6.90 &	6.59 & \color{blue}6.63\color{black}  \\ \hline
    \end{tabular}
\end{center}

\begin{center}
    \begin{tabular}{ | c | c | c | c | c |   }
    \hline
    \multicolumn{5}{|c|}{SVHN results with Lasso regularization}   \\ \hline
    $\log(\lambda)$ & Seed $=$ 1 & Seed $=$ 2 & Seed $=$ 3 & Median \\ \hline
-4.00 &	6.92 &	7.05 &	7.15 &	7.05 	 \\ \hline
-4.20 &	6.63 &	7.33 &	6.17 &	6.63 	 \\ \hline
-4.40 &	6.18 &	6.86 &	6.20 &	6.20 	 \\ \hline
-4.60 &	6.05 &	6.44 &	6.95 &	6.44 	 \\ \hline
-4.80 &	6.53 &	6.50 &	6.70 &	6.53 	 \\ \hline
-5.00 &	6.69 &	5.61 &	6.19 &	6.19 	 \\ \hline
-5.20 &	6.06 &	6.31 &	5.81 &	6.06 	 \\ \hline
-5.40 &	5.42 &	5.34 &	5.70 &	5.42 	 \\ \hline
-5.60 &	5.46 &	5.24 &	5.32 &	\color{blue}5.32\color{black} 	 \\ \hline
-5.80 &	5.55 &	5.54 &	5.39 &	5.54 	 \\ \hline
-6.00 &	5.28 &	5.40 &	5.48 &	5.40 	 \\ \hline
-6.20 &	7.78 &	6.90 &	7.67 &	7.67 	 \\ \hline
-6.40 &	7.37 &	6.78 &	6.41 &	6.78 	 \\ \hline
-6.60 &	7.00 &	6.80 &	7.36 &	7.00 	 \\ \hline
-6.80 &	6.52 &	6.43 &	7.00 &	6.52 	 \\ \hline
-7.00 &	6.70 &	6.73 &	7.58 &	6.73 	 \\ \hline
    \end{tabular}
\end{center}

\begin{center}
    \begin{tabular}{ | c | c | c | c | c |   }
    \hline
    \multicolumn{5}{|c|}{SVHN results with $L_{2}$ regularization}   \\ \hline
    $\log(\lambda)$ & Seed $=$ 1 & Seed $=$ 2 & Seed $=$ 3 & Median \\ \hline
-4.00 &	5.56 &	5.83 &	5.42 &	5.56 	 \\ \hline
-4.20 &	5.37 &	5.49 &	5.82 &	5.49 	 \\ \hline
-4.40 &	5.60 &	5.75 &	5.54 &	5.60 	 \\ \hline
-4.60 &	5.29 &	5.59 &	5.22 &	\color{blue}5.29\color{black} 	 \\ \hline
-4.80 &	7.49 &	7.07 &	6.56 &	7.07 	 \\ \hline
-5.00 &	6.73 &	7.15 &	6.50 &	6.73 	 \\ \hline
-5.20 &	7.48 &	6.42 &	7.54 &	7.48 	 \\ \hline
-5.40 &	6.91 &	6.42 &	7.02 &	6.91 	 \\ \hline
-5.60 &	6.29 &	6.48 &	6.96 &	6.48 	 \\ \hline
-5.80 &	7.18 &	7.40 &	7.55 &	7.40 	 \\ \hline
-6.00 &	6.95 &	7.11 &	7.48 &	7.11 	 \\ \hline
-6.20 &	6.83 &	6.52 &	6.54 &	6.54 	 \\ \hline
-6.40 &	6.32 &	7.62 &	6.92 &	6.92 	 \\ \hline
-6.60 &	7.13 &	6.91 &	6.84 &	6.91 	 \\ \hline
-6.80 &	6.47 &	7.23 &	7.17 &	7.17 	 \\ \hline
-7.00 &	6.76 &	6.63 &	6.15 &	6.63 	 \\ \hline
    \end{tabular}
\end{center}

\begin{center}
    \begin{tabular}{ | c | c | c | c | c |   }
    \hline
    \multicolumn{5}{|c|}{SVHN results with Laplace ($\varepsilon$ = 1e-07) regularization}   \\ \hline
    $\log(\lambda)$ & Seed $=$ 1 & Seed $=$ 2 & Seed $=$ 3 & Median \\ \hline
-4.00 &	8.74 &	7.94 &	7.65 &	7.94 	 \\ \hline
-4.20 &	6.93 &	7.76 &	7.51 &	7.51 	 \\ \hline
-4.40 &	6.97 &	8.44 &	7.67 &	7.67 	 \\ \hline
-4.60 &	7.71 &	7.56 &	7.52 &	7.56 	 \\ \hline
-4.80 &	6.82 &	7.24 &	7.61 &	7.24 	 \\ \hline
-5.00 &	6.85 &	6.94 &	7.25 &	6.94 	 \\ \hline
-5.20 &	7.21 &	7.24 &	6.13 &	7.21 	 \\ \hline
-5.40 &	7.14 &	6.71 &	6.41 &	6.71 	 \\ \hline
-5.60 &	6.84 &	6.41 &	6.43 &	\color{blue}6.43\color{black} 	 \\ \hline
-5.80 &	6.54 &	6.50 &	7.05 &	6.54 	 \\ \hline
-6.00 &	6.52 &	6.86 &	6.71 &	6.71 	 \\ \hline
-6.20 &	6.65 &	6.75 &	6.44 &	6.65 	 \\ \hline
-6.40 &	6.71 &	6.67 &	6.68 &	6.68 	 \\ \hline
-6.60 &	7.41 &	6.80 &	6.60 &	6.80 	 \\ \hline
-6.80 &	6.57 &	6.74 &	6.97 &	6.74 	 \\ \hline
-7.00 &	6.94 &	6.71 &	6.96 &	6.94 	 \\ \hline
    \end{tabular}
\end{center}

\begin{center}
    \begin{tabular}{ | c | c | c | c | c |   }
    \hline
    \multicolumn{5}{|c|}{SVHN results with Arctan ($\gamma$ = 1) regularization}   \\ \hline
    $\log(\lambda)$ & Seed $=$ 1 & Seed $=$ 2 & Seed $=$ 3 & Median \\ \hline
-4.00 &	6.33 &	6.22 &	6.97 &	6.33 	 \\ \hline
-4.20 &	6.71 &	7.26 &	6.69 &	6.71 	 \\ \hline
-4.40 &	6.33 &	6.43 &	6.37 &	6.37 	 \\ \hline
-4.60 &	6.67 &	6.59 &	7.27 &	6.67 	 \\ \hline
-4.80 &	6.09 &	6.91 &	6.63 &	6.63 	 \\ \hline
-5.00 &	5.90 &	5.58 &	5.69 &	5.69 	 \\ \hline
-5.20 &	5.69 &	5.29 &	5.83 &	5.69 	 \\ \hline
-5.40 &	6.00 &	5.55 &	5.45 &	5.55 	 \\ \hline
-5.60 &	5.61 &	5.49 &	5.52 &	5.52 	 \\ \hline
-5.80 &	5.37 &	5.23 &	5.40 &	\color{blue}5.37\color{black} 	 \\ \hline
-6.00 &	6.63 &	6.66 &	6.82 &	6.66 	 \\ \hline
-6.20 &	7.01 &	6.81 &	7.26 &	7.01 	 \\ \hline
-6.40 &	6.49 &	6.62 &	7.06 &	6.62 	 \\ \hline
-6.60 &	7.02 &	6.60 &	6.88 &	6.88 	 \\ \hline
-6.80 &	6.77 &	6.70 &	6.92 &	6.77 	 \\ \hline
-7.00 &	6.48 &	7.08 &	6.83 &	6.83 	 \\ \hline
    \end{tabular}
\end{center}

\begin{center}
    \begin{tabular}{ | c | c | c | c | c |   }
    \hline
    \multicolumn{5}{|c|}{SVHN results with Arctan ($\gamma$ = 100) regularization}   \\ \hline
    $\log(\lambda)$ & Seed $=$ 1 & Seed $=$ 2 & Seed $=$ 3 & Median \\ \hline
-4.00 &	13.01 &	16.90 &	12.88 &	13.01 	 \\ \hline
-4.20 &	10.23 &	9.96 &	9.49 &	9.96 	 \\ \hline
-4.40 &	8.02 &	9.23 &	14.08 &	9.23 	 \\ \hline
-4.60 &	7.51 &	7.68 &	7.97 &	7.68 	 \\ \hline
-4.80 &	5.73 &	5.82 &	6.05 &	5.82 	 \\ \hline
-5.00 &	5.98 &	5.67 &	5.96 &	5.96 	 \\ \hline
-5.20 &	5.61 &	5.80 &	5.70 &	5.70 	 \\ \hline
-5.40 &	5.44 &	5.75 &	5.57 &	5.57 	 \\ \hline
-5.60 &	5.32 &	5.54 &	5.31 &	5.32 	 \\ \hline
-5.80 &	5.50 &	5.47 &	5.33 &	5.47 	 \\ \hline
-6.00 &	5.52 &	5.33 &	5.27 &	5.33 	 \\ \hline
-6.20 &	5.39 &	5.50 &	5.41 &  5.41	 \\ \hline
-6.40 &	5.28 &	5.29 &  5.27 &  5.28	 \\ \hline
-6.60 &	5.19 &	5.17 &  5.27 & \color{blue} 5.19\color{black}     \\ \hline
-6.80 &	5.13 &	5.23 &	5.28 &	5.23 	 \\ \hline
-7.00 &	4.96 &	5.27 &	5.19 & \color{blue} 5.19\color{black} 	 \\ \hline
    \end{tabular}
\end{center}

\subsection*{Reduced Size ImageNet}

\begin{center}
    \begin{tabular}{ | l || c | c | c | c | c | c | c | c | c | c | c | }
    \hline
    \multicolumn{12}{|c|}{ImageNet results with no regularization}   \\ \hline
    Seed & 1 & 2 & 3 & 4 & 5 & 6 & 7 & 8 & 9 & 10 & Median \\ \hline
    Error & 43.86 &	43.09 &	43.31 &	42.36 &	43.29 &	42.44 &	42.24 &	42.27 &	42.60 &	43.60 & \color{blue}42.85\color{black}  \\ \hline
    \end{tabular}
\end{center}

\begin{center}
    \begin{tabular}{ | c | c |    }
    \hline
    \multicolumn{2}{|c|}{ImageNet results with} \\ \hline
    \multicolumn{2}{|c|}{Lasso regularization}   \\ \hline
    $\log(\lambda)$ & Seed $=$ 1 \\ \hline
-4.00 &	 44.65    \\ \hline
-4.20 &  43.82 	 \\ \hline
-4.40 &  41.92	 \\ \hline
-4.60 &	 41.33	 \\ \hline
-4.80 &	 \color{blue}41.32\color{black}	 \\ \hline
-5.00 &	 44.08	 \\ \hline
-5.20 &	 45.70    \\ \hline
-5.40 &	 41.77	 \\ \hline
-5.60 &	 41.60	 \\ \hline
-5.80 &	 43.14	 \\ \hline
-6.00 &  42.24	 \\ \hline
-6.20 &  42.60	 \\ \hline
-6.40 &	 42.49	 \\ \hline
-6.60 &	 42.46	 \\ \hline
-6.80 &	 42.07	 \\ \hline
-7.00 &	 42.82	 \\ \hline
    \end{tabular}
\end{center}

\begin{center}
    \begin{tabular}{ | c | c |   }
    \hline
    \multicolumn{2}{|c|}{ImageNet results with} \\ \hline
    \multicolumn{2}{|c|}{$L_{2}$ regularization}   \\ \hline
    $\log(\lambda)$ & Seed $=$ 1  \\ \hline
-4.00 & \color{blue}41.55\color{black}    \\ \hline
-4.20 &	42.29 	 \\ \hline
-4.40 &	42.27 	 \\ \hline
-4.60 &	42.32 	 \\ \hline
-4.80 &	41.94 	 \\ \hline
-5.00 &	42.38 	 \\ \hline
-5.20 &	41.80    \\ \hline
-5.40 &	42.19 	 \\ \hline
-5.60 &	42.09 	 \\ \hline
-5.80 &	43.60 	 \\ \hline
-6.00 &	42.18 	 \\ \hline
-6.20 &	42.98 	 \\ \hline
-6.40 &	42.45 	 \\ \hline
-6.60 &	43.22 	 \\ \hline
-6.80 &	42.77 	 \\ \hline
-7.00 &	43.10 	 \\ \hline
    \end{tabular}
\end{center}

\begin{center}
    \begin{tabular}{ | c | c |    }
    \hline
    \multicolumn{2}{|c|}{ImageNet results with} \\ \hline
    \multicolumn{2}{|c|}{Laplace ($\varepsilon$ = 1e-07) regularization}   \\ \hline
    $\log(\lambda)$ & Seed $=$ 1  \\ \hline
-4.00 &   63.59   \\ \hline
-4.20 &   50.20	 \\ \hline
-4.40 &	  45.81	 \\ \hline
-4.60 &	  45.37	 \\ \hline
-4.80 &	  43.31	 \\ \hline
-5.00 &	  43.59	 \\ \hline
-5.20 &	  44.85   \\ \hline
-5.40 &	  43.67	 \\ \hline
-5.60 &	  44.66	 \\ \hline
-5.80 &   44.27	 \\ \hline
-6.00 &	  44.54	 \\ \hline
-6.20 &	  42.31	 \\ \hline
-6.40 &	  42.38	 \\ \hline
-6.60 &	  42.23	 \\ \hline
-6.80 &	  42.37	 \\ \hline
-7.00 &	  \color{blue}41.99\color{black}	 \\ \hline
    \end{tabular}%
\end{center}

\normalsize

\section*{Appendix C: Tables of Pairwise Comparison p-values (Supplementary Material)}

\scriptsize

\begin{center}
    %
\end{center}

\end{document}